\documentclass[lettersize,journal]{IEEEtran}
\usepackage{times}
\usepackage{epsfig}
\usepackage{graphicx}
\usepackage{amsmath}
\usepackage{amssymb}
\usepackage{booktabs}
\usepackage{algorithm}
\usepackage{algpseudocode}
\usepackage[table,xcdraw,dvipsnames]{xcolor}
\usepackage{multirow}
\usepackage{subcaption}
\usepackage{enumitem}
\usepackage{balance}
\usepackage{mathrsfs} 
\usepackage{bm}
\usepackage{wrapfig}
\usepackage{bbm}

\makeatletter
\newcommand*\bigcdot{\mathpalette\bigcdot@{.5}}
\newcommand*\bigcdot@[2]{\mathbin{\vcenter{\hbox{\scalebox{#2}{$\m@th#1\bullet$}}}}}
\makeatother

\definecolor{c2}{HTML}{FBD9BD}

\usepackage[pagebackref=true,breaklinks=true,letterpaper=true,colorlinks,bookmarks=false]{hyperref}

\usepackage[capitalize]{cleveref}
\crefname{section}{Sec.}{Secs.}
\Crefname{section}{Section}{Sections}
\Crefname{table}{Table}{Tables}

\ifCLASSOPTIONcompsoc
  \usepackage[nocompress]{cite}
\else
  \usepackage{cite}
\fi

\ifCLASSINFOpdf
\else
\fi

\begin{document}
%

\title{Segment Concealed Objects with \\ Incomplete Supervision}

\author{
Chunming~He,~\IEEEmembership{}
Kai~Li,~
Yachao~Zhang,~\IEEEmembership{} 
Ziyun~Yang,~
Youwei~Pang,~
Longxiang~Tang,~\IEEEmembership{} \\
Chengyu~Fang,~ 
Yulun~Zhang,~
Linghe~Kong,~
Xiu~Li,~
and Sina Farsiu,~\IEEEmembership{Fellow,~IEEE.}
\\ \href{https://github.com/ChunmingHe/SEE}{Official Code} \ \ \ \ \ \ \ \  \href{https://ieeexplore.ieee.org/abstract/document/11023026}{TPAMI Version} 
\IEEEcompsocitemizethanks{
\IEEEcompsocthanksitem This research was supported by the Foundation Fighting Blindness (BR-CL-0621-0812-DUKE); Research to Prevent Blindness (Unrestricted Grant to Duke University), and Foundation Fighting Blindness (PPA-1224-0890-DUKE). (\textit{Corresponding author: Sina Farsiu (E-mail: sina.farsiu@duke.edu) and Kai Li (e-mail: li.gml.kai@gmail.com)})
\IEEEcompsocthanksitem Chunming He, Ziyun Yang, Sina Farsiu are with the Department of Biomedical Engineering, Duke University, Durham, NC 27708 USA (e-mail: chunming.he@duke.edu,
sina.farsiu@duke.edu).
\IEEEcompsocthanksitem Kai Li is with Meta (e-mail: li.gml.kai@gmail.com).
\IEEEcompsocthanksitem Yachao Zhang, Longxiang Tang, Chengyu Fang, and Xiu Li are with Tsinghua Shenzhen International Graduate School, Tsinghua University, Shenzhen 518055, China. 
\IEEEcompsocthanksitem Youwei Pang is with the School of Information and Communication Engineering, Dalian University of Technology, Dalian 116024, China.
\IEEEcompsocthanksitem Yulun Zhang and Linghe Kong are with Shanghai Jiao Tong University, Shanghai 200240, China.
}
}

\markboth{IEEE TPAMI}%
{He \MakeLowercase{\textit{et al.}}: Segment Concealed Objects with Incomplete Supervision}

\maketitle
\begin{abstract}
Incompletely-Supervised Concealed Object Segmentation (ISCOS) involves segmenting objects that seamlessly blend into their surrounding environments, utilizing incompletely annotated data, such as weak and semi-annotations, for model training. This task remains highly challenging due to (1) the limited supervision provided by the incompletely annotated training data, and (2) the difficulty of distinguishing concealed objects from the background, which arises from the intrinsic similarities in concealed scenarios. In this paper, we introduce the first unified method for ISCOS to address these challenges. To tackle the issue of incomplete supervision, we propose a unified mean-teacher framework, SEE, that leverages the vision foundation model, ``\emph{Segment Anything Model (SAM)}'', to generate pseudo-labels using coarse masks produced by the teacher model as prompts. To mitigate the effect of low-quality segmentation masks, we introduce a series of strategies for pseudo-label generation, storage, and supervision. These strategies aim to produce informative pseudo-labels, store the best pseudo-labels generated, and select the most reliable components to guide the student model, thereby ensuring robust network training. Additionally, to tackle the issue of intrinsic similarity, we design a hybrid-granularity feature grouping module that groups features at different granularities and aggregates these results. By clustering similar features, this module promotes segmentation coherence, facilitating more complete segmentation for both single-object and multiple-object images. We validate the effectiveness of our approach across multiple ISCOS tasks, and experimental results demonstrate that our method achieves state-of-the-art performance. Furthermore, SEE can serve as a plug-and-play solution, enhancing the performance of existing models.
\end{abstract}


\IEEEpeerreviewmaketitle

\setlength{\abovedisplayskip}{2pt}
\setlength{\belowdisplayskip}{2pt}

\section{Introduction}\label{sec:introduction}
{\IEEEPARstart{C}{oncealed} object segmentation (COS) targets identifying objects that blend seamlessly with their environments \cite{fan2021concealed,he2019image}. COS encompasses a variety of applications, including camouflaged object detection \cite{he2023strategic}, polyp image segmentation \cite{xiao2024survey}, and transparent object detection \cite{xiao2023concealed}.

The primary challenge in COS lies in distinguishing foreground objects from backgrounds with similar visual characteristics. To overcome this, researchers have developed several innovative methods. Some methods simulate human visual perception \cite{he2023strategic,COD-ZoomNeXt}, while others utilize frequency information \cite{zhong2022detecting,He2023Camouflaged}. Additionally, joint modeling techniques across multiple tasks have been explored to enhance accuracy \cite{he2024diffusion,CDCU-Spider}.

Incompletely Supervised COS (ISCOS), which encompasses semi-supervised COS (SSCOS) and weakly supervised COS (WSCOS), addresses a more challenging yet practical problem: learning a COS model without fully annotated training data. SSCOS significantly reduces annotation costs by fully labeling only a small portion of the dataset, while WSCOS achieves a similar cost reduction by requiring only a few annotated points or scribbles in the foreground or background. However, the sparsity of annotated training data in WSCOS and SSCOS limits the segmenter's discriminative capacity during model learning, thereby restricting segmentation performance in highly challenging tasks.

Several strategies are designed to address incomplete supervision, with pseudo-labeling emerging as a promising direction \cite{he2023weaklysupervised,lai2024camoteacher,fang2024real,he2023reti,he2025unfoldir}. \cref{fig:FrameComp} presents several pseudo-labeling strategies, including the widely used semi-supervised and recently proposed weakly supervised frameworks. While they have verified strong capacity, they also exhibit clear limitations. The mean-teacher framework \cite{lai2024camoteacher} employs a teacher model to generate pseudo-labels that supervise the student model using a Siamese structure. However, this approach often produces low-quality pseudo-labels, especially in the early training stages, which undermines the robustness of the guidance provided and negatively impacts segmentation performance (See SSCOD and SSTOD in \cref{fig:FrameComp}). 
To produce higher-quality pseudo-labels, WS-SAM~\cite{he2023weaklysupervised}, a preliminary version of our work, leverages the vision foundation model, \emph{Segment Anything Model (SAM)}, to generate more reliable pseudo-labels. However, this framework depends on manually annotated prompts to guide SAM in segmentation, limiting its applicability to weakly supervised scenarios. Besides, as depicted in the two weakly supervised tasks in \cref{fig:FrameComp}, WS-SAM delivers suboptimal performance in ISCOS tasks due to its incomplete exploitation of SAM’s prompt potential and insufficient consideration of concealed objects' unique characteristics.

\begin{figure*}[ht]
	\centering
	\setlength{\abovecaptionskip}{-0.1cm}
	\begin{center}
		\includegraphics[width=\linewidth]{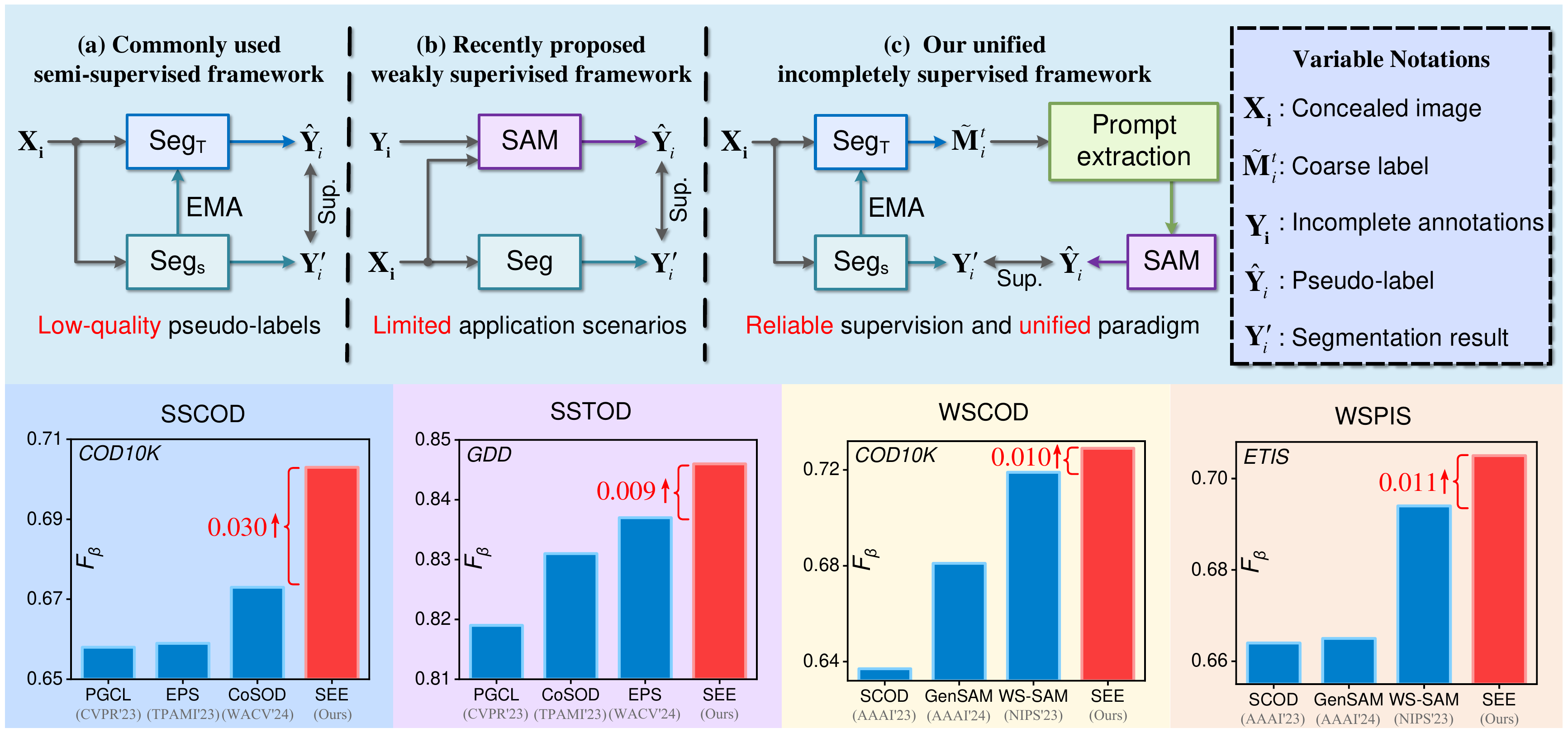}
	\end{center}
\caption{Existing pseudo-labeling frameworks. Compared with previous methods ((a) and (b)), our SEE framework (c) is a unified paradigm that can generate high-quality pseudo-labels and provide segmenters with reliable supervision. In the bottom line, our SEE framework outperforms cutting-edge techniques on four ISCOS tasks, including semi-supervised camouflaged object detection (SSCOD), semi-supervised transparent object detection (SSTOD), weakly supervised camouflaged object detection (WSCOD), and weakly supervised polyp image segmentation (WSPIS). See details in~\cref{Sec:Experiment}.
	}
	\label{fig:FrameComp}
	\vspace{-0.3cm}
\end{figure*}

In this paper, we propose a novel algorithm to tackle the aforementioned challenges of ISCOS tasks. To overcome incomplete supervision, we propose a unified mean-teacher framework, SEE, that harnesses the power of \emph{Segment Anything Model (SAM)} to generate dense masks from the coarse results produced by the teacher model. These generated masks serve as pseudo-labels to train a segmenter. However, due to the intrinsic similarity between foreground objects and the background, the pseudo-labels generated by SAM may not always be reliable. To address this, we introduce a series of strategies focused on pseudo-label generation, storage, and supervision, as illustrated in~\cref{fig:SWSCOS,Alg:Camouflageator}.

For pseudo-label generation, we first propose generating multiple augmentation views for each image and fusing the segmentation masks produced by the teacher model across all views. 
This fusion process enhances the robustness of predictions by focusing on those that are consistent across augmentations, thereby resulting in more accurate and complete masks due to the ensemble effect.
Next, we extract multi-density prompts—points, boxes, and masks—from the fused coarse masks to distill essential information for SAM. This approach not only provides SAM with diverse and comprehensive guidance that aligns with its prompt requirements but also adapts to the evolving accuracy of the mean-teacher model over different training stages. 
As training progresses, increasingly accurate information can be distilled from the coarse fused masks, providing SAM with more reliable guidance. The same ``augmentation then fusion" strategy is applied to SAM, further ensuring the generation of more accurate and comprehensive pseudo-labels.

Moreover, since the quality of generated pseudo-labels does not necessarily improve consistently throughout network training, we propose storing a selection of the most informative and beneficial pseudo-labels for each concealed object. This is achieved using an entropy-based uncertainty map that evaluates each pseudo-label's informativeness, \textit{i.e.}, the extent to which it contains regions with low uncertainty, and its utility to the student model, focusing on areas where the model has not yet achieved high certainty.
The variations in pseudo-labels generated at different epochs, due to differences in extracted prompts, complement each other and help to more effectively reveal concealed objects, ultimately leading to improved segmentation performance.

For pseudo-label supervision, simply discarding valuable but potentially confusing pseudo-labels could hinder model generalization, while using them without refinement risks exacerbating confirmation bias due to mislabeling. 
To address this, we propose a dual strategy that selects reliable pseudo-labels and emphasizes their trustworthy regions during supervision. This is achieved through both image-level and pixel-level strategies. The image-level selection relies on an entropy-based evaluation function to assess the quality of each generated mask and determine whether to include the pseudo-label in training. The pixel-level weighting strategy, also entropy-based, assigns higher weights to predictions with higher certainty. Together, these strategies ensure that only the most reliable regions of the pseudo-labels are emphasized during training, overcoming the deceptive nature of concealed objects and leading to more robust network performance.

Additionally, to overcome the difficulty posed by the intrinsic visual similarity between foreground and background, we present a Hybrid-Granularity Feature Grouping (HGFG) module. This module operates by first extracting discriminative features at multiple granular levels and then integrating these features to effectively address various concealing scenarios. By performing feature grouping, HGFG enhances the features' coherence, thus reducing the likelihood of incomplete segmentation. It achieves this by promoting local feature correlations within individual objects while simultaneously enabling accurate segmentation of multiple objects by establishing global coherence across different objects.

\begin{figure*}[ht]
	\centering
	\setlength{\abovecaptionskip}{-0.1cm}
	\begin{center}
		\includegraphics[width=\linewidth]{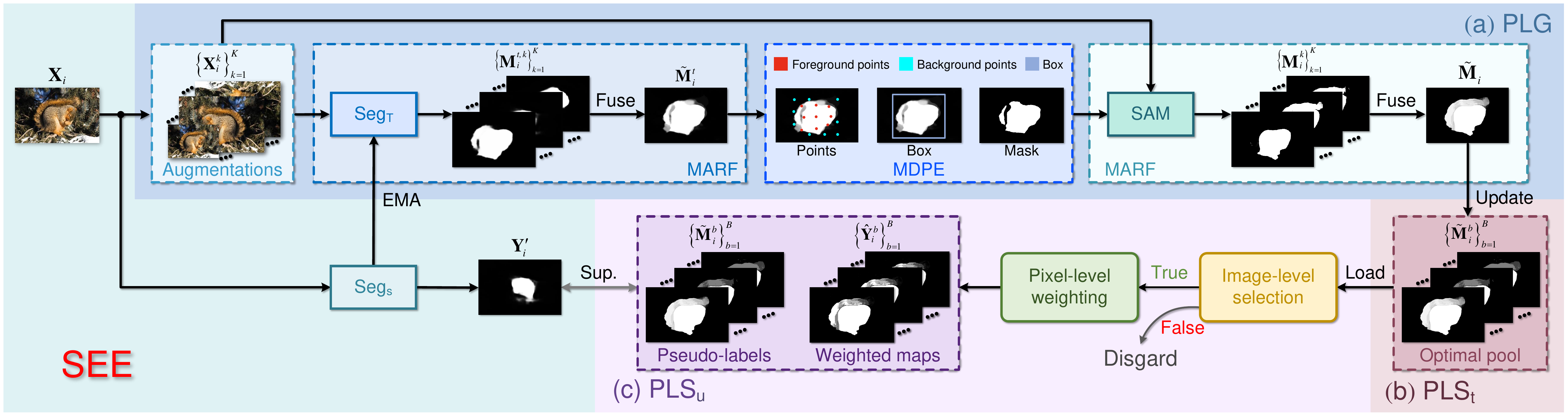}
	\end{center}
	\caption{Framework of SEE for ISCOS tasks. (a) In pseudo-label generation (PLG), SEE uses SAM to generate pseudo-labels with multi-augmentation result fusion (MARF) and multi-density prompt extraction (MDPE) strategies. (b) Then SEE stores the best pseudo-labels generated in an optimal pool for pseudo-label storage ($\text{PLS}_{\text{t}}$). (c) In pseudo-label supervision ($\text{PLS}_{\text{u}}$), SEE selects the most reliable components of these stored pseudo-labels to supervise the student model. These strategies cooperatively ensure robust network training. 
	}
	\label{fig:SWSCOS}
	\vspace{-0.3cm}
\end{figure*}

Our contributions are summarized as follows:}
\begin{itemize}
	\setlist{nolistsep}
	\item[$\bullet$] 
{ We propose utilizing SAM for ISCOS tasks to generate dense segmentation masks for training segmenters. To the best of our knowledge, this is the first attempt to leverage the vision foundation model to address ISCOS tasks.}
	\item[$\bullet$] {We propose a unified mean-teacher framework, SEE, for ISCOS tasks, which comprises a series of pseudo-label generation, storage, and supervision strategies to produce high-quality pseudo-labels. These strategies offer reliable and accurate guidance for model training, thereby leading to improved segmentation performance.
 }
	\item[$\bullet$] {We introduce HGFG to address the intrinsic similarity challenge in ISCOS tasks. HGFG extracts discriminative cues by grouping features at multiple granularities, enhancing coherence and enabling complete segmentation.
 }
	\item[$\bullet$] {We evaluate our method on six ISCOS tasks to verify that our method achieves state-of-the-art performance and that SEE is a plug-and-play framework.  }
\end{itemize}

{A preliminary version of this work was published in~\cite{he2023weaklysupervised}. In this paper, we extend our conference version in the following aspects. 
First, we \textbf{explore semi-supervised COS tasks}. Our target task has been extended from the weakly-supervised COS task to incompletely-supervised COS tasks, incorporating a semi-supervision setting. 
Second, we \textbf{enhance the weakly-supervised WS-SAM framework into the incompletely-supervised unified SEE framework}. This enhancement enables segmentation methods to effectively "SEE" concealed objects even with limited annotations during training. 
Thirdly, we \textbf{improve the performance of our SEE framework by introducing structural extensions} in the pseudo-label generation, restoration, and supervision, such as multi-density prompt extraction and the optimal label pool. These enhancements make our approach not only superior to the conference version, the WS-SAM framework, but also establish it as a more robust plug-and-play framework for improving the performance of existing state-of-the-art segmenters under incomplete supervision. Fourth, we \textbf{introduce and discuss more recently published algorithms for ISCOS tasks} to track developments in the field and ensure fair, comprehensive, and rigorous comparisons. 
Finally, we \textbf{provide additional analytical experiments and visualization results} to further explore the potential of our methods and verify the effectiveness of our SEE framework and HGFG, 
as well as the generalizability of SEE in serving as a plug-and-play technique.}

\section{Related Work}\label{sec:relatedwork}
{\noindent \textbf{Segment Anything Model}. SAM~\cite{kirillov2023segment}, a milestone vision foundation model trained on
over 1 billion masks, aims to segment any object in any given image without requiring additional task-specific adaptation. Its outstanding quality in segmentation results and zero-shot generalization to new scenes make SAM a promising candidate for various computer vision tasks~\cite{tang2023can}. However, recent studies have highlighted that SAM encounters difficulties when segmenting objects with poor visibility, such as camouflaged objects~\cite{ji2023sam,ji2023segment}, medical polyps~\cite{mazurowski2023segment,he2023HQG}, and transparent glasses~\cite{ji2023segment}. These findings suggest that SAM still has limitations in COS tasks. 

To address this issue, SAM-Adapter~\cite{chen2023sam} employs an adapter to fine-tune the framework, improving results by implicitly enhancing data alignment. However, SAM-Adapter achieves suboptimal results under incomplete supervision due to data distribution biases introduced by incomplete annotations. Several strategies, such as WS-SAM~\cite{he2023weaklysupervised} and GenSAM~\cite{hu2024relax}, focus on weak supervision and leverage the pretrained SAM model to generate pseudo-labels for network training. Nonetheless, these methods rely heavily on manually annotated prompts, which limits their applicability to semi-supervised tasks. Additionally, both approaches fall short of fully exploring SAM’s prompt potential and fail to adequately account for the unique characteristics of concealed objects, leading to suboptimal performance in the ISCOS task.

In this paper, we propose the first unified framework, SEE, to address ISCOS tasks. SEE employs an iterative mean-teacher structure that utilizes SAM to generate high-quality pseudo-labels for network training. The framework introduces a series of strategies for pseudo-label generation, storage, and supervision, designed to fully leverage the strengths of both SAM and the mean-teacher structure. By accounting for the unique challenges of concealed objects, such as hidden discriminative cues and ambiguous boundaries, SEE provides precise and stable guidance throughout the learning process. Ultimately, SEE functions as a plug-and-play solution, boosting the segmentation performance of existing ISCOS models.}

{\noindent \textbf{Concealed Object Segmentation}. With the advancement of deep learning, learning-based segmenters have achieved significant success in fully-supervised COS tasks~\cite{fan2020pranet,fan2020camouflaged,lin2021rich,zhang2025referring,yin2024camoformer,he2025run}. 
PraNet~\cite{fan2020pranet} proposed a parallel reverse attention network for polyp segmentation in colonoscopy images. 
{ACUMEN~\cite{zhang2024unlocking} introduced the COD-TAX dataset and developed a robust framework for joint textual-visual understanding. ProMaC~\cite{hu2025leveraging} harnessed the capabilities of MLLMs, focusing on mitigating hallucinations during the reasoning phase to generate more precise instance-specific prompts.
}
Inspired by biological processes, Camouflageator~\cite{he2023strategic} introduced a predator network to locate and capture camouflaged objects, while GSDNet~\cite{lin2021rich} designed a reflection refinement module to improve the detection of transparent objects. 
However, research on incompletely supervised COS tasks remains limited. SCOD~\cite{he2022weakly} brought the first weakly-supervised COD framework, but it relies only on sparse annotations, which limits its discriminatory capacity and inhibits performance. Besides, CamoTeacher~\cite{lai2024camoteacher} developed the first semi-supervised COD method, updating the student model using pseudo-labels generated by the teacher model. However, the initial instability of these pseudo-labels restricts the student model's optimization. 

To address these challenges, we propose SEE, a unified framework for incompletely supervised COS tasks that utilizes SAM to generate high-quality pseudo-labels, facilitating robust network optimization.
Furthermore, to address the intrinsic similarity challenge in COS, we present the hybrid-granularity feature grouping module. This module extracts discriminative features at different granularities, thereby enhancing feature coherence and improving segmentation performance.
}

\setlength{\textfloatsep}{4pt}
\begin{algorithm}[t]
	\caption{Updating the student model under the SEE framework with pseudo-labels.}
	\label{Alg:Camouflageator}
	\textbf{Input}: concealed object image $\mathbf{X_i}$,  incomplete annotations $\mathbf{Y_i}$, segment anything model $SAM$, epoch number $N$ \\
	\textbf{Output}: student model $S_s$, teacher model $S_t$ 
	\begin{algorithmic}[1]
		\State Initialize the student model $S_s$ and the teacher model $S_t$
		\State current epoch n $\gets 0$
		\For{each epoch $n\in \left[1,N\right]$} \\
        \textcolor{gray}{// Pseudo-label generation}
            \State $\tilde{\mathbf{M}}^t_i$ $ \leftarrow$ coarse fused mask generated by $S_t$ \Comment{\cref{Eq:CoarseMask}}
            \State $\mathbf{Y}_i^p=\left\{\{\mathbf{Y}_i^{fp,c}\}_{c=1}^C, \{\mathbf{Y}_i^{bp,c}\}_{c=1}^C, \mathbf{Y}_i^{b}, \mathbf{Y}_i^{m}\right\}$ \ $ \leftarrow$ multi-density prompts extracted from $\tilde{\mathbf{M}}^t_i$ 
            \State $\tilde{\mathbf{M}}_i$ $ \leftarrow$ pseudo-label generated by $SAM$ with $\mathbf{Y_i^p}$ \Comment{\cref{Eq:RefinedPL}} 
            \\
            \textcolor{gray}{// Pseudo-label storage}
		\State $\{\tilde{\mathbf{M}}_i^{b,n}\}_{b=1}^B$ $ \leftarrow$ top-$B$ best labels at the $n^{th}$ epoch stored in the optimal pool $\mathcal{P}$ \Comment{\cref{Eq:pls}}
  \\
            \hspace{15pt} \textcolor{gray}{// Pseudo-label supervision}
            \State $\hat{\mathbf{Y}}_i^b$ $ \leftarrow$ weighted map processed by image-level selection and pixel-level weighting \Comment{\cref{Eq:PLSup}} 
		\\ \textcolor{gray}{// Update the student model}
  \\ \textbf{if} in the weakly supervised setting:
		\State $S_s^{(n)} \leftarrow$ update the student model with $L_w$ \Comment{\cref{Eq:WeakSup}}
     \\ \textbf{else if} in the semi-supervised setting:
   \State $S_s^{(n)} \leftarrow$ update the student model with $L_s$ \Comment{\cref{Eq:SemiSup}}
   \\ \textcolor{gray}{//  Update the teacher model}
        \State $S_t^{(n)}$ $\leftarrow$ update the teacher model with EMA \Comment{\cref{Eq:EMA}}
		\EndFor
	\end{algorithmic}
\end{algorithm}

\section{Methodology}\label{Sec:Method}
{Incompletely Supervised Concealed Object Segmentation (ISCOS), including semi-supervised COS (SSCOS) and weak-supervised COS (WSCOS), is an extremely challenging task to learn a segmentation model from an incompletely annotated training dataset $\mathcal{S} = \{\mathbf{X}_i, \mathbf{Y}_i\}_{i=1}^{S}$ and test the model on a test dataset $\mathcal{T} = \{\mathbf{T}_i\}_{i=1}^{T}$, where $\mathbf{X}_i$ and $\mathbf{T}_i$ represent the training and test images, respectively. $\mathbf{Y}_i$ denotes the incomplete annotations, where they can be missing for a part of images in the setting of SSCOS and a few points or scribbles annotated as foreground or background in the WSCOS task.    

Learning an ISCOS model is challenging because concealed objects often blend with their surroundings, making it difficult to distinguish the foreground from the background.  Moreover, the incomplete annotations, $\mathbf{Y}_i$, may not provide sufficient supervision for accurate dense predictions. To address these challenges, we propose an iterative mean-teacher framework, SEE, which leverages the power of \emph{Segment Anything Model (SAM)} to generate high-quality segmentation masks. 
As depicted in~\cref{fig:SWSCOS,Alg:Camouflageator},
our framework generates pseudo-labels, stores the best pseudo-labels generated, and selects the most reliable components of these stored pseudo-labels to supervise the student model, ensuring robust network training. Furthermore, as shown in~\cref{fig:MFG}, we propose a Hybrid-Granularity Feature Grouping (HGFG) module to group features at various granularities. This module promotes segmentation coherence, enabling the model to produce complete segmentation results even in diverse concealing scenarios. By integrating these two innovations, our method can accurately distinguish and segment concealed objects within complex environments even with incomplete supervision.
}

\begin{figure*}[t]
	\centering
	\setlength{\abovecaptionskip}{-0.2cm}
	\begin{center}
		\includegraphics[width=0.9\linewidth]{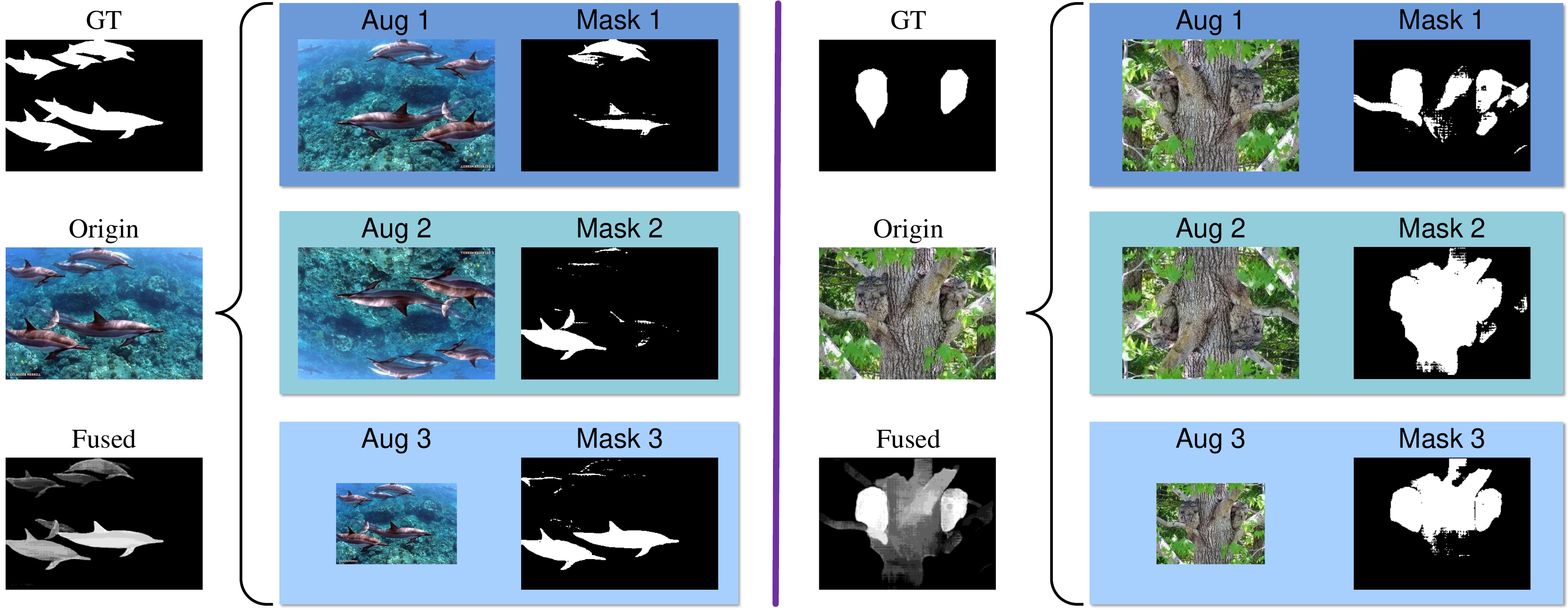}
	\end{center}
	\caption{Segmentation masks with different augmented images. We inversely transform the masks to keep consistent with the original image. It is observed that fused masks contain more accurate and complete segmentation information.
	}
	\label{fig:Augmentation}
	\vspace{-0.6cm}
\end{figure*}

\begin{figure}[t]
	\centering
	\setlength{\abovecaptionskip}{-0.2cm}
	\begin{center}
		\includegraphics[width=\linewidth]{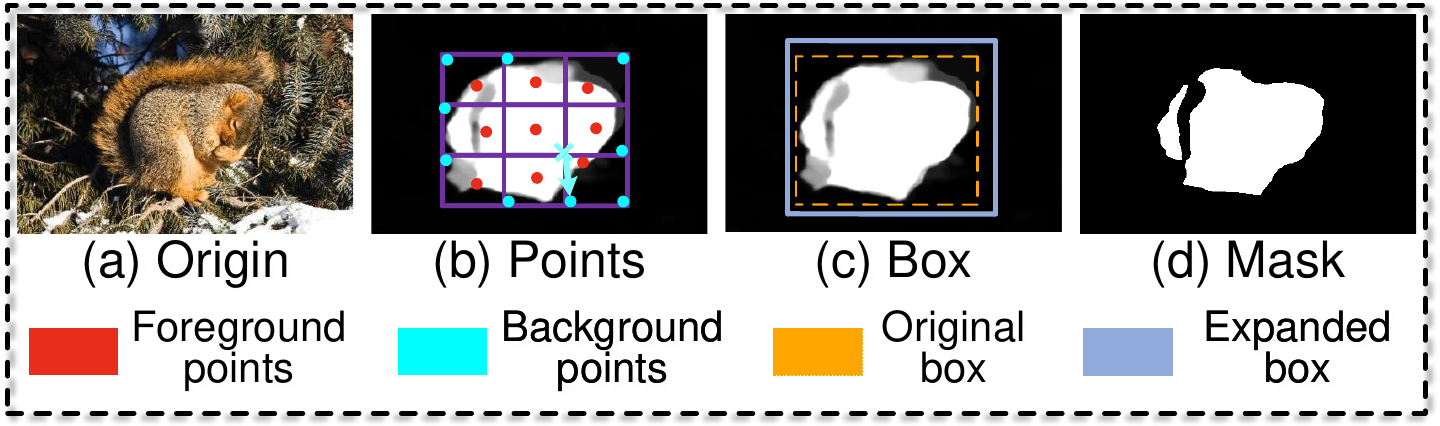}
	\end{center}
	\caption{Multi-density prompt extraction, including points, box, and mask. Foreground and background points are sampled by the Nine-block strategy. The box prompt is expanded in four directions based on their corresponding expansion coefficients.
	}
	\label{fig:NBS}
\end{figure}
\subsection{Pseudo Labeling with SAM}\label{Sec:SWSCOS}
{SAM is designed for generic object segmentation~\cite{kirillov2023segment}, showing impressive capabilities in producing precise segmentation masks across a wide range of object categories (hence the term ``\emph{segment anything}''). Nevertheless, contrary to claims that SAM has ``solved'' the segmentation task, we find that SAM falls short, at least in concealed object segmentation~\cite{he2023weaklysupervised}. One key limitation is that SAM requires ``prompts''—hints about the objects of interest—to perform segmentation. While these prompts can take various forms (\textit{e.g.}, points, masks, or bounding boxes), they must be provided by humans or other external sources, such as other algorithms. This reliance on additional input renders SAM unsuitable for applications where only test images are available. Additionally, although SAM performs admirably in general scene images, it struggles with concealed object images due to the intrinsic similarity between the foreground objects and the background.

This paper proposes a SAM-based iterative mean-teacher framework, SEE, for ISCOS tasks, including WSCOS and SSCOS, to generate high-quality pseudo-labels to train a COS model. Given that SAM-generated pseudo-labels may not be entirely reliable, as shown in~\cref{fig:SWSCOS,Alg:Camouflageator}, we propose a comprehensive pipeline, consisting of pseudo-label generation, storage, and supervision, to address this problem.}

\subsubsection{Pseudo-Label Generation}\label{Sec:PLG}
{To improve the accuracy of SAM-generated masks, we propose providing SAM with more precise and comprehensive prompts by designing two strategies: multi-augmentation result fusion (MARF) and multi-density prompt extraction (MDPE). As depicted in~\cref{fig:SWSCOS,Alg:Camouflageator}, MARF is first applied to the teacher model to generate a fused coarse mask. Next, MDPE extracts multi-density prompts, including points, boxes, and masks, from the coarse mask. Finally, MARF is also applied to SAM, along with the extracted prompts, to generate the fused pseudo-label, effectively capturing camouflaged objects.
}

{\noindent\textbf{Multi-augmentation result fusion}. Given a concealed image $(\mathbf{X}_i, \mathbf{Y}_i)\in\mathcal{S}$, we generate $K$ augmented images $\{\mathbf{X}^k_i\}^K_{k=1}$ by applying stochastic augmentations randomly sampled from combinations of image flipping, rotation ($0^{\circ}$, $90^{\circ}$, $180^{\circ}$, $270^{\circ}$), and scaling ($\times0.5$, $\times1.0$, $\times2.0$). We apply only weak augmentations, avoiding stronger ones. This is because ISCOS is an extremely challenging task, and strong augmentations may distort the concealed characteristics, thus negatively affecting segmentation performance in an incomplete supervision setting.
We send $\{\mathbf{X}^k_i\}^K_{k=1}$ to teacher model $S_t$ and generate segmentation masks $\{\mathbf{M}^{t,k}_i\}^K_{k=1}$, where 
\begin{equation}
	\mathbf{M}_i^{t,k}=S_t\left(\mathbf{X}^k_i\right),
\end{equation}
note that $\mathbf{M}_i^{t,k}$ shares the same structure with the input image $\mathbf{X}^k_i$, which may differ in shape from the original camouflaged image $\mathbf{X}_i$; we perform inverse image transformation to ensure all masks have the same shape as the original image.

As depicted in Fig. \ref{fig:Augmentation}, although the masks generated from different augmentations vary significantly in shape, they overlap in certain regions consistently predicted by the teacher model, irrespective of image transformations. These overlapping regions often correspond to accurately predicted foreground areas. Moreover, the masks complement one another, as regions of the foreground missed by one mask are captured by others. Based on these observations, we propose fusing the segmentation masks from different augmented images to improve overall accuracy, as
\begin{equation}\label{Eq:CoarseMask}
	\tilde{\mathbf{M}}^t_i = \frac{1}{K}\sum_{k=1}^K \mathbf{M}^{t,k}_i,
\end{equation}
where $\tilde{\mathbf{M}}^t_i$ is the fused coarse mask generated by the teacher model. We expect $\tilde{\mathbf{M}}^t_i$ to be more reliable than the individual masks, as it is an ensemble over multiple augmented images.

\noindent\textbf{Multi-density prompt extraction}. After obtaining the fused coarse mask $\tilde{\mathbf{M}}^t_i$, we propose using the information embedded in this mask to guide SAM in generating high-quality pseudo-labels. Simply using the coarse mask as a prompt is inadequate, as the teacher model's capacity is notably constrained during the early training stages, often resulting in inaccurate predictions that risk misguiding SAM. To overcome this, we introduce a multi-density prompt extraction (MDPE) strategy, which derives diverse prompts $\mathbf{Y}_i^{p}$—including points, boxes, and masks—from $\tilde{\mathbf{M}}^t_i$. These multi-density prompts collectively capture comprehensive information from the coarse mask, providing SAM with diverse and complementary guidance. This adaptive approach aligns with the evolving accuracy of the mean-teacher model throughout training, mitigating the adverse effects of low-quality predictions.

During the early training stages, when the teacher model can only approximate the location of the concealed object, accurate point prompts provide SAM with essential positional information. Hence, we adopt a nine-block strategy to extract foreground and background point prompts, $\{\mathbf{Y}_i^{fp,c}\}_{c=1}^C$ and $\{\mathbf{Y}_i^{bp,c}\}_{c=1}^C$. As shown in \cref{fig:NBS}, the minimum bounding box of the coarse mask is divided into nine equal blocks. Foreground points are selected based on pixel confidence values, with high-confidence pixels (value 1) chosen at the center of each block or, if unavailable, their nearest valid neighbors. Background points are identified as the farthest pixels relative to the foreground with a pixel value of 0, reducing interference from ambiguous regions. For weakly supervised settings, scribbles are processed using the same nine-block strategy to extract key point prompts effectively.

During the intermediate training phase, as the teacher model begins to detect camouflaged objects with finer detail, the extracted box prompt $\mathbf{Y}_i^{b}$ provides SAM with more precise spatial information, effectively minimizing the segmentation of visually similar but irrelevant background elements. 
{{To address this, as shown in~\cref{fig:NBS}, we propose expanding the minimum bounding box in four directions—left, right, up, and down—based on their corresponding expansion coefficients $C_e$. These coefficients are designed to reflect the teacher model's confidence in the segmentation for each direction, with lower confidence resulting in greater expansion in that direction. This adaptive approach ensures that the expanded box captures potential false-negative regions while avoiding excessive enlargement, thus balancing inclusion and precision in the prompt generation process.
For instance, in the left direction, following the same strategy used in point extraction, we first divide the bounding box into nine equal blocks. Then, we calculate the left expansion coefficient $C_e^l$, which represents the proportion of pixels in the three leftmost blocks that have values other than 0 or 1, relative to the total number of pixels in those blocks. The coefficient $C_e^l$ is then used to enlarge the bounding box in the left direction by $C_e^l\,\%$ of the corresponding block's length. The same method is applied to the right, up, and down directions to obtain the fully expanded box. 
This approach ensures that the expanded box captures false-negative regions while avoiding excessive enlargement.}}

Finally, we extract the mask prompt $\mathbf{Y}_i^{m}$. In the later stages of training, as the teacher model increasingly identifies the concealed object but struggles with certain ambiguous regions, SAM requires refinement to eliminate these areas. The purpose of the mask prompt is to address false-positive regions predicted by the teacher model and prevent them from misleading SAM. Therefore, $\mathbf{Y}_i^{m}$ marks only the pixels with a value of 1 as belonging to the foreground object, while all other pixels are directly classified as background.


Our density-varied hybrid prompts work together effectively, adapting to the evolving characteristics of the mean-teacher model across different training stages. This approach enables thorough extraction of key information from the coarse mask and provides SAM with diverse, complementary guidance for producing more accurate pseudo-labels.

Once the multi-density prompts $\mathbf{Y}_i^{p}$ are obtained, they are input to SAM along with the augmented images $\{\mathbf{X}^k_i\}^K_{k=1}$ to generate segmentation masks. These masks are then fused to the final pseudo-label to enhance both accuracy and comprehensiveness, which is formulated as:
\begin{equation}
	\tilde{\mathbf{M}}_i = \frac{1}{K}\sum_{k=1}^K \mathrm{SAM}\left(\mathbf{X}^k_i, \mathbf{Y}_i^p \right),
	\label{Eq:RefinedPL}
\end{equation}
where $\tilde{\mathbf{M}}_i$ represents the pseudo-label of $\mathbf{X}_i$ in the current iteration. As illustrated in~\cref{fig:SWSCOS}, the overlap foreground regions in $\tilde{\mathbf{M}}_i$, consistently segmented by SAM across augmentations, 
are highly likely to accurately represent
the foreground object. This highlights the effectiveness of our fusion strategy in facilitating the generation of informative pseudo-labels.

}

\subsubsection{Pseudo-Label Storage}
{During network training, SAM generally produces progressively better pseudo-labels to facilitate network updates. However, we identified two key conditions: \textbf{(1)} While label quality typically improves over time, the quality of pseudo-labels in the $n^{th}$ epoch is not always superior to those in the $(n-1)^{th}$ epoch. \textbf{(2)} Pseudo-labels generated across different epochs vary due to differences in the extracted prompts. In some cases, these variations can complement one another, resulting in a more complete representation of the concealed object.

Given these insights, we propose establishing an optimal label pool $\mathcal{P}$ to dynamically store the top-$B$ best-ever pseudo-labels, ensuring that the student model receives accurate, diverse, and comprehensive supervision. In practice, during the $n^{th}$ epoch, the newly generated pseudo-label $\tilde{\mathbf{M}}_i^n$ is compared with the top-$B$ labels $\{\tilde{\mathbf{M}}_i^b\}_{b=1}^B$ in the pool. If $\tilde{\mathbf{M}}_i^n$ surpasses any of the labels in $\{\tilde{\mathbf{M}}_i^b\}_{b=1}^B$, it replaces that one. If $\tilde{\mathbf{M}}_i^n$ exceeds multiple labels, it randomly replaces one of them.

Building on the finding discussed at the end of~\cref{Sec:PLG}, which highlights that highly overlapped regions in pseudo-labels often correspond to correct predictions, this paper proposes evaluating pseudo-label quality from an informational perspective. We argue that an effective pseudo-label should be both \textbf{informative} and \textbf{helpful} to the student model. To evaluate these characteristics, we define an information entropy-based uncertainty map ${\mathbf{E}}(\bigcdot)$ with base 2. For instance, the uncertainty map for $\tilde{\mathbf{M}}_i^n$ is computed as:
\begin{equation}
	{\mathbf{E}}(\tilde{\mathbf{M}}_i^n) = -\tilde{\mathbf{M}}_i^n \log \tilde{\mathbf{M}}_i^n -(1-\tilde{\mathbf{M}}_i^n) \log (1-\tilde{\mathbf{M}}_i^n).
	\label{Eq:entropy}
\end{equation}

\cref{Eq:entropy} measures pixel-wise prediction uncertainty, where lower uncertainty values signify more confident and consistent predictions across all augmented images. By employing this uncertainty map, we assess the pseudo-label's informativeness—defined by the presence of regions with low uncertainty—and its usefulness to the student model, emphasizing areas where the model has not yet achieved high certainty. Specifically, the evaluation uses three metrics: absolute uncertainty $U_a$, relative uncertainty $U_r$, and differential uncertainty $U_d$. Lower values for 
each metric
indicate better performance. 

Absolute uncertainty $U_a$ measures the proportion of high-uncertainty pixels among all pixels, while relative uncertainty $U_r$ indicates the proportion of high-uncertainty pixels to foreground pixels with low uncertainty, designed to accommodate small-object scenarios. Both $U_a$ and $U_r$ directly assess the quality of the mask, determining its informativeness.
Besides, differential uncertainty $U_d$ evaluates how much the pseudo-label can aid the student model. This is calculated using the residual uncertainty map, ${\mathbf{E}}( |\tilde{\mathbf{M}}_i^n-\mathbf{(Y')}_i^{(n-1)}|)$, where $\tilde{\mathbf{M}}_i^n$ is the pseudo-label, and $\mathbf{(Y')}_i^{(n-1)}$ is the mask predicted by the student model in the previous epoch. The goal is to assess how much important information the pseudo-label contributes that the student model has not yet considered. 
It is worth noting that a pixel is deemed high-uncertainty if its uncertainty exceeds 0.9 when using base 2 in~\cref{Eq:entropy}.

If the newly generated pseudo-label $\tilde{\mathbf{M}}_i^n$ has lower values in at least two of the three metrics ($U_a$, $U_r$, and $U_d$) compared to any of the best-ever pseudo-labels $\{\tilde{\mathbf{M}}_i^b\}_{b=1}^B$, then $\tilde{\mathbf{M}}_i^n$ replaces one of the top-$B$ labels. Otherwise, the best-ever pseudo-labels remain unchanged. In the $n^{th}$ epoch, this process can be formulated as:
\begin{equation} \hspace{-3mm}
	\{\tilde{\mathbf{M}}_i^{b,n}\}_{b=1}^B = PLS_t\left(\{\tilde{\mathbf{M}}_i^{b,(n-1)}\}_{b=1}^B, \tilde{\mathbf{M}}_i^n, \mathbf{(Y')}_i^{(n-1)}\right),
 \label{Eq:pls}
\end{equation}
where $B$ is empirically set as 3 to balance efficiency and performance, although a larger $B$ can improve segmentation performance by providing more comprehensive pseudo-labels.

}
\begin{figure*}[t]
	\centering
	\setlength{\abovecaptionskip}{-0.2cm}
	\begin{center}
		\includegraphics[width=\linewidth]{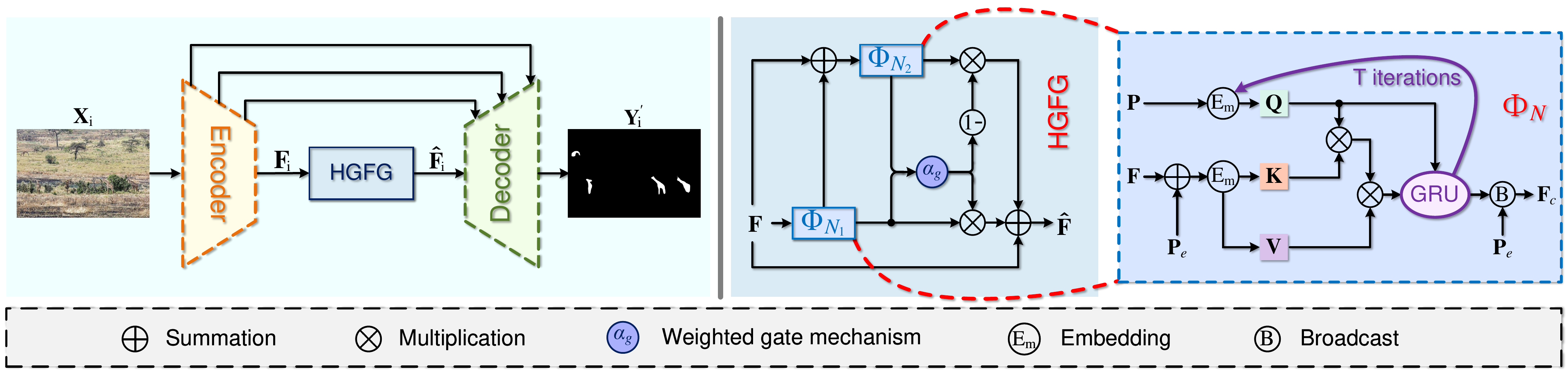}
	\end{center}
	\vspace{0.5mm}
	\caption{Details of HGFG. $\Phi_N$ is feature grouping with $N$ prototypes, whose 
 broadcast process is simplified for space limitation.}
	\label{fig:MFG}
	\vspace{-0.6cm}
\end{figure*}

\begin{figure}[t]
	\centering
	\setlength{\abovecaptionskip}{-0.2cm}
	\begin{center}
		\includegraphics[width=\linewidth]{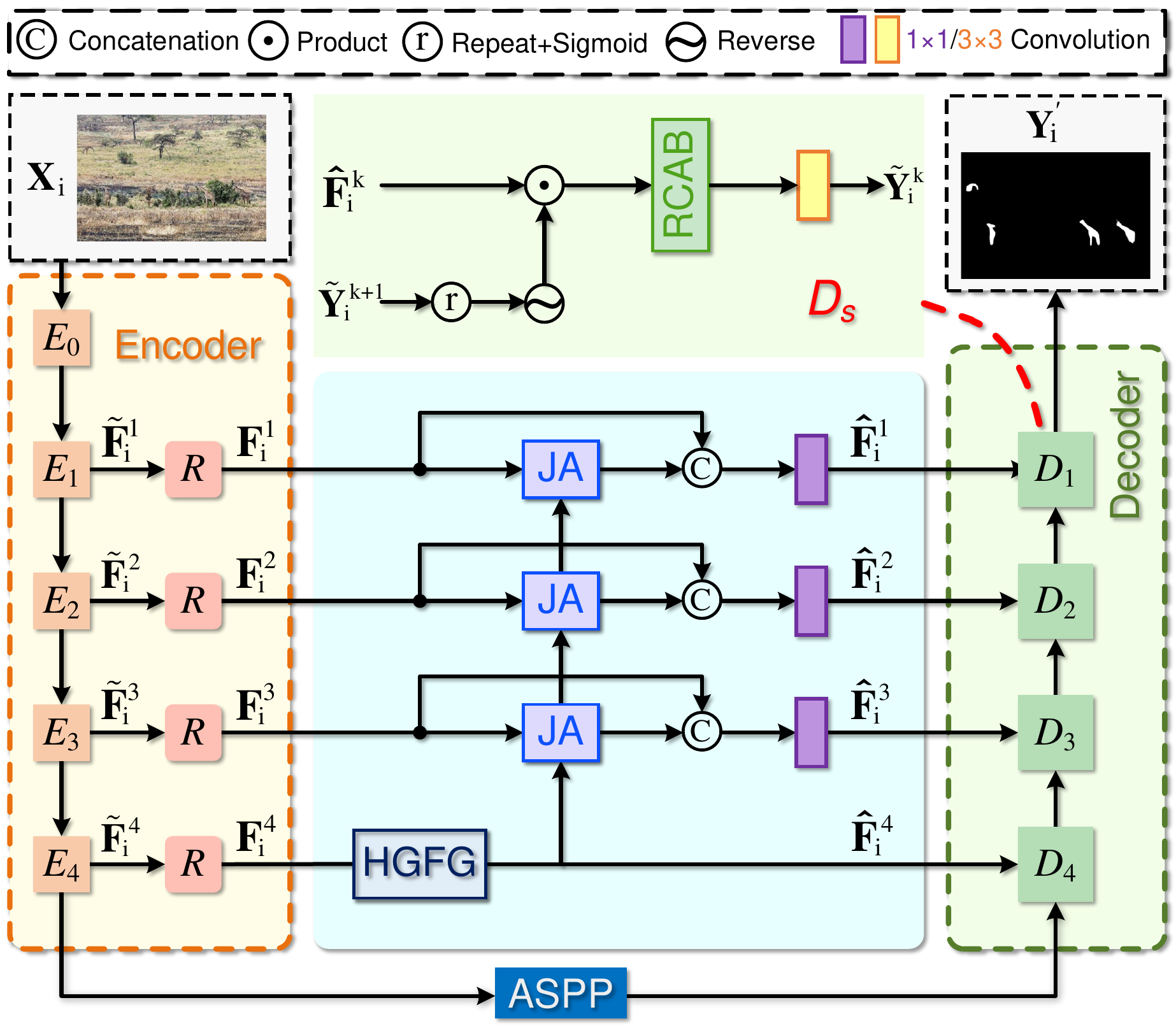}
	\end{center}
	\vspace{0.5mm}
	\caption{Framework of our segmenter with HGFG. Note that HGFG can be integrated with various existing methods.}
	\label{fig:Segmenter}
	\vspace{-0.1cm}
\end{figure}

\subsubsection{Pseudo-Label Supervision}
{ Given that discarding valuable yet confusing pseudo-labels could compromise model generalization, while using them for training might exacerbate confirmation bias due to inevitable mislabeling, we propose selecting reliable pseudo-labels and highlighting their reliable regions for supervision. To achieve this, we design an image-level selection and a pixel-level weighting strategy.

We have observed that for highly challenging concealed images, SAM frequently fails to generate reasonably accurate results, even when using multi-density prompts, regardless of the type of augmented images applied. To address this, we propose an image-level selection mechanism to identify informative pseudo-labels for training, intending to provide more reliable supervision. Specifically, to determine whether an image should be retained for training, we employ the absolute uncertainty $U_a$ and relative uncertainty $U_r$, where a pixel is regarded as a high-uncertainty pixel when its entropy is above 0.9. Given one of the top-$B$ pseudo-labels $\tilde{\mathbf{M}}_i^b$,  the indicator function $\mathbf{I}$, taking values of 0 or 1, is defined as: 
\begin{equation}
	\mathbf{I}(\tilde{\mathbf{M}}_i^b) = \mathbbm{1}[U_a(\tilde{\mathbf{M}}_i^b) < \tau_{a}] \times \mathbbm{1}[U_r(\tilde{\mathbf{M}}_i^b) < \tau_{r}],
\end{equation}
where $\tau_{a}$ and $\tau_{r}$ are the thresholds that are set as 0.1 and 0.5.

Having filtered out high-quality pseudo-labels, we propose using an entropy-based uncertainty map to weight the predictions, highlighting those more reliable regions, which are 
\begin{equation}
	\mathbf{E}(\tilde{\mathbf{M}}_i^b) = -\tilde{\mathbf{M}}_i^b \log \tilde{\mathbf{M}}_i^b -(1-\tilde{\mathbf{M}}_i^b) \log (1-\tilde{\mathbf{M}}_i^b). 
\end{equation}

By applying the uncertainty weights $\mathbf{E}(\tilde{\mathbf{M}}_i^b)$ to the image selection indicator $I(\tilde{\mathbf{M}}_i^b)$, we obtain the final weighted map $\hat{\mathbf{Y}}_i^b$ utilized for training the student segmenter:
\begin{equation}\label{Eq:PLSup}
	\hat{\mathbf{Y}}_i^b = (1 - \mathbf{E}(\tilde{\mathbf{M}}_i^b)) 
	\times \mathbf{I}(\tilde{\mathbf{M}}_i^b).
\end{equation}
Our method leverages SAM to generate pseudo-labels, stores the best-ever pseudo-labels, and selects the most reliable contents from them to supervise the segmenter, thereby ensuring robust network training.

}

{\subsubsection{Discussion}
Our SEE framework introduces the first unified solution for ISCOS tasks that operate orthogonally to existing methods by leveraging the capabilities of SAM. Inspired by SAM's remarkable performance on natural scene images, we extend its application to concealed object images by designing a comprehensive pipeline. This pipeline encompasses pseudo-label generation, storage, and supervision, addressing the challenge of providing the segmenter with consistent supervision in highly concealed scenarios. Our approach establishes a robust foundation for enhancing the segmentation performance.}

\subsection{Hybrid-Granularity Feature Grouping}
\label{Sec:MFG}
{The intrinsic similarity among concealed objects can significantly hinder effective segmentation, resulting in incomplete object localization in multi-object segmentation tasks. This problem is particularly pronounced in incompletely supervised scenarios, where the segmenter's limited discriminative capacity compounds the issue. To tackle this challenge, we propose a novel Hybrid-Granularity Feature Grouping (HGFG) module designed to extract discriminative cues across varying levels of granularity. HGFG operates by analyzing the coherence between foreground and background regions, allowing for feature grouping at multiple scales. By fostering feature coherence, HGFG mitigates incomplete segmentation by enhancing local correlations within individual objects and promotes comprehensive multi-object segmentation by seeking global coherence across all detected objects. The architecture of HGFG is shown in~\cref{fig:MFG}.

\noindent\textbf{Feature grouping}. 
Let $\mathbf{F}\in \mathbb{R}^{H\times W\times C}$ denote the feature of an input image. We perform feature grouping by mapping $\mathbf{F}$ to $N$ learnable cluster prototypes $\mathbf{P}\in \mathbb{R}^{N\times C}$, which are initialized randomly. Initially, we append a learnable spatial positional embedding $\mathbf{P}_e$ to the input feature $\mathbf{F}$, getting $\mathbf{F}_p$. Then, we linearly transform the prototypes $\mathbf{P}$ and the positioned feature $\mathbf{F}_p$ to obtain $\mathbf{Q}\in\mathbb{R}^{N\times C}$, $\mathbf{K}\in\mathbb{R}^{H W\times C}$, and $\mathbf{V}\in\mathbb{R}^{H W\times C}$:
\begin{equation}\label{eq:prototype_qkv}
	\mathbf{Q} = \mathbf{W}_q \mathbf{P}, \ \ \ \mathbf{K} = \mathbf{W}_k \mathbf{F}_p, \ \ \ \mathbf{V} = \mathbf{W}_v \mathbf{F}_p, 
\end{equation}
where $\mathbf{W}_q, \mathbf{W}_k, \mathbf{W}_v \in \mathbb{R}^{C\times C}$ are learnable weight matrices. 
To ensure the exclusive assignment of features to the cluster prototypes, we normalize the coefficients over all prototypes, 
\begin{equation}
	\bar{\mathbf{A}}_{i,j}=\frac{e^{\mathbf{A}_{i,j}}}{\sum_l e^{\mathbf{A}_{i,l}}}, \quad \mathrm{where} \quad \mathbf{A}=\frac{1}{\sqrt{C}} \mathbf{K} \mathbf{Q}^{\top}.
\end{equation}
We then calculate the integral value $\mathbf{U}$ of the input values concerning the prototypes as 
\begin{equation}
	\mathbf{U} = \mathbf{D}^{\top} \mathbf{V}, \quad \mathrm{where} \quad \mathbf{D}_{i,j}= \frac{\bar{\mathbf{A}}_{i,j}}{\sum_l \bar{\mathbf{A}}_{i,l}}, 
\end{equation}
and update the prototypes $\mathbf{P}$ by feeding it into a Gated Recurrent Units $GRU(\cdot)$ along with $\mathbf{U}$:
\begin{equation}\label{eq:prototype_GRU}
	\mathbf{P} = GRU\left(inputs=\mathbf{U}, states = \mathbf{P}\right).
\end{equation}
By repeating Eqs.~\eqref{eq:prototype_qkv} - \eqref{eq:prototype_GRU} for $T$ iterations, the cluster prototypes are iteratively updated, gradually strengthening the association between similar features, where $T=3$ here.

We broadcast each prototype onto a 2D grid augmented with the learnable spatial position embedding $\mathbf{P}_e$ to obtain $\{\mathbf{F}'_i\}_{i=1}^N \in \mathbb{R}^{H\times W\times C}$. Subsequently, we apply a $1\times1$ convolution to downsample each prototype, yielding $\{\mathbf{F}''_i\}_{i=1}^N \in \mathbb{R}^{H\times W\times C/N}$. We then concatenate these prototypes and obtain $\mathbf{F}_c\in \mathbb{R}^{H\times W\times C}$. 
We denote the feature grouping process with $N$ prototypes as $\mathbf{F}_c= \Phi_N(\mathbf{F})$ for convenience in references.}

{\noindent\textbf{Hybrid-granularity feature aggregation}. 
The number of prototypes $N$ in the above feature grouping technique controls the granularity of the grouping: a smaller $N$ facilitates the extraction of global information, while a larger $N$ yields more detailed information. To strike a balance, we propose aggregating hybrid-granularity grouping features with varying numbers of prototypes. Inspired by the second-order Runge-Kutta (RK2) structure, which is known for its superior numerical solutions compared to the traditional residual structure~\cite{he2016deep,he2023degradation}, we employ RK2 to aggregate these features. Additionally, as shown in Fig.~\ref{fig:MFG}, we adopt a weighted gate mechanism $\mathbf{\alpha}_g$ to adaptively estimate the trade-off parameter rather than using a fixed coefficient. Given the feature $\mathbf{F}$, the adaptively aggregated feature $\hat{\mathbf{F}}$ is formulated as follows:
\begin{equation}
	\hat{\mathbf{F}} = \mathbf{F} +\alpha_g \Phi_{N_1}(\mathbf{F}) +(1-\alpha_g) \Phi_{N_2}(\mathbf{F}+\Phi_{N_1}(\mathbf{F})), 
\end{equation}
where $\alpha_g=S(\sigma \ cat(\Phi_{N_1}(\mathbf{F}), \Phi_{N_2}(\mathbf{F}+\Phi_{N_1}(\mathbf{F})))+\mu)$. Here, $S$ denotes the Sigmoid function, while $\sigma$ and $\mu$ are the learnable parameters in $\alpha_g$. The values of $N_1$ and $N_2$, the numbers of groups, are empirically set to 2 and 4.

\noindent\textbf{Discussion}. Our HGFG technique is inspired by the slot attention method~\cite{locatello2020object}; however, we diverge from slot attention in several key aspects. While slot attention focuses on instance-level grouping in a self-supervised manner, our HGFG aims to adaptively extract feature-level coherence for complete segmentation and precise multi-object localization. To enhance the segmenter's flexibility and ensure generalization, we eliminate the auxiliary decoder used in slot attention for image reconstruction, along with the reconstruction constraint. Besides, we implement an RK2 structure to aggregate multiscale grouping features with varying numbers of prototypes, which further facilitates the extraction of feature coherence and enhances segmentation performance.}

\begin{table*}[ht]
	\centering
	\setlength{\abovecaptionskip}{0cm}
	\caption{Results on COD of the WSCOS task with point supervision and scribble supervision. 
		SCOD+ and SCOD++ indicate integrating SCOD with WS-SAM~\cite{he2023weaklysupervised} and our SEE framework, respectively. 
		The best two results are in {\color[HTML]{FF0000} \textbf{red}} and {\color[HTML]{00B0F0} \textbf{blue}} fonts.}
	\resizebox{2\columnwidth}{!}{
		\setlength{\tabcolsep}{1mm}
		\begin{tabular}{l|c|cccc|cccc|cccc|cccc}
			\toprule
			\multicolumn{1}{l|}{} & \multicolumn{1}{c|}{} & \multicolumn{4}{c|}{\textit{CHAMELEON} } & \multicolumn{4}{c|}{\textit{CAMO}} & \multicolumn{4}{c|}{\textit{COD10K}} & \multicolumn{4}{c}{\textit{NC4K}} \\ \cline{3-18}
			\multicolumn{1}{l|}{\multirow{-2}{*}{Methods}} & \multicolumn{1}{c|}{\multirow{-2}{*}{Pub.}} & {\cellcolor{gray!40}$M$~$\downarrow$} &{\cellcolor{gray!40}$F_\beta$~$\uparrow$} &{\cellcolor{gray!40}$E_\phi$~$\uparrow$} & \multicolumn{1}{c|}{\cellcolor{gray!40}$S_\alpha$~$\uparrow$}& {\cellcolor{gray!40}$M$~$\downarrow$} &{\cellcolor{gray!40}$F_\beta$~$\uparrow$} &{\cellcolor{gray!40}$E_\phi$~$\uparrow$} & \multicolumn{1}{c|}{\cellcolor{gray!40}$S_\alpha$~$\uparrow$}& {\cellcolor{gray!40}$M$~$\downarrow$} &{\cellcolor{gray!40}$F_\beta$~$\uparrow$} &{\cellcolor{gray!40}$E_\phi$~$\uparrow$} & \multicolumn{1}{c|}{\cellcolor{gray!40}$S_\alpha$~$\uparrow$}& {\cellcolor{gray!40}$M$~$\downarrow$} &{\cellcolor{gray!40}$F_\beta$~$\uparrow$} &{\cellcolor{gray!40}$E_\phi$~$\uparrow$} & \multicolumn{1}{c}{\cellcolor{gray!40}$S_\alpha$~$\uparrow$}\\ \midrule
			\multicolumn{18}{c}{Scribble Supervision} \\ \midrule
			SAM~\cite{kirillov2023segment}                                         & \multicolumn{1}{c|}{ICCV23}                              & 0.207                                 & 0.595                                 & 0.647                                 & 0.635                                 & 0.160                                 & 0.597                                 & 0.639                                 & 0.643                                 & 0.093                                 & 0.673                                 & 0.737                                 & 0.730                                 & 0.118                                 & 0.675                                 & 0.723                                 & 0.717                                 \\
		 $\text{SAM-W}_\text{S}$~\cite{chen2023sam}                                         & \multicolumn{1}{c|}{---}                              & 0.069                & 0.751                & 0.835                & 0.661                & 0.097                & 0.696                & 0.788                & 0.738                & 0.049                & 0.712                & 0.833                & 0.770                & 0.066                & 0.757                & 0.842                & 0.768                \\
			WSSA~\cite{zhang2020weakly}                                          & CVPR20                                             & 0.067                                 & 0.692                                 & 0.860                                 & 0.782                                 & 0.118                                 & 0.615                                 & 0.786                                 & 0.696                                 & 0.071                                 & 0.536                                 & 0.770                                 & 0.684                                 & 0.091                                 & 0.657                                 & 0.779                                 & 0.761                                 \\
			SCWS~\cite{yu2021structure}                                       & AAAI21                                             & 0.053                                 & 0.758                                 & 0.881                                 & 0.792                                 & 0.102                                 & 0.658                                 & 0.795                                 & 0.713                                 & 0.055                                 & 0.602                                 & 0.805                                 & 0.710                                 & 0.073                                 & 0.723                                 & 0.814                                 & 0.784                                 \\
			TEL~\cite{liang2022tree}                                           & CVPR22                                             & 0.073                                 & 0.708                                 & 0.827                                 & 0.785                                 & 0.104                                 & 0.681                                 & 0.797                                 & 0.717                                 & 0.057                                 & 0.633                                 & 0.826                                 & 0.724                                 & 0.075                                 & 0.754                                 & 0.832                                 & 0.782                                 \\
			SCOD~\cite{he2022weakly}                                         & AAAI23                                             & 0.046                                 & 0.791                                 & 0.897                                 & 0.818                                 & 0.092                                 & 0.709                                 & 0.815                                 & 0.735                                 & 0.049                                 & 0.637                                 & 0.832                                 & 0.733                                 & 0.064                                 & 0.751                                 & 0.853                                 & 0.779                                 \\
      GenSAM~\cite{hu2024relax}   & AAAI24 & 0.090 & 0.680 & 0.807 & 0.764 & 0.113 & 0.659 & 0.775 & 0.719 & 0.067 & 0.681 & 0.838 & 0.775 & 0.097 & 0.687 & 0.750 & 0.732 \\
   SCOD+~\cite{he2023weaklysupervised} & NIPS23 & 0.046                                 & {\color[HTML]{00B0F0} \textbf{0.797}} & 0.900                                 & 0.820                                 & {\color[HTML]{00B0F0} \textbf{0.090}} & 0.716                                 & 0.818                                 & 0.741                                 & 0.047                                 & 0.650                                 & 0.845                                 & 0.742                                 & 0.060                                 & 0.766                                 & 0.862                                 & 0.785                                 \\
{GenSAM+~\cite{he2023weaklysupervised}}  & NIPS23 & 0.087 & 0.688 & 0.814 & 0.766 & 0.110 & 0.667 & 0.787 & 0.723 & 0.063 & 0.690 & 0.841 & 0.810 & 0.085 & 0.695 & 0.763 & 0.736 \\
WS-SAM~\cite{he2023weaklysupervised} & NIPS23 & 0.046                                 & 0.777                                 & 0.897                                 & {\color[HTML]{00B0F0} \textbf{0.824}} & 0.092                                 & {\color[HTML]{00B0F0} \textbf{0.742}} & 0.818                                 & {\color[HTML]{00B0F0} \textbf{0.759}} & {\color[HTML]{00B0F0} \textbf{0.038}} & {\color[HTML]{00B0F0} \textbf{0.719}} & {\color[HTML]{00B0F0} \textbf{0.878}} & {\color[HTML]{00B0F0} \textbf{0.803}} & {\color[HTML]{00B0F0} \textbf{0.052}} & {\color[HTML]{00B0F0} \textbf{0.802}} & {\color[HTML]{00B0F0} \textbf{0.886}} & {\color[HTML]{00B0F0} \textbf{0.829}} \\ 
   \rowcolor{c2!20}SCOD++ & \multicolumn{1}{c|}{---}& {\color[HTML]{00B0F0} \textbf{0.045}} & {\color[HTML]{FF0000} \textbf{0.802}} & {\color[HTML]{FF0000} \textbf{0.904}} & 0.823                                 & {\color[HTML]{FF0000} \textbf{0.089}} & 0.722                                 & {\color[HTML]{00B0F0} \textbf{0.823}} & 0.744                                 & 0.044                                 & 0.673                                 & 0.858                                 & 0.751                                 & 0.058                                 & 0.776                                 & 0.868                                 & 0.797                                 \\
   \rowcolor{c2!20}{GenSAM++} & \multicolumn{1}{c|}{---} & 0.083 & 0.695 & 0.826 & 0.771 & 0.105 & 0.675 & 0.799 & 0.731 & 0.057 & 0.702 & 0.853 & 0.818 & 0.072 & 0.715 & 0.782 & 0.742 \\
   \rowcolor{c2!20} SEE (Ours) & \multicolumn{1}{c|}{---}& {\color[HTML]{FF0000} \textbf{0.044}} & 0.785                                 & {\color[HTML]{00B0F0} \textbf{0.903}} & {\color[HTML]{FF0000} \textbf{0.826}} & {\color[HTML]{00B0F0} \textbf{0.090}} & {\color[HTML]{FF0000} \textbf{0.747}} & {\color[HTML]{FF0000} \textbf{0.826}} & {\color[HTML]{FF0000} \textbf{0.765}} & {\color[HTML]{FF0000} \textbf{0.036}} & {\color[HTML]{FF0000} \textbf{0.729}} & {\color[HTML]{FF0000} \textbf{0.883}} & {\color[HTML]{FF0000} \textbf{0.807}} & {\color[HTML]{FF0000} \textbf{0.051}} & {\color[HTML]{FF0000} \textbf{0.808}} & {\color[HTML]{FF0000} \textbf{0.891}} & {\color[HTML]{FF0000} \textbf{0.836}} \\
   \midrule
			\multicolumn{18}{c}{Point Supervision} \\ \midrule
			SAM~\cite{kirillov2023segment} & \multicolumn{1}{c|}{ICCV23} & 0.207 & 0.595                                 & 0.647                                 & 0.635                                 & 0.160                                 & 0.597                                 & 0.639                                 & 0.643                                 & 0.093                                 & 0.673                                 & 0.737                                 & 0.730                                 & 0.118                                 & 0.675                                 & 0.723                                 & 0.717                                 \\
		$\text{SAM-W}_\text{P}$~\cite{chen2023sam} & \multicolumn{1}{c|}{---} & 0.095 & 0.708 & 0.752 & 0.688 & 0.117 & 0.653 & 0.698 & 0.681 & 0.068 & 0.697 & 0.805 & 0.767 & 0.078 & 0.736 & 0.793 & 0.782 \\
			WSSA~\cite{zhang2020weakly}                                          & CVPR20                                             & 0.105                                 & 0.660                                 & 0.712                                 & 0.711                                 & 0.148                                 & 0.607                                 & 0.652                                 & 0.649                                 & 0.087                                 & 0.509                                 & 0.733                                 & 0.642                                 & 0.104                                 & 0.688                                 & 0.756                                 & 0.743                                 \\
			SCWS~\cite{yu2021structure}                                       & AAAI21                                             & 0.097                                 & 0.684                                 & 0.739                                 & 0.714                                 & 0.142                                 & 0.624                                 & 0.672                                 & {{0.687}}                                 & 0.082                                 & 0.593                                 & 0.777                                 & 0.738                                 & 0.098                                 & 0.695                                 & 0.767                                 & 0.754                                 \\
			TEL~\cite{liang2022tree}                                           & CVPR22                                             & 0.094                                 & {{0.712}}                                 & 0.751                                 & {{0.746}}                                 & 0.133                                 & {{0.662}}                                 & 0.674                                 & 0.645                                 & 0.063                                 & 0.623                                 & 0.803                                 & 0.727                                 & 0.085                                 & 0.725                                 & 0.795                                 & 0.766                                 \\
			SCOD~\cite{he2022weakly}                                         & AAAI23                                             & 0.092 & 0.688                                 & 0.746                                 & 0.725                                 & 0.137                                 & 0.629                                 & 0.688                                 & 0.663                                 & 0.060                                 & 0.607                                 & 0.802                                 & 0.711                                 & 0.080                                 & 0.744                                 & 0.796                                 & 0.758                                 \\
   GenSAM~\cite{hu2024relax}   &   AAAI24    & 0.090 & 0.680 & 0.807 & 0.764 & 0.113 & 0.659 & 0.775 & 0.719 & 0.067 & 0.681 & 0.838 & 0.775 & 0.097 & 0.687 & 0.750 & 0.732 \\
	SCOD+~\cite{he2023weaklysupervised} & NIPS23& 0.089                                 & 0.704                                 & 0.757                                 & 0.731                                 & 0.129                                 & 0.642                                 & 0.693                                 & 0.666                                 & 0.058                                 & 0.618                                 & 0.812                                 & 0.719                                 & 0.075                                 & 0.767                                 & 0.825                                 & 0.771                                 \\
 GenSAM+~\cite{he2023weaklysupervised}   &   NIPS23  & 0.089 & 0.676 & 0.810 & 0.763 & 0.108 & 0.662 & 0.777 & 0.720 & 0.066 & 0.686 & 0.843 & 0.777 & 0.095 & 0.702 & 0.764 & 0.738 \\
		WS-SAM~\cite{he2023weaklysupervised}  & NIPS23 & {\color[HTML]{00B0F0} \textbf{0.056}} & {\color[HTML]{00B0F0} \textbf{0.767}} & {\color[HTML]{00B0F0} \textbf{0.868}} & {\color[HTML]{00B0F0} \textbf{0.805}} & {\color[HTML]{00B0F0} \textbf{0.102}} & {\color[HTML]{00B0F0} \textbf{0.703}} & {\color[HTML]{00B0F0} \textbf{0.757}} & {\color[HTML]{00B0F0} \textbf{0.718}} & {\color[HTML]{00B0F0} \textbf{0.039}} & {\color[HTML]{00B0F0} \textbf{0.698}} & {\color[HTML]{00B0F0} \textbf{0.856}} & {\color[HTML]{00B0F0} \textbf{0.790}} & {\color[HTML]{00B0F0} \textbf{0.057}} & {\color[HTML]{00B0F0} \textbf{0.801}} & {\color[HTML]{00B0F0} \textbf{0.859}} & {\color[HTML]{00B0F0} \textbf{0.813}} \\ 
      \rowcolor{c2!20}SCOD++ & \multicolumn{1}{c|}{---}& 0.086                                 & 0.722                                 & 0.766                                 & 0.735                                 & 0.122                                 & 0.667                                 & 0.714                                 & 0.676                                 & 0.055                                 & 0.637                                 & 0.823                                 & 0.726                                 & 0.071                                 & 0.783                                 & 0.836                                 & 0.780                                 \\
       \rowcolor{c2!20} GenSAM++ &  \multicolumn{1}{c|}{---} & 0.087 & 0.689 & 0.815 & 0.768 & 0.106 & 0.666 & 0.779 & 0.719 & 0.063 & 0.692 & 0.850 & 0.778 & 0.090 & 0.723 & 0.780 & 0.743 \\
   \rowcolor{c2!20} SEE (Ours)& \multicolumn{1}{c|}{---}& {\color[HTML]{FF0000} \textbf{0.055}} & {\color[HTML]{FF0000} \textbf{0.772}} & {\color[HTML]{FF0000} \textbf{0.872}} & {\color[HTML]{FF0000} \textbf{0.806}} & {\color[HTML]{FF0000} \textbf{0.098}} & {\color[HTML]{FF0000} \textbf{0.712}} & {\color[HTML]{FF0000} \textbf{0.769}} & {\color[HTML]{FF0000} \textbf{0.721}} & {\color[HTML]{FF0000} \textbf{0.038}} & {\color[HTML]{FF0000} \textbf{0.706}} & {\color[HTML]{FF0000} \textbf{0.862}} & {\color[HTML]{FF0000} \textbf{0.796}} & {\color[HTML]{FF0000} \textbf{0.055}} & {\color[HTML]{FF0000} \textbf{0.806}} & {\color[HTML]{FF0000} \textbf{0.867}} & {\color[HTML]{FF0000} \textbf{0.817}}\\
			\bottomrule
	\end{tabular}}
	\label{table:CODQuanti}
	\vspace{-0.2cm}
\end{table*}
\begin{figure*}[t]
	\centering
	\setlength{\abovecaptionskip}{-0cm}
	\begin{center}
		\includegraphics[width=0.85\linewidth]{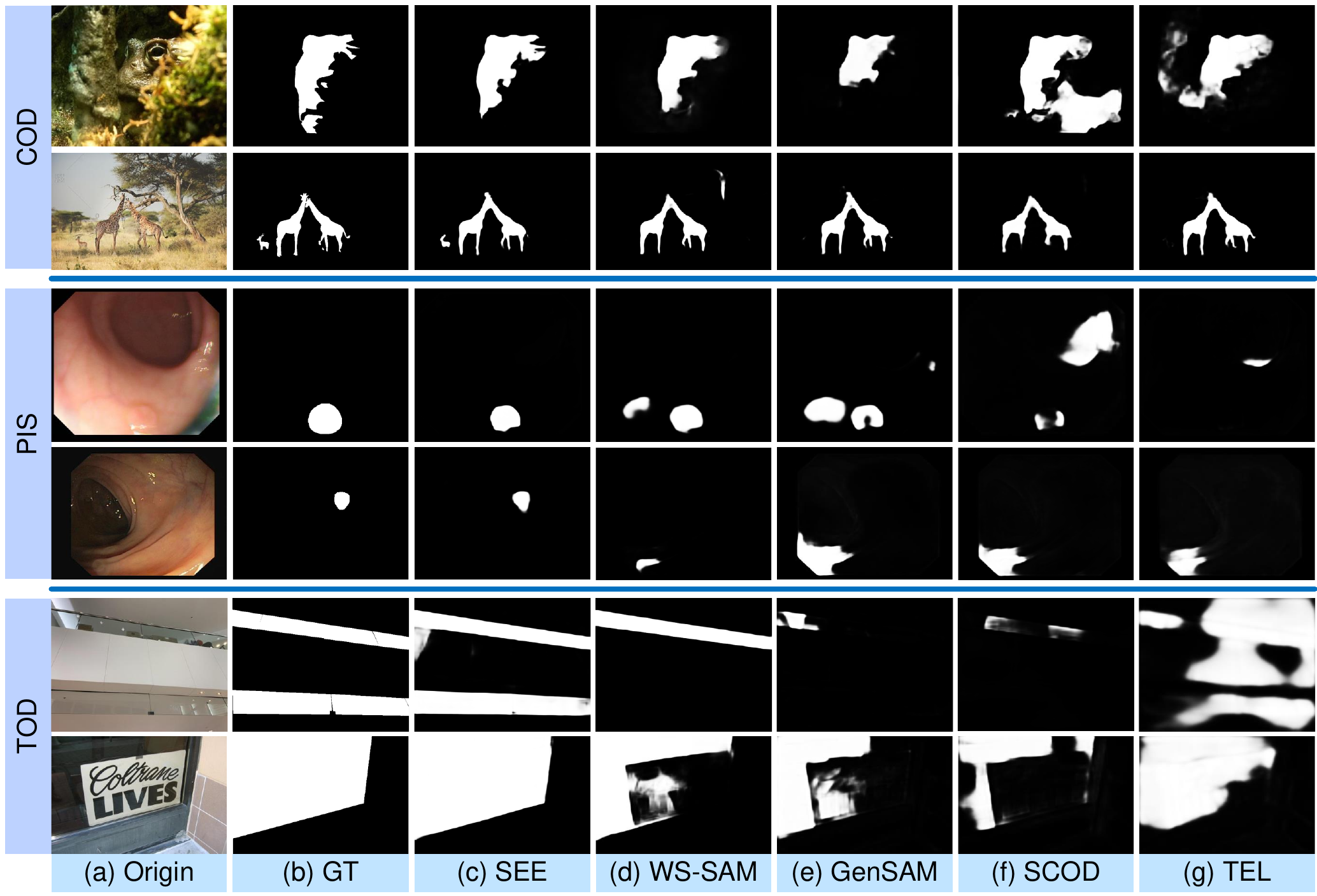}
	\end{center}
	\caption{Visualized results for COS tasks with point supervision, including COD, PIS, and TOD.}
	\label{fig:COSQuali}
	\vspace{-0.45cm}
\end{figure*}
\begin{table*}[t]
	\centering
	\setlength{\abovecaptionskip}{0cm}
	\caption{Results on PIS and TOD of the WSCOS task with point supervision. 
	}
	\resizebox{2\columnwidth}{!}{
		\setlength{\tabcolsep}{0.8mm}
		\begin{tabular}{l|cccc|cccc|cccc|cccc|cccc}
			\toprule
			& \multicolumn{12}{c|}{Polyp Image Segmentation (PIS)}& \multicolumn{8}{c}{Transparant Object Detection (TOD)}\\ \cline{2-21}
			& \multicolumn{4}{c|}{\textit{CVC-ColonDB}}& \multicolumn{4}{c|}{\textit{ETIS}}& \multicolumn{4}{c|}{\textit{Kvasir}}& \multicolumn{4}{c|}{\textit{GDD}}& \multicolumn{4}{c}{\textit{GSD}}\\ \cline{2-21}
			\multirow{-3}{*}{Methods} & {\cellcolor{gray!40}$M$~$\downarrow$} &{\cellcolor{gray!40}$F_\beta$~$\uparrow$} &{\cellcolor{gray!40}$E_\phi$~$\uparrow$} & \multicolumn{1}{c|}{\cellcolor{gray!40}$S_\alpha$~$\uparrow$}& {\cellcolor{gray!40}$M$~$\downarrow$} &{\cellcolor{gray!40}$F_\beta$~$\uparrow$} &{\cellcolor{gray!40}$E_\phi$~$\uparrow$} & \multicolumn{1}{c|}{\cellcolor{gray!40}$S_\alpha$~$\uparrow$}& {\cellcolor{gray!40}$M$~$\downarrow$} &{\cellcolor{gray!40}$F_\beta$~$\uparrow$} &{\cellcolor{gray!40}$E_\phi$~$\uparrow$} & \multicolumn{1}{c|}{\cellcolor{gray!40}$S_\alpha$~$\uparrow$}& {\cellcolor{gray!40}$M$~$\downarrow$} &{\cellcolor{gray!40}$F_\beta$~$\uparrow$} &{\cellcolor{gray!40}$E_\phi$~$\uparrow$} & \multicolumn{1}{c|}{\cellcolor{gray!40}$S_\alpha$~$\uparrow$}& {\cellcolor{gray!40}$M$~$\downarrow$} &{\cellcolor{gray!40}$F_\beta$~$\uparrow$} &{\cellcolor{gray!40}$E_\phi$~$\uparrow$} & \multicolumn{1}{c}{\cellcolor{gray!40}$S_\alpha$~$\uparrow$}\\ \midrule
			SAM~\cite{kirillov2023segment}& 0.479                                 & 0.343                                 & 0.419                                 & 0.427                                 & 0.429                                 & 0.439                                 & 0.512                                 & 0.503                                 & 0.320                                 & 0.545                                 & 0.564                                 & 0.582                                 & 0.245                                 & 0.512                                 & 0.530                                 & 0.551                                 & 0.266                                 & 0.473                                 & 0.501                                 & 0.514                                 \\
			$\text{SAM-W}_\text{P}$~\cite{chen2023sam}& 0.146 & 0.612 & 0.690 & 0.683 & 0.125 & 0.650 & 0.731 & 0.728 & 0.093 & 0.818 & 0.823 & 0.815 & 0.143 & 0.691 & 0.733 & 0.636 & 0.137 & 0.715 & 0.752 & 0.681 \\
			WSSA~\cite{zhang2020weakly} & 0.127                                 & 0.645                                 & 0.732                                 & 0.713                                 & 0.123                                 & 0.647                                 & 0.733                                 & 0.762                                 & 0.082                                 & 0.822                                 & 0.852                                 & 0.828                                 & 0.173                                 & 0.652                                 & 0.710                                 & 0.616                                 & 0.185                                 & 0.661                                 & 0.712                                 & 0.650                                 \\
			SCWS~\cite{yu2021structure}                      & 0.082                                 & 0.674                                 & 0.758                                 & 0.787                                 & 0.085                                 & 0.646                                 & 0.768                                 & 0.731                                 & 0.078                                 & 0.837                                 & 0.860                                 & 0.831                                 & 0.170                                 & 0.631                                 & 0.702                                 & 0.613                                 & 0.172                                 & 0.706                                 & 0.738                                 & 0.673                                 \\
			TEL~\cite{liang2022tree}                       & 0.089                                 & 0.669                                 & 0.743                                 & 0.761                                 & 0.083                                 & 0.639                                 & 0.776                                 & 0.726                                 & 0.091                                 & 0.810                                 & 0.826                                 & 0.804                                 & 0.230                                 & 0.640                                 & 0.586                                 & 0.536                                 & 0.275                                 & 0.571                                 & 0.501                                 & 0.495                                 \\
			SCOD~\cite{he2022weakly}                      & 0.077                                 & 0.691                                 & 0.795                                 & 0.802                                 & 0.071                                 & 0.664                                 & 0.802                                 & 0.766                                 & 0.071                                 & 0.853                                 & 0.877                                 & {{0.836}} & 0.146                                 & 0.801                                 & 0.778                                 & 0.723                                 & 0.154                                 & 0.743                                 & 0.751                                 & 0.710                                 \\
   GenSAM~\cite{hu2024relax}& 0.079 & 0.665 & 0.739 & 0.793 & 0.073 & 0.665 & 0.813 & 0.772 & 0.073 & 0.841 & 0.866 & 0.829 & 0.115 & 0.712 & 0.795 & 0.737 & 0.127 & 0.757 & 0.760 & 0.724 \\
	SCOD+~\cite{he2023weaklysupervised}              & 0.074                                 & 0.702                                 & 0.806                                 & 0.803                                 & 0.066                                 & 0.670                                 & 0.811                                 & 0.769                                 & 0.068                                 & 0.860                                 & 0.880                                 & 0.836                                 & 0.129                                 & 0.818                                 & 0.796                                 & 0.732                                 & 0.145                                 & 0.761                                 & 0.765                                 & 0.720                                 \\
 GenSAM+~\cite{he2023weaklysupervised}   & 0.076 & 0.682 & 0.752 & 0.797 & 0.072 & 0.673 & 0.819 & 0.776 & 0.072 & 0.846 & 0.868 & 0.833 & 0.112 & 0.719 & 0.802 & 0.740 & 0.126 & 0.763 & 0.763 & 0.727 \\
		WS-SAM~\cite{he2023weaklysupervised}                & {\color[HTML]{00B0F0} \textbf{0.043}} & {\color[HTML]{00B0F0} \textbf{0.721}} & {\color[HTML]{00B0F0} \textbf{0.839}} & {\color[HTML]{00B0F0} \textbf{0.816}} & {\color[HTML]{00B0F0} \textbf{0.037}} & {\color[HTML]{00B0F0} \textbf{0.694}} & {\color[HTML]{00B0F0} \textbf{0.849}} & {\color[HTML]{00B0F0} \textbf{0.797}} & {\color[HTML]{00B0F0} \textbf{0.046}} & {\color[HTML]{00B0F0} \textbf{0.878}} & {\color[HTML]{00B0F0} \textbf{0.917}} & {\color[HTML]{00B0F0} \textbf{0.877}} & {\color[HTML]{00B0F0} \textbf{0.078}} & {\color[HTML]{00B0F0} \textbf{0.858}} & {\color[HTML]{00B0F0} \textbf{0.863}} & {\color[HTML]{00B0F0} \textbf{0.775}} & {\color[HTML]{00B0F0} \textbf{0.089}} & {\color[HTML]{00B0F0} \textbf{0.839}} & {\color[HTML]{00B0F0} \textbf{0.841}} & {\color[HTML]{00B0F0} \textbf{0.764}} \\
  		\rowcolor{c2!20}SCOD++ & 0.070                                 & 0.712                                 & 0.818                                 & 0.807                                 & 0.064                                 & 0.682                                 & 0.819                                 & 0.773                                 & 0.066                                 & 0.871                                 & 0.890                                 & 0.841                                 & 0.112                                 & 0.830                                 & 0.812                                 & 0.740                                 & 0.122                                 & 0.776                                 & 0.774                                 & 0.729                                 \\
    \rowcolor{c2!20} GenSAM++ & 0.070 & 0.707 & 0.773 & 0.802 & 0.068 & 0.681 & 0.824 & 0.785 & 0.069 & 0.855 & 0.873 & 0.838 & 0.107 & 0.727 & 0.819 & 0.744 & 0.122 & 0.768 & 0.775 & 0.731 \\
    		\rowcolor{c2!20} SEE (Ours) & {\color[HTML]{FF0000} \textbf{0.042}} & {\color[HTML]{FF0000} \textbf{0.726}} & {\color[HTML]{FF0000} \textbf{0.842}} & {\color[HTML]{FF0000} \textbf{0.819}} & {\color[HTML]{FF0000} \textbf{0.035}} & {\color[HTML]{FF0000} \textbf{0.705}} & {\color[HTML]{FF0000} \textbf{0.857}} & {\color[HTML]{FF0000} \textbf{0.800}} & {\color[HTML]{FF0000} \textbf{0.045}} & {\color[HTML]{FF0000} \textbf{0.883}} & {\color[HTML]{FF0000} \textbf{0.918}} & {\color[HTML]{FF0000} \textbf{0.879}} & {\color[HTML]{FF0000} \textbf{0.073}} & {\color[HTML]{FF0000} \textbf{0.867}} & {\color[HTML]{FF0000} \textbf{0.871}} & {\color[HTML]{FF0000} \textbf{0.777}} & {\color[HTML]{FF0000} \textbf{0.081}} & {\color[HTML]{FF0000} \textbf{0.846}} & {\color[HTML]{FF0000} \textbf{0.849}} & {\color[HTML]{FF0000} \textbf{0.768}} \\
  \bottomrule
	\end{tabular}} \label{table:MISTOD_Quanti}
	\vspace{-0.1cm}
\end{table*}

\begin{table*}[ht]
	\centering
	\setlength{\abovecaptionskip}{0cm}
	\caption{Results on COD of the SSCOS task with 1/8 and 1/16 labeled training data. 
  }
	\resizebox{2\columnwidth}{!}{
		\setlength{\tabcolsep}{1mm}
		\begin{tabular}{l|c|cccc|cccc|cccc|cccc}
			\toprule
			\multicolumn{1}{l|}{} & \multicolumn{1}{c|}{} & \multicolumn{4}{c|}{\textit{CHAMELEON} } & \multicolumn{4}{c|}{\textit{CAMO}} & \multicolumn{4}{c|}{\textit{COD10K}} & \multicolumn{4}{c}{\textit{NC4K}} \\ \cline{3-18}
			\multicolumn{1}{l|}{\multirow{-2}{*}{Methods}} & \multicolumn{1}{c|}{\multirow{-2}{*}{Pub.}} & {\cellcolor{gray!40}$M$~$\downarrow$} &{\cellcolor{gray!40}$F_\beta$~$\uparrow$} &{\cellcolor{gray!40}$E_\phi$~$\uparrow$} & \multicolumn{1}{c|}{\cellcolor{gray!40}$S_\alpha$~$\uparrow$}& {\cellcolor{gray!40}$M$~$\downarrow$} &{\cellcolor{gray!40}$F_\beta$~$\uparrow$} &{\cellcolor{gray!40}$E_\phi$~$\uparrow$} & \multicolumn{1}{c|}{\cellcolor{gray!40}$S_\alpha$~$\uparrow$}& {\cellcolor{gray!40}$M$~$\downarrow$} &{\cellcolor{gray!40}$F_\beta$~$\uparrow$} &{\cellcolor{gray!40}$E_\phi$~$\uparrow$} & \multicolumn{1}{c|}{\cellcolor{gray!40}$S_\alpha$~$\uparrow$}& {\cellcolor{gray!40}$M$~$\downarrow$} &{\cellcolor{gray!40}$F_\beta$~$\uparrow$} &{\cellcolor{gray!40}$E_\phi$~$\uparrow$} & \multicolumn{1}{c}{\cellcolor{gray!40}$S_\alpha$~$\uparrow$}\\ \midrule
			\multicolumn{18}{c}{1/8 Labeled Training Data} \\ \midrule
			SAM~\cite{kirillov2023segment}        & ICCV23       & 0.207                                 & 0.595                                 & 0.647                                 & 0.635                                 & 0.160                                 & 0.597                                 & 0.639                                 & 0.643                                 & 0.093                                 & 0.673                                 & 0.737                                 & 0.730                                 & 0.118                                 & 0.675                                 & 0.723                                 & 0.717                                 \\
SAM-S~\cite{chen2023sam}      &  ---    & 0.136                                 & 0.667                                 & 0.695                                 & 0.672                                 & 0.133                                 & 0.637                                 & 0.678                                 & 0.662                                 & 0.073                                 & {\color[HTML]{00B0F0} \textbf{0.695}} & 0.770                                 & 0.751                                 & 0.085                                 & 0.715                                 & 0.756                                 & 0.767                                 \\
DTEN~\cite{ren2023towards}       & CVPR23 & 0.062                                 & 0.775                                 & 0.862                                 & 0.803                                 & 0.103                                 & 0.688                                 & 0.781                                 & 0.742                                 & 0.054                                 & 0.635                                 & 0.791                                 & 0.747                                 & 0.070                                 & 0.733                                 & 0.833                                 & 0.790                                 \\
PGCL~\cite{basak2023pseudo}       & CVPR23  & 0.051                                 & 0.792                                 & 0.878                                 & 0.833                                 & 0.096                                 & 0.705                                 & 0.803                                 & 0.755                                 & 0.051                                 & 0.658                                 & 0.798                                 & 0.752                                 & 0.063                                 & 0.753                                 & 0.838                                 & 0.803                                 \\
EPS~\cite{lee2023saliency}        & TPAMI23 & 0.042                                 & 0.810                                 & 0.891                                 & 0.862                                 & 0.090                                 & 0.721                                 & 0.815                                 & 0.753                                 & 0.047                                 & 0.659                                 & 0.806                                 & 0.761                                 & 0.058                                 & 0.765                                 & 0.855                                 & 0.809                                 \\
CoSOD~\cite{chakraborty2024unsupervised}      & WACV24  & 0.047                                 & 0.802                                 & 0.883                                 & 0.850                                 & 0.092                                 & 0.730                                 & 0.822                                 & 0.761                                 & 0.046                                 & 0.673                                 & 0.813                                 & 0.767                                 & 0.061                                 & 0.764                                 & 0.862                                 & 0.818                                 \\
\rowcolor{c2!20} DTEN++     &   ---   & 0.060                                 & 0.786                                 & 0.870                                 & 0.809                                 & 0.098                                 & 0.695                                 & 0.792                                 & 0.750                                 & 0.051                                 & 0.653                                 & 0.808                                 & 0.754                                 & 0.067                                 & 0.743                                 & 0.842                                 & 0.795                                 \\
\rowcolor{c2!20} PGCL++     &   ---      & 0.048                                 & 0.805                                 & 0.898                                 & 0.839                                 & 0.093                                 & 0.712                                 & 0.810                                 & 0.760                                 & 0.049                                 & 0.663                                 & 0.804                                 & 0.754                                 & 0.062                                 & 0.756                                 & 0.840                                 & 0.807                                 \\
\rowcolor{c2!20} EPS++      &   ---      & {\color[HTML]{00B0F0} \textbf{0.039}} & {\color[HTML]{00B0F0} \textbf{0.818}} & {\color[HTML]{00B0F0} \textbf{0.900}} & {\color[HTML]{00B0F0} \textbf{0.866}} & {\color[HTML]{00B0F0} \textbf{0.087}} & 0.729                                 & 0.826                                 & 0.757                                 & 0.044                                 & 0.672                                 & 0.819                                 & 0.766                                 & {\color[HTML]{00B0F0} \textbf{0.055}} & 0.772                                 & 0.866                                 & 0.815                                 \\
\rowcolor{c2!20} CoSOD++    &   ---      & 0.043                                 & 0.815                                 & 0.895                                 & 0.858                                 & 0.088                                 & {\color[HTML]{00B0F0} \textbf{0.746}} & {\color[HTML]{00B0F0} \textbf{0.837}} & {\color[HTML]{00B0F0} \textbf{0.769}} & {\color[HTML]{00B0F0} \textbf{0.043}} & 0.686                                 & {\color[HTML]{00B0F0} \textbf{0.825}} & {\color[HTML]{00B0F0} \textbf{0.771}} & 0.057                                 & {\color[HTML]{00B0F0} \textbf{0.774}} & {\color[HTML]{00B0F0} \textbf{0.875}} & {\color[HTML]{00B0F0} \textbf{0.827}} \\
\rowcolor{c2!20} SEE (Ours) &   ---      & {\color[HTML]{FF0000} \textbf{0.035}} & {\color[HTML]{FF0000} \textbf{0.825}} & {\color[HTML]{FF0000} \textbf{0.903}} & {\color[HTML]{FF0000} \textbf{0.873}} & {\color[HTML]{FF0000} \textbf{0.083}} & {\color[HTML]{FF0000} \textbf{0.753}} & {\color[HTML]{FF0000} \textbf{0.843}} & {\color[HTML]{FF0000} \textbf{0.776}} & {\color[HTML]{FF0000} \textbf{0.040}} & {\color[HTML]{FF0000} \textbf{0.703}} & {\color[HTML]{FF0000} \textbf{0.839}} & {\color[HTML]{FF0000} \textbf{0.786}} & {\color[HTML]{FF0000} \textbf{0.053}} & {\color[HTML]{FF0000} \textbf{0.778}} & {\color[HTML]{FF0000} \textbf{0.889}} & {\color[HTML]{FF0000} \textbf{0.839}} \\
   \midrule
			\multicolumn{18}{c}{1/16 Labeled Training Data} \\ \midrule
			SAM~\cite{kirillov2023segment}        & ICCV23 & 0.207                                 & 0.595                                 & 0.647                                 & 0.635                                 & 0.160                                 & 0.597                                 & 0.639                                 & 0.643                                 & 0.093                                 & 0.673                                 & 0.737                                 & 0.730                                 & 0.118                                 & 0.675                                 & 0.723                                 & 0.717                                 \\
SAM-S~\cite{chen2023sam}      &     ---        & 0.150                                 & 0.642                                 & 0.680                                 & 0.657                                 & 0.146                                 & 0.624                                 & 0.663                                 & 0.647                                 & 0.078                                 & {\color[HTML]{FF0000} \textbf{0.682}} & 0.752                                 & 0.738                                 & 0.093                                 & 0.692                                 & 0.741                                 & 0.753                                 \\
DTEN~\cite{ren2023towards}       & CVPR23               & 0.070                                 & 0.731                                 & 0.827                                 & 0.776                                 & 0.123                                 & 0.665                                 & 0.765                                 & 0.718                                 & 0.080                                 & 0.618                                 & 0.762                                 & 0.714                                 & 0.083                                 & 0.704                                 & 0.797                                 & 0.772                                 \\
PGCL~\cite{basak2023pseudo}       & CVPR23                & 0.057                                 & 0.752                                 & 0.850                                 & 0.801                                 & 0.116                                 & 0.682                                 & 0.782                                 & 0.722                                 & 0.061                                 & 0.637                                 & 0.779                                 & 0.728                                 & 0.071                                 & 0.719                                 & 0.809                                 & 0.789                                 \\
EPS~\cite{lee2023saliency}        & TPAMI23               & 0.049                                 & 0.763                                 & 0.843                                 & 0.828                                 & 0.103                                 & 0.697                                 & 0.796                                 & 0.735                                 & 0.056                                 & 0.646                                 & 0.787                                 & 0.736                                 & 0.066                                 & 0.737                                 & 0.833                                 & 0.801                                 \\
CoSOD~\cite{chakraborty2024unsupervised}      & WACV24                & 0.055                                 & 0.758                                 & 0.856                                 & 0.830                                 & 0.099                                 & 0.702                                 & 0.793                                 & 0.730                                 & 0.055                                 & 0.650                                 & 0.795                                 & 0.740                                 & 0.070                                 & 0.726                                 & 0.825                                 & 0.792                                 \\
\rowcolor{c2!20} DTEN++     &    ---        & 0.066                                 & 0.742                                 & 0.840                                 & 0.781                                 & 0.116                                 & 0.683                                 & 0.779                                 & 0.725                                 & 0.075                                 & 0.627                                 & 0.774                                 & 0.719                                 & 0.081                                 & 0.714                                 & 0.806                                 & 0.782                                 \\
\rowcolor{c2!20} PGCL++     &     ---        & 0.053                                 & 0.764                                 & 0.862                                 & 0.808                                 & 0.110                                 & 0.691                                 & 0.789                                 & 0.726                                 & 0.056                                 & 0.655                                 & 0.791                                 & 0.736                                 & 0.068                                 & 0.730                                 & 0.818                                 & 0.796                                 \\
\rowcolor{c2!20} EPS++      &     ---        & {\color[HTML]{00B0F0} \textbf{0.046}} & {\color[HTML]{00B0F0} \textbf{0.775}} & 0.852                                 & 0.833                                 & 0.096                                 & {\color[HTML]{FF0000} \textbf{0.717}} & 0.807                                 & {\color[HTML]{00B0F0} \textbf{0.741}} & 0.052                                 & 0.661                                 & 0.796                                 & 0.742                                 & {\color[HTML]{00B0F0} \textbf{0.063}} & {\color[HTML]{00B0F0} \textbf{0.749}} & {\color[HTML]{00B0F0} \textbf{0.845}} & {\color[HTML]{00B0F0} \textbf{0.810}} \\
\rowcolor{c2!20} CoSOD++    &         ---       & 0.053                                 & 0.767                                 & {\color[HTML]{00B0F0} \textbf{0.869}} & {\color[HTML]{00B0F0} \textbf{0.838}} & {\color[HTML]{00B0F0} \textbf{0.095}} & 0.713                                 & {\color[HTML]{00B0F0} \textbf{0.808}} & 0.736                                 & {\color[HTML]{00B0F0} \textbf{0.051}} & 0.667                                 & {\color[HTML]{FF0000} \textbf{0.803}} & {\color[HTML]{FF0000} \textbf{0.747}} & 0.064                                 & 0.738                                 & 0.840                                 & 0.803                                 \\
\rowcolor{c2!20} SEE (Ours) &      ---       & {\color[HTML]{FF0000} \textbf{0.040}} & {\color[HTML]{FF0000} \textbf{0.793}} & {\color[HTML]{FF0000} \textbf{0.885}} & {\color[HTML]{FF0000} \textbf{0.852}} & {\color[HTML]{FF0000} \textbf{0.093}} & {\color[HTML]{00B0F0} \textbf{0.716}} & {\color[HTML]{FF0000} \textbf{0.810}} & {\color[HTML]{FF0000} \textbf{0.747}} & {\color[HTML]{FF0000} \textbf{0.046}} & {\color[HTML]{00B0F0} \textbf{0.679}} & {\color[HTML]{FF0000} \textbf{0.803}} & {\color[HTML]{00B0F0} \textbf{0.745}} & {\color[HTML]{FF0000} \textbf{0.060}} & {\color[HTML]{FF0000} \textbf{0.757}} & {\color[HTML]{FF0000} \textbf{0.863}} & {\color[HTML]{FF0000} \textbf{0.812}} \\ \bottomrule
	\end{tabular}}
	\label{table:CODSemi}
	\vspace{-0.4cm}
\end{table*} 
\begin{figure*}[ht]
	\centering
	\setlength{\abovecaptionskip}{-0.1cm}
	\begin{center}
		\includegraphics[width=0.85\linewidth]{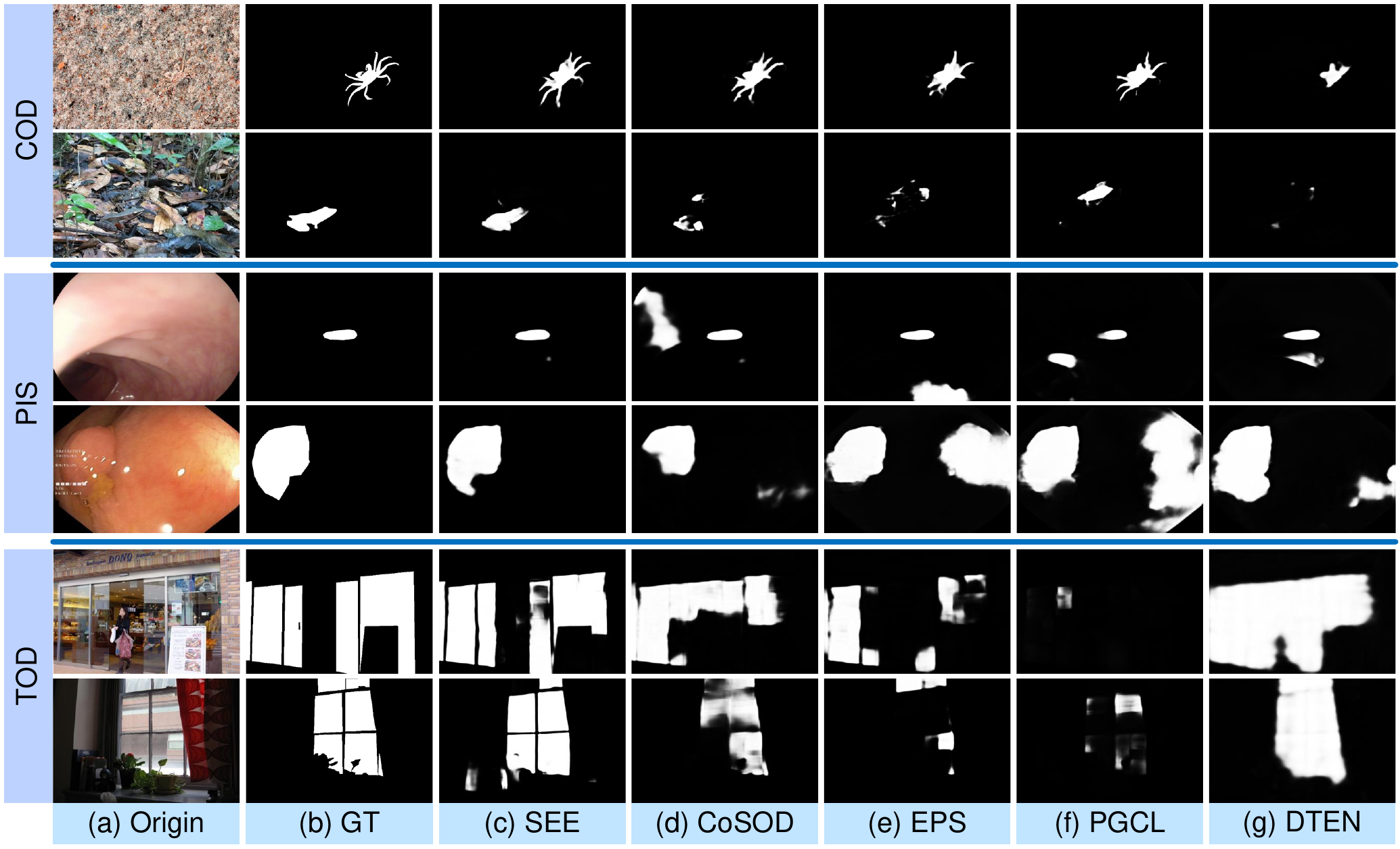}
	\end{center}
	\caption{Visualized results for COS tasks with 1/8 semi-supervision, including COD, PIS, and TOD.}
	\label{fig:COSQuali-semi}
	\vspace{-0.4cm}
\end{figure*}

\begin{table*}[t]
	\centering
	\setlength{\abovecaptionskip}{0cm}
	\caption{Results on PIS and TOD of the SSCOS task. 
	}
	\resizebox{2\columnwidth}{!}{
		\setlength{\tabcolsep}{0.8mm}
		\begin{tabular}{l|cccc|cccc|cccc|cccc|cccc}
			\toprule
			& \multicolumn{12}{c|}{Polyp Image Segmentation (PIS)}& \multicolumn{8}{c}{Transparant Object Detection (TOD)}\\ \cline{2-21}
			& \multicolumn{4}{c|}{\textit{CVC-ColonDB}}& \multicolumn{4}{c|}{\textit{ETIS}}& \multicolumn{4}{c|}{\textit{Kvasir}}& \multicolumn{4}{c|}{\textit{GDD}}& \multicolumn{4}{c}{\textit{GSD}}\\ \cline{2-21}
			\multirow{-3}{*}{Methods} & {\cellcolor{gray!40}$M$~$\downarrow$} &{\cellcolor{gray!40}$F_\beta$~$\uparrow$} &{\cellcolor{gray!40}$E_\phi$~$\uparrow$} & \multicolumn{1}{c|}{\cellcolor{gray!40}$S_\alpha$~$\uparrow$}& {\cellcolor{gray!40}$M$~$\downarrow$} &{\cellcolor{gray!40}$F_\beta$~$\uparrow$} &{\cellcolor{gray!40}$E_\phi$~$\uparrow$} & \multicolumn{1}{c|}{\cellcolor{gray!40}$S_\alpha$~$\uparrow$}& {\cellcolor{gray!40}$M$~$\downarrow$} &{\cellcolor{gray!40}$F_\beta$~$\uparrow$} &{\cellcolor{gray!40}$E_\phi$~$\uparrow$} & \multicolumn{1}{c|}{\cellcolor{gray!40}$S_\alpha$~$\uparrow$}& {\cellcolor{gray!40}$M$~$\downarrow$} &{\cellcolor{gray!40}$F_\beta$~$\uparrow$} &{\cellcolor{gray!40}$E_\phi$~$\uparrow$} & \multicolumn{1}{c|}{\cellcolor{gray!40}$S_\alpha$~$\uparrow$}& {\cellcolor{gray!40}$M$~$\downarrow$} &{\cellcolor{gray!40}$F_\beta$~$\uparrow$} &{\cellcolor{gray!40}$E_\phi$~$\uparrow$} & \multicolumn{1}{c}{\cellcolor{gray!40}$S_\alpha$~$\uparrow$}\\ \midrule
			\multicolumn{21}{c}{1/8 Labeled Training Data} \\ \midrule
			SAM~\cite{kirillov2023segment}        & 0.479                                 & 0.343                                 & 0.419                                 & 0.427                                 & 0.429                                 & 0.439                                 & 0.512                                 & 0.503                                 & 0.320                                 & 0.545                                 & 0.564                                 & 0.582                                 & 0.245                                 & 0.512                                 & 0.530                                 & 0.551                                 & 0.266                                 & 0.473                                 & 0.501                                 & 0.514                                 \\
SAM-S~\cite{chen2023sam}       & 0.185                                 & 0.517                                 & 0.627                                 & 0.661                                 & 0.172                                 & 0.558                                 & 0.706                                 & 0.682                                 & 0.131                                 & 0.738                                 & 0.772                                 & 0.783                                 & 0.185                                 & 0.674                                 & 0.703                                 & 0.618                                 & 0.177                                 & 0.688                                 & 0.724                                 & 0.652                                 \\
DTEN~\cite{ren2023towards}       & 0.073                                 & 0.607                                 & 0.722                                 & 0.743                                 & 0.064                                 & 0.607                                 & 0.759                                 & 0.757                                 & 0.078                                 & 0.837                                 & 0.867                                 & 0.833                                 & 0.096                                 & 0.815                                 & 0.818                                 & 0.733                                 & 0.102                                 & 0.803                                 & 0.792                                 & 0.736                                 \\
PGCL~\cite{basak2023pseudo}       & 0.068                                 & 0.629                                 & 0.743                                 & 0.749                                 & 0.057                                 & 0.623                                 & 0.771                                 & 0.764                                 & 0.070                                 & 0.838                                 & 0.869                                 & 0.855                                 & 0.097                                 & 0.819                                 & 0.823                                 & 0.747                                 & 0.095                                 & 0.817                                 & 0.804                                 & 0.741                                 \\
EPS~\cite{lee2023saliency}            & 0.060                                 & 0.645                                 & 0.758                                 & 0.756                                 & 0.053                                 & 0.642                                 & 0.786                                 & 0.772                                 & 0.063                                 & 0.847                                 & 0.883                                 & 0.858                                 & 0.085                                 & 0.837                                 & 0.842                                 & 0.759                                 & 0.089                                 & 0.829                                 & 0.817                                 & 0.754                                 \\
CoSOD~\cite{chakraborty2024unsupervised}      & 0.057                                 & 0.657                                 & 0.776                                 & 0.763                                 & 0.047                                 & 0.639                                 & 0.802                                 & 0.780                                 & 0.056                                 & 0.859                                 & 0.886                                 & 0.863                                 & 0.088                                 & 0.831                                 & 0.835                                 & 0.756                                 & 0.091                                 & 0.826                                 & 0.813                                 & 0.749                                 \\
\rowcolor{c2!20} DTEN++      & 0.071                                 & 0.620                                 & 0.733                                 & 0.750                                 & 0.062                                 & 0.622                                 & 0.773                                 & 0.765                                 & 0.074                                 & 0.849                                 & 0.882                                 & 0.838                                 & 0.092                                 & 0.827                                 & 0.829                                 & 0.738                                 & 0.095                                 & 0.819                                 & 0.808                                 & 0.747                                 \\
\rowcolor{c2!20} PGCL++      & 0.065                                 & 0.647                                 & 0.756                                 & 0.755                                 & 0.053                                 & 0.645                                 & 0.788                                 & 0.773                                 & 0.067                                 & 0.856                                 & 0.884                                 & 0.862                                 & 0.093                                 & 0.832                                 & 0.841                                 & 0.757                                 & 0.091                                 & 0.832                                 & 0.825                                 & 0.749                                 \\
\rowcolor{c2!20} EPS++       & 0.057                                 & 0.660                                 & 0.771                                 & 0.763                                 & 0.050                                 & {\color[HTML]{00B0F0} \textbf{0.663}} & 0.798                                 & 0.778                                 & 0.060                                 & 0.861                                 & 0.898                                 & 0.865                                 & {\color[HTML]{00B0F0} \textbf{0.081}} & {\color[HTML]{00B0F0} \textbf{0.844}} & {\color[HTML]{00B0F0} \textbf{0.853}} & {\color[HTML]{00B0F0} \textbf{0.763}} & {\color[HTML]{00B0F0} \textbf{0.086}} & {\color[HTML]{FF0000} \textbf{0.837}} & {\color[HTML]{00B0F0} \textbf{0.827}} & {\color[HTML]{00B0F0} \textbf{0.759}} \\
\rowcolor{c2!20} CoSOD++  & {\color[HTML]{00B0F0} \textbf{0.053}} & {\color[HTML]{00B0F0} \textbf{0.674}} & {\color[HTML]{00B0F0} \textbf{0.793}} & {\color[HTML]{00B0F0} \textbf{0.768}} & {\color[HTML]{00B0F0} \textbf{0.041}} & 0.657                                 & {\color[HTML]{00B0F0} \textbf{0.822}} & {\color[HTML]{00B0F0} \textbf{0.788}} & {\color[HTML]{00B0F0} \textbf{0.052}} & {\color[HTML]{FF0000} \textbf{0.874}} & {\color[HTML]{00B0F0} \textbf{0.900}} & {\color[HTML]{FF0000} \textbf{0.870}} & 0.083                                 & 0.842                                 & 0.846                                 & 0.760                                 & 0.087                                 & 0.832                                 & 0.820                                 & 0.755                                 \\
\rowcolor{c2!20} SEE (Ours) & {\color[HTML]{FF0000} \textbf{0.045}} & {\color[HTML]{FF0000} \textbf{0.706}} & {\color[HTML]{FF0000} \textbf{0.828}} & {\color[HTML]{FF0000} \textbf{0.801}} & {\color[HTML]{FF0000} \textbf{0.039}} & {\color[HTML]{FF0000} \textbf{0.692}} & {\color[HTML]{FF0000} \textbf{0.832}} & {\color[HTML]{FF0000} \textbf{0.792}} & {\color[HTML]{FF0000} \textbf{0.051}} & {\color[HTML]{00B0F0} \textbf{0.867}} & {\color[HTML]{FF0000} \textbf{0.905}} & {\color[HTML]{00B0F0} \textbf{0.869}} & {\color[HTML]{FF0000} \textbf{0.076}} & {\color[HTML]{FF0000} \textbf{0.846}} & {\color[HTML]{FF0000} \textbf{0.855}} & {\color[HTML]{FF0000} \textbf{0.768}} & {\color[HTML]{FF0000} \textbf{0.085}} & {\color[HTML]{00B0F0} \textbf{0.835}} & {\color[HTML]{FF0000} \textbf{0.832}} & {\color[HTML]{FF0000} \textbf{0.763}} \\ \midrule
\multicolumn{21}{c}{1/16 Labeled Training Data} \\ \midrule
SAM~\cite{kirillov2023segment}        & 0.479                                 & 0.343                                 & 0.419                                 & 0.427                                 & 0.429                                 & 0.439                                 & 0.512                                 & 0.503                                 & 0.320                                 & 0.545                                 & 0.564                                 & 0.582                                 & 0.245                                 & 0.512                                 & 0.530                                 & 0.551                                 & 0.266                                 & 0.473                                 & 0.501                                 & 0.514                                 \\
SAM-S~\cite{chen2023sam}        & 0.201                                 & 0.471                                 & 0.593                                 & 0.643                                 & 0.183                                 & 0.533                                 & 0.677                                 & 0.660                                 & 0.145                                 & 0.712                                 & 0.753                                 & 0.762                                 & 0.203                                 & 0.657                                 & 0.672                                 & 0.604                                 & 0.193                                 & 0.654                                 & 0.684                                 & 0.627                                 \\
DTEN~\cite{ren2023towards}      & 0.080                                 & 0.586                                 & 0.696                                 & 0.732                                 & 0.071                                 & 0.592                                 & 0.730                                 & 0.750                                 & 0.086                                 & 0.813                                 & 0.824                                 & 0.833                                 & 0.112                                 & 0.797                                 & 0.810                                 & 0.736                                 & 0.115                                 & 0.786                                 & 0.770                                 & 0.705                                 \\
PGCL~\cite{basak2023pseudo}     & 0.075                                 & 0.608                                 & 0.719                                 & 0.745                                 & 0.063                                 & 0.614                                 & 0.753                                 & 0.754                                 & 0.078                                 & 0.836                                 & 0.845                                 & 0.835                                 & 0.108                                 & 0.794                                 & 0.803                                 & 0.735                                 & 0.108                                 & 0.800                                 & 0.788                                 & 0.718                                 \\
EPS~\cite{lee2023saliency}     & 0.065                                 & 0.631                                 & 0.746                                 & 0.750                                 & 0.057                                 & 0.631                                 & 0.774                                 & 0.755                                 & 0.068                                 & 0.835                                 & 0.863                                 & 0.846                                 & 0.089                                 & 0.818                                 & 0.825                                 & 0.750                                 & 0.097                                 & 0.812                                 & 0.803                                 & 0.737                                 \\
CoSOD~\cite{chakraborty2024unsupervised}     & 0.063                                 & 0.638                                 & 0.754                                 & 0.755                                 & 0.050                                 & 0.632                                 & 0.781                                 & 0.767                                 & 0.066                                 & 0.840                                 & 0.871                                 & 0.852                                 & 0.095                                 & 0.813                                 & 0.816                                 & 0.746                                 & 0.105                                 & 0.793                                 & 0.792                                 & 0.732                                 \\
\rowcolor{c2!20} DTEN++     & 0.077                                 & 0.608                                 & 0.715                                 & 0.739                                 & 0.065                                 & 0.615                                 & 0.747                                 & 0.759                                 & 0.081                                 & 0.827                                 & 0.843                                 & 0.840                                 & 0.101                                 & 0.810                                 & 0.819                                 & 0.743                                 & 0.107                                 & 0.798                                 & 0.784                                 & 0.716                                 \\
\rowcolor{c2!20} PGCL++      & 0.071                                 & 0.623                                 & 0.733                                 & 0.751                                 & 0.060                                 & 0.620                                 & 0.769                                 & 0.760                                 & 0.073                                 & 0.849                                 & 0.861                                 & 0.843                                 & 0.099                                 & 0.809                                 & 0.821                                 & 0.752                                 & 0.097                                 & 0.812                                 & 0.803                                 & 0.727                                 \\
\rowcolor{c2!20} EPS++        & 0.062                                 & 0.645                                 & {\color[HTML]{00B0F0} \textbf{0.768}} & 0.762                                 & 0.053                                 & {\color[HTML]{00B0F0} \textbf{0.650}} & 0.792                                 & 0.763                                 & 0.063                                 & 0.849                                 & 0.882                                 & 0.854                                 & {\color[HTML]{00B0F0} \textbf{0.085}} & {\color[HTML]{00B0F0} \textbf{0.832}} & {\color[HTML]{FF0000} \textbf{0.839}} & {\color[HTML]{00B0F0} \textbf{0.758}} & {\color[HTML]{00B0F0} \textbf{0.091}} & {\color[HTML]{FF0000} \textbf{0.828}} & {\color[HTML]{FF0000} \textbf{0.826}} & 0.745                                 \\
\rowcolor{c2!20} CoSOD++     & {\color[HTML]{00B0F0} \textbf{0.059}} & {\color[HTML]{00B0F0} \textbf{0.653}} & 0.767                                 & {\color[HTML]{00B0F0} \textbf{0.764}} & {\color[HTML]{00B0F0} \textbf{0.047}} & {\color[HTML]{00B0F0} \textbf{0.650}} & {\color[HTML]{00B0F0} \textbf{0.798}} & {\color[HTML]{00B0F0} \textbf{0.773}} & {\color[HTML]{00B0F0} \textbf{0.062}} & {\color[HTML]{00B0F0} \textbf{0.857}} & {\color[HTML]{00B0F0} \textbf{0.889}} & {\color[HTML]{00B0F0} \textbf{0.858}} & 0.090                                 & 0.826                                 & 0.830                                 & 0.753                                 & 0.098                                 & 0.815                                 & 0.812                                 & {\color[HTML]{00B0F0} \textbf{0.747}} \\
\rowcolor{c2!20} SEE (Ours)    & {\color[HTML]{FF0000} \textbf{0.050}} & {\color[HTML]{FF0000} \textbf{0.693}} & {\color[HTML]{FF0000} \textbf{0.812}} & {\color[HTML]{FF0000} \textbf{0.795}} & {\color[HTML]{FF0000} \textbf{0.043}} & {\color[HTML]{FF0000} \textbf{0.674}} & {\color[HTML]{FF0000} \textbf{0.811}} & {\color[HTML]{FF0000} \textbf{0.784}} & {\color[HTML]{FF0000} \textbf{0.057}} & {\color[HTML]{FF0000} \textbf{0.858}} & {\color[HTML]{FF0000} \textbf{0.893}} & {\color[HTML]{FF0000} \textbf{0.861}} & {\color[HTML]{FF0000} \textbf{0.080}} & {\color[HTML]{FF0000} \textbf{0.827}} & {\color[HTML]{00B0F0} \textbf{0.831}} & {\color[HTML]{FF0000} \textbf{0.759}} & {\color[HTML]{FF0000} \textbf{0.092}} & {\color[HTML]{00B0F0} \textbf{0.824}} & {\color[HTML]{00B0F0} \textbf{0.817}} & {\color[HTML]{FF0000} \textbf{0.750}} \\
  \bottomrule
	\end{tabular}} \label{table:MISTODSemi}
	\vspace{-0.2cm}
\end{table*}

\begin{table*}[t]
	\begin{minipage}[c]{0.457\textwidth}
		\centering
		\setlength{\abovecaptionskip}{0cm}
		\caption{Breakdown studies for the SEE framework. 
		}
		\resizebox{\columnwidth}{!}{
			\setlength{\tabcolsep}{1mm}
			\begin{tabular}{cccc|cccc}
				\toprule
				Baseline & PLG & $\text{PLS}_{\text{t}}$ & $\text{PLS}_{\text{u}}$ & $M$~$\downarrow$ & $F_\beta$~$\uparrow$ & $E_\phi$~$\uparrow$ & $S_\alpha$~$\uparrow$ \\ \midrule
				\checkmark  &  &   & & 0.061          & 0.627          & 0.815          & 0.720          \\
				\checkmark & \checkmark  &   &  &0.045          & 0.695          & 0.861          & 0.783          \\
				\checkmark &\checkmark &\checkmark  & & 0.040          & 0.703          & 0.871          & 0.795          \\
				\rowcolor{c2!20}\checkmark & \checkmark&\checkmark  &\checkmark & \textbf{0.036} & \textbf{0.729} & \textbf{0.883} & \textbf{0.807} \\ \bottomrule
		\end{tabular}}\label{table:AblationSAM} \vspace{-3mm}
	\end{minipage}
	\begin{minipage}[c]{0.533\textwidth}
		\centering
		\setlength{\abovecaptionskip}{0cm}
		\caption{Ablation studies for the HGFG module. 
		}
		\resizebox{\columnwidth}{!}{
			\setlength{\tabcolsep}{1mm}
			\begin{tabular}{c|ccccc}
				\toprule
				Metrics & w/o HGFG & SA$\rightarrow$FG & w/o RK2 & WGM$\rightarrow$FC &\cellcolor{c2!20} w/ HGFG         \\ \midrule
				$M$~$\downarrow$  & 0.043   & 0.038               & 0.039          & 0.039                &\cellcolor{c2!20} \textbf{0.036} \\
				$F_\beta$~$\uparrow$ & 0.688   & 0.712               & 0.704          & 0.711                &\cellcolor{c2!20} \textbf{0.729} \\
				$E_\phi$~$\uparrow$ & 0.861   & 0.872               & 0.861          & 0.874                &\cellcolor{c2!20} \textbf{0.883} \\
				$S_\alpha$~$\uparrow$  & 0.785   & 0.798               & 0.785          & 0.793                &\cellcolor{c2!20} \textbf{0.807} \\ \bottomrule
		\end{tabular}}\label{table:AblationMFG} \vspace{-3mm}
	\end{minipage}
\end{table*}

\begin{table}[t]
		\centering
		\setlength{\abovecaptionskip}{0cm}
		\caption{Ablations of SEE with SAM2 in \textit{COD10K}. 
		}
		\resizebox{0.95\columnwidth}{!}{
			\setlength{\tabcolsep}{1mm}
			\begin{tabular}{cccc|cccc}
				\toprule
				Baseline & PLG & $\text{PLS}_{\text{t}}$ & $\text{PLS}_{\text{u}}$ & $M$~$\downarrow$ & $F_\beta$~$\uparrow$ & $E_\phi$~$\uparrow$ & $S_\alpha$~$\uparrow$ \\ \midrule
				\checkmark  &  &   &  & 0.061          & 0.627          & 0.815          & 0.720          \\
				\checkmark & \checkmark  &   &  & 0.044          & 0.702          & 0.867          & 0.786          \\
				\checkmark &\checkmark &\checkmark  & & 0.040          & 0.711          & 0.880          & 0.798          \\
				\rowcolor{c2!20}\checkmark & \checkmark&\checkmark  &\checkmark & \textbf{0.036} & \textbf{0.730} & \textbf{0.885} & \textbf{0.806} \\ \bottomrule
		\end{tabular}}\label{table:SAM2}
		\vspace{-0.2cm}
\end{table}

\begin{table*}[t]
\begin{minipage}[c]{\textwidth}
\centering
		\setlength{\abovecaptionskip}{0cm}
		\caption{Ablation studies in pseudo-label generation. SA and WA are strong augmentations and weak augmentations, where SA is provided by RandAugment~\cite{cubuk2020randaugment}. MARF-T and MARF-S refer to applying the fusion strategy to the teacher model and SAM. Prom. \#1 to \#3 are the breakdown of prompt usage {while Prom. \#4 to \#6 are three prompt selection strategies.} 
		}
		\resizebox{\columnwidth}{!}{
			\setlength{\tabcolsep}{0.6mm}
\begin{tabular}{l|cc|cc|cccccc|c|c}
\toprule
    & \multicolumn{2}{c|}{{Different augmentations}} & \multicolumn{2}{c|}{{Effect of MARF}} & \multicolumn{6}{c|}{{Effect of MDPE}} & Training with&{\cellcolor{c2!20} }           \\  \cline{2-11}
\multirow{-2}{*}{Metrics} & SA$\rightarrow$WA   & SA \& WA  & w/o MARF-T              & w/o MARF-S             & Prom. \#1  & Prom. \#2  & Prom. \#3  & {Prom. \#4} & {Prom. \#5} & {Prom. \#6} & coarse mask $\tilde{\mathbf{M}}^t_i$& \multirow{-2}{*}{{\cellcolor{c2!20}  Ours}} \\ \midrule
$M$~$\downarrow$        & 0.050                                 & 0.038                    & 0.039                         & 0.041                     & 0.039        & 0.037        & 0.038         &0.038  & 0.037 &  \textbf{0.036}  &0.037    &\cellcolor{c2!20}  \textbf{0.036}                                \\
$F_\beta$~$\uparrow$          & 0.676                                 & 0.708                    & 0.710                         & 0.702                     & 0.713        & 0.721        & 0.716      &  0.717   & 0.723  &  \textbf{0.732}  &0.725     &\cellcolor{c2!20}  {0.729}                                \\
$E_\phi$~$\uparrow$            & 0.839                                 & 0.865                    & 0.868                         & 0.853                     & 0.869        & 0.875        & 0.868      &  0.865   & 0.877  &  \textbf{0.889}   &0.872     &\cellcolor{c2!20}  {0.883}                                \\
$S_\alpha$~$\uparrow$         & 0.746                                 & 0.799                    & 0.794                         & 0.786                     & 0.797        & 0.806        & 0.802       & 0.803   & 0.803    &     \textbf{0.808}&0.801     &\cellcolor{c2!20}  {0.807}    \\  \bottomrule                           
\end{tabular}}\label{table:AblationPLG}
\vspace{1mm}
\end{minipage}
\\ 
\begin{minipage}[c]{0.66\textwidth}
\centering
		\setlength{\abovecaptionskip}{0cm}
		\caption{Explorations of the extraction manner of three prompts.
		}
		\resizebox{\columnwidth}{!}{
			\setlength{\tabcolsep}{0.8mm}
\begin{tabular}{l|cc|ccc|c|c}
\toprule
                          & \multicolumn{2}{c|}{Points} & \multicolumn{3}{c|}{Box }          & Mask    & {\cellcolor{c2!20} }                       \\ \cline{2-7}
\multirow{-2}{*}{Metrics} & w/o NBS      & NBS \#1     & w/o Expand. & Expand. \#1 & Expand. \#2 & w/o Filter & \multirow{-2}{*}{{\cellcolor{c2!20} Ours}} \\ \midrule
$M$~$\downarrow$  & 0.038        & 0.037             & 0.038       & 0.037       & 0.036       & 0.037           &\cellcolor{c2!20} \textbf{0.036}                                \\
$F_\beta$~$\uparrow$        & 0.713        & 0.721             & 0.710       & 0.715       & 0.717       & 0.720           &\cellcolor{c2!20} \textbf{0.729}                                \\
$E_\phi$~$\uparrow$                   & 0.870        & 0.880             & 0.872       & 0.873       & 0.882       & 0.872           &\cellcolor{c2!20} \textbf{0.883}                                \\
$S_\alpha$~$\uparrow$                       & 0.792        & 0.800             & 0.794       & 0.806       & 0.803       & \textbf{0.808}  &\cellcolor{c2!20} 0.807          \\  \bottomrule                                
\end{tabular}}\label{table:AblationPromptExtraction}
\vspace{1mm}
\end{minipage}
\begin{minipage}[c]{0.33\textwidth}
\centering
	\setlength{\abovecaptionskip}{-1mm}
	\begin{center}
		\includegraphics[width=0.95\linewidth]{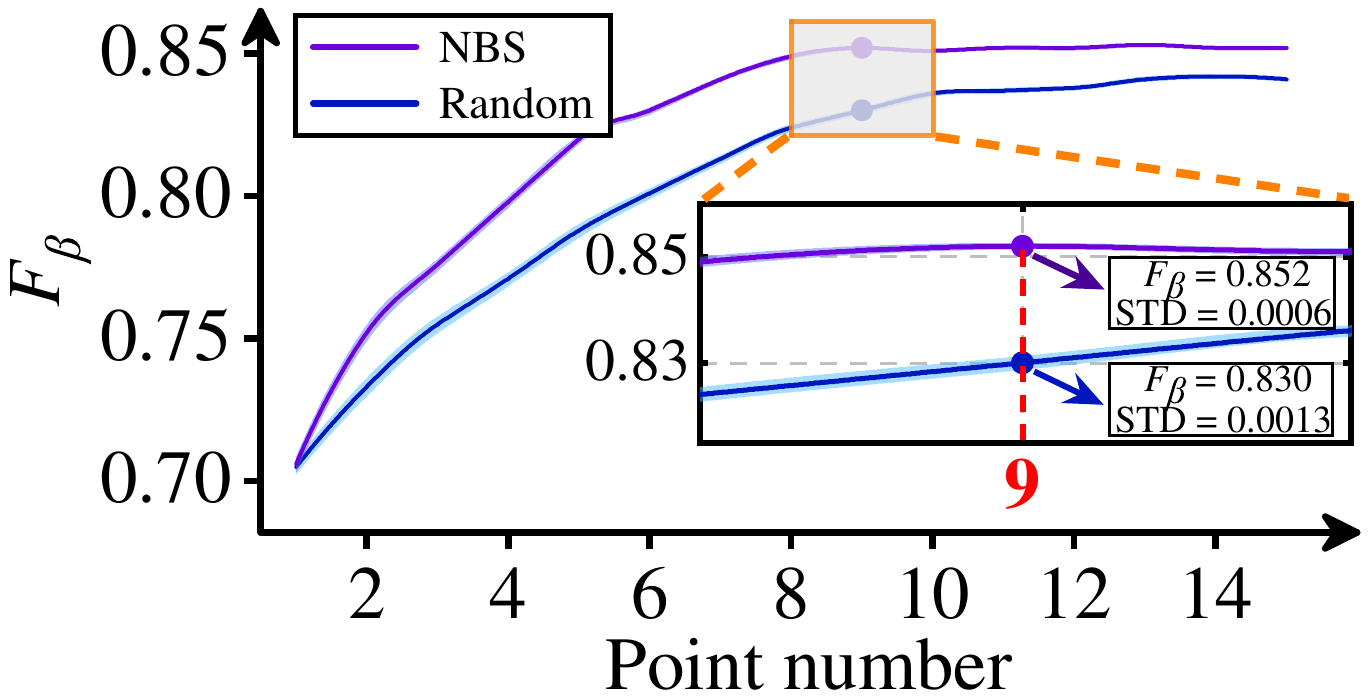}
	\end{center}
	\captionof{figure}{Analysis of point number $C$.}
	\label{fig:PointNumber}
	\vspace{1mm}
\end{minipage}
\\
\begin{minipage}[c]{0.67\textwidth}
\centering
		\setlength{\abovecaptionskip}{0cm}
		\caption{Ablation studies of pseudo-label storage and supervision.
		}
		\resizebox{\columnwidth}{!}{
			\setlength{\tabcolsep}{1.2mm}
\begin{tabular}{l|cccc|ccc|c}
\toprule
Metrics & Stor. \#1 & Stor. \#2    & Stor. \#3 & Stor. \#4 & Sup. \#1 & Sup. \#2 & Sup. \#3 & {\cellcolor{c2!20}   Ours} \\ \midrule
$M$~$\downarrow$   & 0.037       & 0.036          & 0.038       & 0.038       & 0.039           & 0.038           & 0.038           &\cellcolor{c2!20}  \textbf{0.036}              \\
$F_\beta$~$\uparrow$ & 0.724       & 0.723          & 0.716       & 0.718       & 0.708           & 0.715           & 0.723           &\cellcolor{c2!20}  \textbf{0.729}              \\
$E_\phi$~$\uparrow$ & 0.879       & \textbf{0.885} & 0.869       & 0.875       & 0.873           & 0.880           & 0.876           &\cellcolor{c2!20}  0.883                       \\
$S_\alpha$~$\uparrow$    & 0.803       & 0.801          & 0.790       & 0.797       & 0.798           & 0.805           & 0.802           &\cellcolor{c2!20}  \textbf{0.807}         \\ \bottomrule    
\end{tabular}}\label{table:AblationPLS}
\vspace{2mm}
\end{minipage}
\begin{minipage}[c]{0.32\textwidth}
\centering
	\setlength{\abovecaptionskip}{-1mm}
	\begin{center}
		\includegraphics[width=0.9\linewidth]{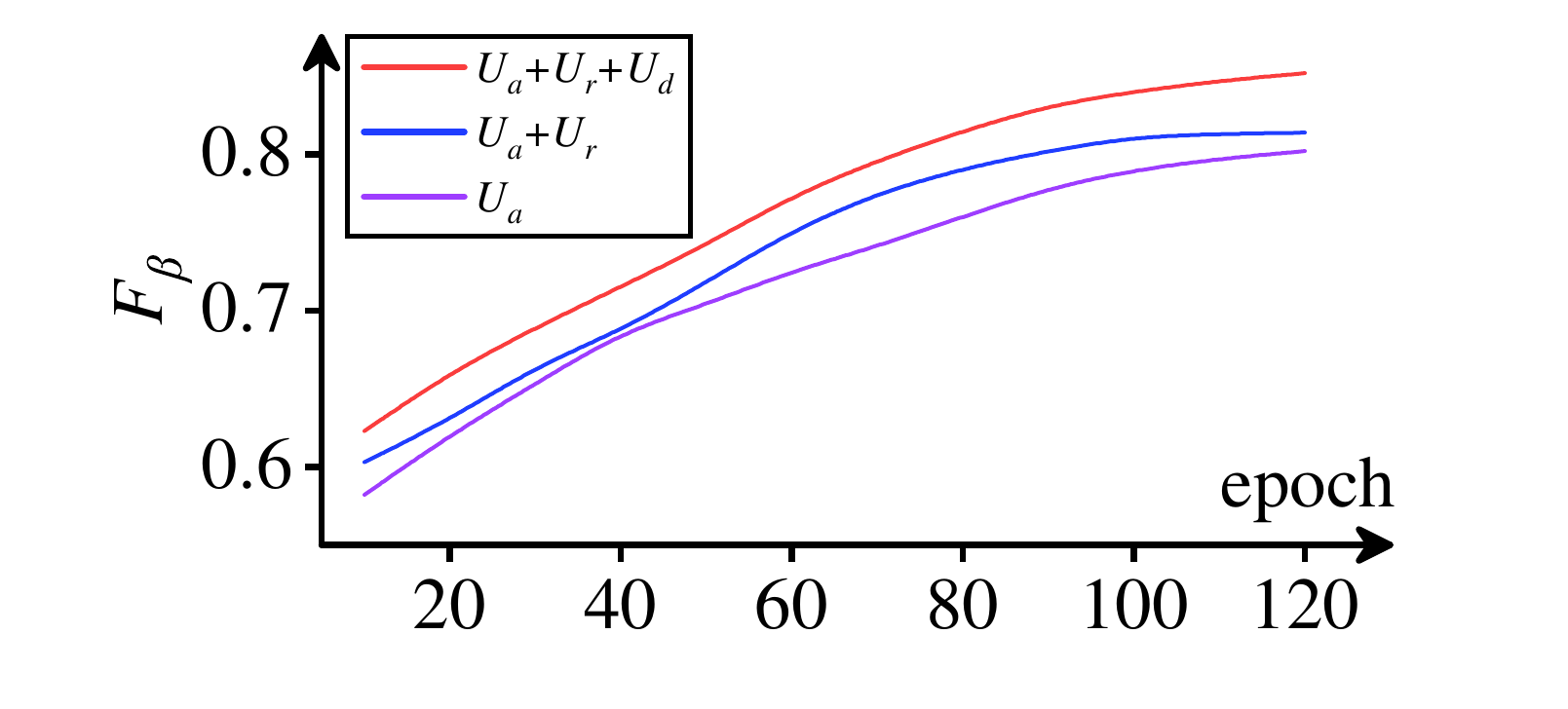}
	\end{center}
	\captionof{figure}{Quality of stored pseudo-labels.}
	\label{fig:PLSt}
	\vspace{2mm}
\end{minipage}
\\
\begin{minipage}[c]{\textwidth}
\centering
	\setlength{\abovecaptionskip}{0cm}
	\caption{Parameter analysis on the number of aumented views $K$ in pseudo-label generation, the number of stored best labels $B$ in pseudo-label storage, and the uncertainty thresholds $\tau_a$ and $\tau_r$ in pseudo-label selection.}
	\resizebox{\columnwidth}{!}{
		\setlength{\tabcolsep}{1.4mm}
		\begin{tabular}{c|cccc|cccc|cccc|cccc}
			\toprule
			\multicolumn{1}{l|}{\multirow{2}{*}{Metrics}} & \multicolumn{4}{c|}{$K$}  & \multicolumn{4}{c|}{$B$} & \multicolumn{4}{c|}{$\tau_a$} & \multicolumn{4}{c}{$\tau_r$}\\\cline{2-17}
			\multicolumn{1}{l|}{}  & 1     & 6     & \cellcolor{c2!20}12             & 18             & 1 &2           & \cellcolor{c2!20}3            & 4  & 0.05           & \cellcolor{c2!20}0.1            & 0.2            & 0.3   & 0.3            & \cellcolor{c2!20}0.5            & 0.7   & 0.9            \\ \midrule
			$M$~$\downarrow$ & 0.051 & 0.039 &\cellcolor{c2!20} \textbf{0.036}            & \textbf{0.036} & 0.038 & \textbf{0.036} &\cellcolor{c2!20} \textbf{0.036}           & \textbf{0.036} & \textbf{0.037} & \cellcolor{c2!20}0.038          & 0.038          & 0.040 & \textbf{0.038} & \cellcolor{c2!20}\textbf{0.038} & 0.039 & 0.040 \\
			$F_\beta$~$\uparrow$ & 0.674 & 0.697 &\cellcolor{c2!20} \textbf{0.729}            & 0.728          & 0.704 & 0.718          &\cellcolor{c2!20} 0.729           & \textbf{0.731} & 0.706          & \cellcolor{c2!20}\textbf{0.719} & 0.716          & 0.704 & \textbf{0.723} & \cellcolor{c2!20}0.719          & 0.715 & 0.700  \\
			$E_\phi$~$\uparrow$ & 0.838 & 0.857 &\cellcolor{c2!20} \textbf{0.883}            & 0.880          & 0.862 & 0.877          &\cellcolor{c2!20} \textbf{0.883}           & 0.882  & 0.868          & \cellcolor{c2!20}\textbf{0.878} & 0.876          & 0.865 & 0.866          & \cellcolor{c2!20}\textbf{0.878} & 0.874 & 0.851  \\
			$S_\alpha$~$\uparrow$ & 0.737 & 0.776 &\cellcolor{c2!20} \textbf{0.807}            & 0.805          & 0.786 & 0.796          &\cellcolor{c2!20} \textbf{0.807}           & 0.803  & 0.795          & \cellcolor{c2!20}0.803          & \textbf{0.805} & 0.793 & 0.792          & \cellcolor{c2!20}\textbf{0.803} & 0.789 & 0.781     \\ \bottomrule
	\end{tabular}}\label{table:ParameterAnalysisKB}
	\vspace{-0.2cm}
\end{minipage}
\end{table*}

\subsection{Incompletely Supervised Concealed Object Segmentation}
{In this section, we present a complete ISCOS segmenter, updated within the SEE framework, which employs HGFG for feature aggregation. As shown in Fig.~\ref{fig:Segmenter}, we begin by integrating HGFG with a commonly-used basic encoder-decoder architecture, which has been used in several methods, such as SINet~\cite{fan2020camouflaged} and FEDER~\cite{He2023Camouflaged}
to construct a novel segmenter. 
 
\noindent \textbf{Encoder}. Given a concealed image $\mathbf{X}_i$ of size $H\times W$, we start by using the basic encoder $E$, implemented with ResNet50~\cite{he2016deep}, to encode $\mathbf{X}_i$ into a set of features $\{\tilde{\mathbf{F}}_i^k\}_{k=0}^4$ with the resolution of $\frac{H}{2^{k+1}}\times \frac{W}{2^{k+1}}$. Next, the R-Net~\cite{fan2020camouflaged} $R$ is cascaded to transform $\{\tilde{\mathbf{F}}_i^k\}_{k=1}^4$ to $\{\mathbf{F}_i^k\}_{k=1}^4$, reducing the feature space to a more compact form with 64 channels for channel reduction. Besides, since $\tilde{\mathbf{F}}_i^4$ is rich in semantic information, we pass it through an atrous spatial pyramid pooling (ASPP)~\cite{yang2018denseaspp} $A_s$ to generate a coarse prediction map $\tilde{\mathbf{Y}}_i^5$, to guide the segmentation: $\tilde{\mathbf{Y}}_i^5=A_s(\tilde{\mathbf{F}}_i^4)$.

\noindent \textbf{HGFG}. To balance performance and efficiency, we apply HGFG only to the bottom layer, obtaining $\dot{\mathbf{F}}_i^4$, as the semantic information enriched by the deep features in this layer is most conducive to HGFG for mining feature coherence. For cross-layer feature aggregation, we employ the joint attention module $JA(\cdot)$~\cite{He2023Camouflaged}, which integrates spatial attention~\cite{zhao2018psanet} and channel attention~\cite{wang2020eca}:
\begin{equation}
	\ddot{\mathbf{F}}_i^k = JA(\mathbf{F}_i^k,up(\ddot{\mathbf{F}}_i^{k+1})),
\end{equation}
where $\ddot{\mathbf{F}}_i^4 = \dot{\mathbf{F}}_i^4$ and $up(\bigcdot)$ denotes the up-sampling operation.
Having combined the cross-layer features, we further define the latent features $\{\hat{\mathbf{F}}_i^k\}_{k=1}^3$ conveyed to the decoder:
\begin{equation}
	\hat{\mathbf{F}}_i^k = conv1(cat(\mathbf{F}_i^k, \ddot{\mathbf{F}}_i^k)),
\end{equation}
where $cat(\bigcdot)$ denote the concatenation operation and $conv1(\bigcdot)$ means $1\times1$ convolution. Notice that $\hat{\mathbf{F}}_i^4 = \dot{\mathbf{F}}_i^4$ here.

\noindent \textbf{Decoder}. Prediction maps generated by segmenters trained with incomplete annotations often contain low-confidence, ambiguous regions due to the complex scenes involving concealed objects and the limited discriminative capacity of the segmenter. To address this issue, we leverage the prediction map from the previous decoder to extract cues from these low-confidence regions, enabling the segmenter to more effectively detect previously undetected parts. The prediction map $\{\tilde{\mathbf{Y}}_i^k\}_{k=1}^4$ is defined as follows:
\begin{equation}
	\tilde{\mathbf{Y}}_i^k = conv3(RCAB(\hat{\mathbf{F}}_i^k \odot rv(S(rp(\tilde{\mathbf{Y}}_i^{k+1}))))),
\end{equation}
where $rp(\bigcdot)$, $S(\bigcdot)$, $rv(\bigcdot)$, $\odot$, and $conv3(\bigcdot)$ denote repeat, Sigmoid, reverse (element-wise subtraction with 1), Hadamard product, and $3\times 3$ convolution. To emphasize noteworthy information, we employ the residual channel attention block~\cite{zhang2018image} $RCAB(\bigcdot)$. 
The final segmentation result $\mathbf{Y}'_i$ is $\tilde{\mathbf{Y}}_i^1$.}

{ \noindent \textbf{Optimization with weak supervision}.
Following common practice \cite{he2022weakly}, we train the student model using the sparse annotations $\mathbf{Y}_i$, the top-$B$ best-ever pseudo-label $\{\tilde{\mathbf{M}}_i^b\}_{b=1}^B$, and the corresponding weighted maps $\{\hat{\mathbf{Y}}_i^b\}_{b=1}^B$, as 
\begin{equation}
\begin{aligned}
	L_w& = 
 \sum_{b=1}^B \left[\hat{\mathbf{Y}}_i^b L_{ce}(\mathbf{Y}'_i, \tilde{\mathbf{M}}_i^b)
 +\hat{\mathbf{Y}}_i^b L_{IoU}(\mathbf{Y}'_i, \tilde{\mathbf{M}}_i^b)\right] \\
 &+ L_{pce}\left(\mathbf{Y}_i', \mathbf{Y}_i\right),
\end{aligned} \label{Eq:WeakSup}
\end{equation}
where the first term combines the cross-entropy loss $L_{ce}$ and the intersection-over-union loss $L_{IoU}$, both 
focusing on
the pseudo-labels $\{\tilde{\mathbf{M}}_i^b\}_{b=1}^B$ and the weights $\{\hat{\mathbf{Y}}_i^b\}_{b=1}^B$. The second term is the partial cross-entropy loss $L_{pce}$, ensuring consistency between the predictions and the sparse annotations.

Concurrently, the teacher model is updated using the exponential moving average (EMA) strategy, which distills valuable knowledge from the student model while minimizing noise interference. This update is formulated as follows:
\begin{equation}\label{Eq:EMA}
    \theta_{t} = \eta \theta_{t} + (1- \eta)\theta_{s},
\end{equation}
where $\theta_{t}$ and $\theta_{s}$ denote the weights of the teacher and student models. The momentum parameter $\eta$ is set to 0.996 here.

}

{\noindent \textbf{Optimization with semi-supervision}. 
In semi-supervision, the segmenters are optimized similarly, where the student model is optimized by $L_s$:
\begin{equation}\label{Eq:SemiSup}
\begin{aligned}
	L_s  &= \sum_{b=1}^B \left[\hat{\mathbf{Y}}_i^b L_{ce}(\mathbf{Y}'_i, \tilde{\mathbf{M}}_i^b)
 +\hat{\mathbf{Y}}_i^b L_{IoU}(\mathbf{Y}'_i, \tilde{\mathbf{M}}_i^b)\right] \\
 &+\left[L_{ce}\left(\mathbf{Y}_i', \mathbf{Y}_i\right)+L_{IoU}\left(\mathbf{Y}_i', \mathbf{Y}_i\right)\right],
\end{aligned}
\end{equation}
where the first and second components correspond to ground-truth-independent and ground-truth-dependent constraints, respectively. 
Besides, the teacher model is updated using the EMA strategy, as  in~\cref{Eq:EMA}.

}

\section{Experiments}\label{Sec:Experiment}
\subsection{Experimental Setup}\label{Sec:ExperimentSet}
{\noindent \textbf{Implementation details}. We implement our method with PyTorch and run experiments on RTX4090 GPUs. For the proposed segmenter, the image encoder uses ResNet50 as the backbone and is pre-trained on ImageNet~\cite{deng2009imagenet}. The batch size is 12 and the learning rate is initialized as 0.0001, decreased by 0.1 every 80 epochs. Adam is employed to optimize the segmenter with momentum terms $(0.9,0.999)$.
 For the SEE framework, we adopt the ViT-H SAM model to generate pseudo-labels.
The number of augmentation images $K$ and stored images $B$ are set to 12 and 3, respectively. Following standard practices~\cite{fan2020camouflaged}, all images are resized as $352\times352$ during both the training and testing phases.

\noindent \textbf{Baselines}. In our research, we investigate the application of SAM \cite{kirillov2023segment} for the ISCOS task by generating segmentation masks from multi-density prompts and using these masks to train a COS segmenter. An alternative and more straightforward approach would be to fine-tune SAM using the incomplete annotations and directly apply it for testing. To evaluate the effect of our method compared to this straightforward approach, we have developed two baseline methods: SAM-W and SAM-S. These methods fine-tune SAM under weak and semi-supervised conditions, respectively, following the SAM-adapter strategy \cite{chen2023sam}. To ensure a fair comparison, both SAM-W and SAM-S employ a mean-teacher framework similar to our SEE framework, comprising a teacher SAM and a student SAM. In the weak supervision scenario, SAM-W includes two models: $\text{SAM-W}_\text{P}$, trained with point annotations, and $\text{SAM-W}_\text{S}$, trained with scribble annotations. We will present the performance of these baselines alongside our SEE method in comparative evaluations. For additional context, we also report the performance of the vanilla SAM. When applying SAM and its variants (SAM-S and SAM-W) to test images, we utilize an automatic prompt generation strategy and report the results with the highest Intersection over Union (IoU) scores.

\noindent \textbf{Metrics}. Building on previous methods \cite{he2025run,fang2025integrating}, we employ four widely recognized metrics to evaluate segmentation performance: mean absolute error ($M$), adaptive F-measure ($F_\beta$) \cite{margolin2014evaluate}, mean E-measure ($E_\phi$) \cite{fan2021cognitive}, and structure measure ($S_\alpha$) \cite{fan2017structure}. Superior segmentation performance is indicated by a lower $M$ value, and higher values of $F_\beta$, $E_\phi$, and $S_\alpha$. { Detailed descriptions can be seen in the supplementary materials.}

\noindent \textbf{Datasets}. We conducted extensive experiments on camouflaged object detection (COD), polyp image segmentation (PIS), and transparent object detection (TOD).
For COD, we use four datasets: \textit{CHAMELEON} \cite{skurowski2018animal}, \textit{CAMO} \cite{le2019anabranch}, \textit{COD10K} \cite{fan2021concealed}, and \textit{NC4K} \cite{lv2021simultaneously}. Following established practices \cite{he2023weaklysupervised}, our training set comprised 1,000 images from \textit{CAMO} and 3,040 images from \textit{COD10K}, with the remaining images serving as the test set. 
In the PIS task, we assess performance using three benchmarks: \textit{CVC-ColonDB} \cite{tajbakhsh2015automated}, \textit{ETIS} \cite{silva2014toward}, and \textit{Kvasir} \cite{jha2020kvasir}. Our training dataset included 900 images from \textit{Kvasir}, with the remaining images designated for testing. 
For the TOD task, we employed two datasets: \textit{GDD} \cite{mei2020don} and \textit{GSD} \cite{lin2021rich}. The training set consisted of 2,980 images from \textit{GDD} and 3,202 images from \textit{GSD}, with the remaining images forming the test set.

}

\subsection{Weakly-supervised Comparative Evaluation}
{
For all three WSCOS tasks, performance is evaluated using point annotations, following 
WS-SAM~\cite{he2023weaklysupervised}. Additionally, for COD, we further assess performance using scribble annotations, utilizing the scribble data provided in~\cite{gao2022weakly}.}

\noindent\textbf{Camouflaged object detection}.
{
As shown in Table \ref{table:CODQuanti}, our method, SEE, outperforms all competing methods and baselines. 
Notably, while SAM has verified impressive performance in natural scenarios, its performance on the challenging COD task still falls short.
We observe performance gains after fine-tuning SAM with point ($\text{SAM-W}_\text{P}$) and scribble ($\text{SAM-W}_\text{S}$) supervision, yet the results still trail behind our method.
Furthermore, our SEE framework surpasses the previous WS-SAM \cite{he2023weaklysupervised} framework on all metrics with both scribble and point supervision. This highlights our superiority in generating higher-quality pseudo-labels.
To ensure that our improvement over existing WSCOS methods is not solely due to the use of SAM, we integrated SCOD \cite{he2022weakly} and GenSAM~\cite{hu2024relax}, into our SEE framework to leverage additional mask supervision. This resulted in the methods ``SCOD++'' and ``GenSAM++''. Our results show that SCOD++ generally achieves average performance gains of $3.15\%$ and $5.04\%$ across the four datasets with 16 metrics in scribble and point supervision, which are higher than those achieved by WS-SAM ($1.54\%$ and $2.45\%$ in scribble and point settings). Similar performance gains are also observed in the comparison between GenSAM++ and GenSAM+, further validating our method's advantages.
Fig.~\ref{fig:COSQuali} displays the prediction maps with point supervision, demonstrating our advancement. This superior performance is attributed not only to the accurate pseudo-labels produced by our SEE framework, which enhances the segmenter's discrimination capacity, but also to the HGFG module, which adaptively enhances feature coherence through clustering. Hence, our method can alleviate incomplete segmentation and facilitate multi-object segmentation.

\noindent\textbf{Polyp image segmentation}. 
Polyps often share intrinsic consistency with their background organs and thus PIS is often considered highly challenging. As shown in Table \ref{table:MISTOD_Quanti}, our method significantly surpasses the second-best method, WS-SAM, under point supervision. SAM and $\text{SAM-W}_\text{P}$ perform poorly on this task, highlighting their limitations in this challenging task.
To verify the generalization of our framework, we integrated it with SCOD and GenSAM to create SCOD++ and GenSAM++. Our results show that SCOD++ and GenSAM++ improve the average performance of SCOD by $3.54\%$ and $3.73\%$, which is significantly higher than the $1.82\%$ and $1.26\%$ improvement brought by WS-SAM (SCOD+ and GenSAM+). Notably, although incorporating SEE into SCOD and GenSAM enhances their performance, the results still fall short of those achieved by our method alone. 
As illustrated in Fig.~\ref{fig:COSQuali}, our method can better localize those concealed polyps and segment those objects with more accurate boundaries.

\noindent\textbf{Transparent object detection}.
Recognizing transparent objects is an important yet challenging task that enhances the capabilities of vision systems in robotics and automation.
As shown in Table~\ref{table:MISTOD_Quanti}, our method outperforms existing cutting-edge methods for this task. Specifically, our SEE framework surpasses WS-SAM by $2.49\%$ overall. Besides, integrating SCOD and GenSAM with our SEE framework significantly boosts their performance by $8.07\%$ and $2.67\%$, compared to $3.61\%$ and $0.91\%$ improvements achieved by WS-SAM. This verifies the robustness and generalizability of our framework, SEE. As exhibited in Fig.~\ref{fig:COSQuali}, our method not only accurately identifies the positions of concealed objects but also sharpens their boundaries and helps filter out detailed structures. For instance, it effectively delineates the vase in the first row and the border in the second row, ensuring accurate segmentation.}

\subsection{Semi-supervised Comparative Evaluation}
We further evaluate our effectiveness in semi-supervision. In specific, we report results using only 1/8 and 1/16 labeled training data. We adhere to the previous semi-supervised methods~\cite{lee2023saliency,fang2024real} for labeled data selection.

As shown in~\cref{table:CODSemi,table:MISTODSemi}, similar conclusions can be drawn as that made in the weakly-supervised setting, that is our SEE framework achieves optimal results in all three tasks. For example, in camouflaged object detection, our method outperforms the second-best method, EPS~\cite{lee2023saliency}, by $5.43\%$ and $5.56\%$ with 1/8 and 1/16 labeled training data, respectively.
Additionally, we discover that the integration of our SEE framework with existing methods can obtain performance gains. For instance, in the semi-supervision with 1/16 labeled training data, our SEE framework can further improve the performance of existing methods by $3.17\%$ (DTEN++), $3.70\%$ (PGCL++), $2.63\%$ (EPS++), and $2.94\%$ (CoSOD++) in the polyp image segmentation task. Visualization results provided in~\cref{fig:COSQuali-semi} further illustrate the effect of our approach. As shown in~\cref{fig:COSQuali-semi}, our method accurately segments concealed objects, achieving more complete and detailed segmentation outputs.

\subsection{Ablation Study} \vspace{-1mm}
{
We conduct ablation studies on \textit{COD10K} of the COD task in the weak supervision with scribble. }

{\noindent \textbf{Ablation study for SEE}. 
We establish a baseline by using only the mean-teacher framework without SAM and employing the consistency loss for supervision, aiming to explore how the common network can address this extremely challenging problem without the help of the large foundation model. On top of this baseline, we add pseudo-label generation (PLG),  pseudo-label storage ($\text{PLS}_{\text{t}}$), and pseudo-label supervision ($\text{PLS}_{\text{u}}$). Table~\ref{table:AblationSAM} showing adding these components helps improve the performance, thus demonstrating their effectiveness. {Additionally, to explore the potential of SAM2~\cite{ravi2024sam}, a recently proposed state-of-the-art segmentation foundation model, we integrate SAM2 into our SEE framework by directly replacing the originally used SAM with SAM2. As shown in~\cref{table:SAM2}, SAM2 initially provides a performance boost; however, these gains become marginal when our complete pseudo-label generation strategy is fully applied. This observation underscores the robustness and effectiveness of our framework.}} 

{\noindent \textbf{Ablation study for HGFG}. We evaluate the effectiveness of HGFG by conducting several ablation studies: completely removing the HGFG module, replacing the proposed feature grouping (FG) with slot attention (SA)\cite{locatello2020object}, removing the hybrid-granularity strategy, corresponding to w/o RK2, and substituting the weighted gate mechanism (WGM) with a fixed coefficient (FC). As shown in Table~\ref{table:AblationMFG}, our proposed designs outperform these alternative configurations. 

{\noindent \textbf{Ablation studies in pseudo-label generation}. We assess how specific components contribute to segmentation in pseudo-label generation, including various augmentation strategies and the effect of MARF and MDPE. First, we investigate how augmentations affect segmentation. As shown in~\cref{table:AblationPLG}, compared to augmenting $K$ times using strong augmentations~\cite{cubuk2020randaugment}, or using strong and weak augmentations $K/2$ times each, solely applying weak augmentations yields the best performance. Besides, we verify the effect of MARF, which benefits both the teacher model and SAM by enhancing the quality of pseudo-labels, thus improving segmentation performance. Finally, we explore which prompt combination best facilitates segmentation. Prom. \#1 to \#3 in~\cref{table:AblationPLG} represent ``only using point prompts'', ``using point and box prompts'', and ``using point and mask prompts''. As illustrated in~\cref{table:AblationPLG}, the results from Prom. \#1 to \#3 confirm that point, box, and mask prompts all aid segmentation. 
{ Additionally, we explore three prompt selection strategies (Prompts \#4 to \#6): Prom. \#4 is a staged selection strategy that divides the training process into three phases (epochs 1–40, 41–80, and 81–120), applying point, box, and mask prompts, respectively, in each phase. Prom. \#5 is a fusion strategy where three separate prompts guide the network independently, and their resulting segmentation outputs are fused. While Prom. \#6 is a more sophisticated strategy that generates six sets of segmentation masks from all possible prompt combinations. These masks are then fused into six intermediate masks, denoted as $\{\tilde{\textbf{M}}_t^k\}_{k=1}^{6}$. The six fused masks are updated in an optimal pool, where they compete with previously stored optimal masks to obtain the best mask at each epoch. This strategy is considered a complete and adaptive approach to prompt selection, as it dynamically selects the most effective prompt combination at each epoch. The lower performance of Prompts \#4 and \#5, combined with the comparable results of Prom. \#6—a significantly more complex strategy—suggests that our multi-density prompts collaboratively guide SAM more comprehensively and efficiently.
}
Lastly, we investigate whether further training our framework with the coarse mask generated by the teacher model yields performance gains. We find that this approach decreases performance. This decline is due to two factors: (1) our MDPE strategy effectively extracts valuable information from coarse masks, enabling SAM to generate higher-quality pseudo-labels, and (2) the mean-teacher framework produces low-quality masks during the early training phases, which can mislead the framework.

\begin{table*}[t]
\begin{minipage}[c]{0.46\textwidth}
		\centering
		\setlength{\abovecaptionskip}{0cm}
		\caption{Effect of HGFG. 
  $(0,0,0,1)$ means that HGFG only exists in the $4^{th}$ stage, and so on for others.}
		\resizebox{1\columnwidth}{!}{
			\setlength{\tabcolsep}{1mm}
			\begin{tabular}{l|ccccc}
				\toprule
				Metrics & (0,0,0,0)& \cellcolor{c2!20}(0,0,0,1) &(0,0,1,1) & (0,1,1,1) & (1,1,1,1)\\ \midrule
				$M$~$\downarrow$       & 0.043          &\cellcolor{c2!20} 0.036          & 0.036          & 0.036 & \textbf{0.035}         \\
				$F_\beta$~$\uparrow$ & 0.688          &\cellcolor{c2!20} \textbf{0.729}          & 0.726          & 0.728 & \textbf{0.729} \\
				$E_\phi$~$\uparrow$       & 0.861          &\cellcolor{c2!20} 0.883          & 0.882          & \textbf{0.887} &  0.883         \\
				$S_\alpha$~$\uparrow$       & 0.785          &\cellcolor{c2!20} \textbf{0.807}          & 0.806 & 0.803   & 0.805       \\ \bottomrule
		\end{tabular}}\label{table:GMFA_Combination}
		\vspace{-0.3cm}
  \end{minipage}
\begin{minipage}[c]{0.53\textwidth}
\centering
	\setlength{\abovecaptionskip}{0cm}
	\caption{Parameter analysis on $T$ and $(N_1,N_2)$ in HGFG.}
	\resizebox{\columnwidth}{!}{
		\setlength{\tabcolsep}{1mm}
		\begin{tabular}{c|cccc|cccc}
			\toprule
			\multicolumn{1}{l|}{\multirow{2}{*}{Metrics}} & \multicolumn{4}{c|}{$T$} & \multicolumn{4}{c}{$(N_1,N_2)$} \\\cline{2-9}
			\multicolumn{1}{l|}{}   & 1     & 2     & \cellcolor{c2!20}3              & 4              & \cellcolor{c2!20}(2,4)          & (2,8)          & (4,8) & (2,4,8)        \\ \midrule
			$M$~$\downarrow$   & 0.039 & 0.039 & \cellcolor{c2!20}\textbf{0.038} & \textbf{0.038} & \cellcolor{c2!20}\textbf{0.038} & \textbf{0.038} & 0.039 & \textbf{0.038} \\
			$F_\beta$~$\uparrow$   & 0.706 & 0.715 & \cellcolor{c2!20}0.719          & \textbf{0.720} & \cellcolor{c2!20}0.719          & 0.714          & 0.711 & \textbf{0.721} \\
			$E_\phi$~$\uparrow$ & 0.862 & 0.872 & \cellcolor{c2!20}\textbf{0.878} & 0.876          & \cellcolor{c2!20}\textbf{0.878} & 0.875          & 0.873 & \textbf{0.878} \\
			$S_\alpha$~$\uparrow$   & 0.794 & 0.800 & \cellcolor{c2!20}0.803          & \textbf{0.805} & \cellcolor{c2!20}\textbf{0.803} & 0.802          & 0.799 & 0.802        \\ \bottomrule                        
	\end{tabular}}\label{table:ParameterAnalysis}
	\vspace{-0.3cm}
    \end{minipage}
\end{table*}

\begin{figure*}[h]
	\centering
	\setlength{\abovecaptionskip}{-0.2cm}
	\begin{center}
		\includegraphics[width=\linewidth]{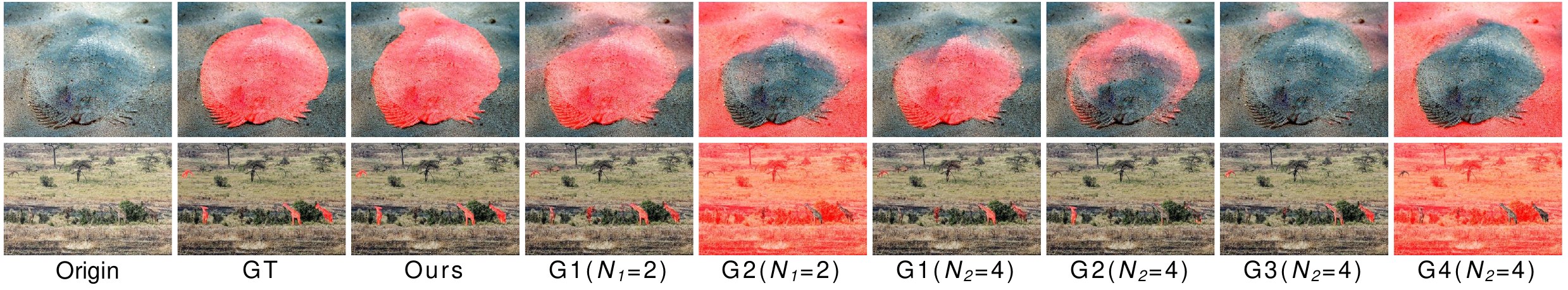}
	\end{center}
	\caption{Visualization of feature similarity among group prototypes and grid features for $N_1=2$ and $N_2=4$, where G1 denotes group 1 and so on. For each pixel point, the darker red color indicates a higher correlation.}
	\label{fig:GroupVisualization}
	\vspace{-0.2cm}
\end{figure*}
\begin{table*}[t]
	\centering
	\setlength{\abovecaptionskip}{0.1cm}
	\caption{Generalization of HGFG in \textit{COD10K}. The suffix $^\dagger$ indicates that HFGF is combined with the respective method. 
    }\label{table:GeneralizationMFG} \vspace{-1mm}
		\begin{subtable}{.3055\textwidth}
		\centering
		\setlength{\abovecaptionskip}{0cm}
		\caption{Weakly-supervised COS.
		}
		\resizebox{\columnwidth}{!}{
			\setlength{\tabcolsep}{0.6mm}
			\begin{tabular}{c|cc|cc}
				\toprule
				Metrics  & TEL~\cite{liang2022tree} &\cellcolor{c2!20}  TEL$^\dagger$ & SCOD~\cite{he2022weakly} &\cellcolor{c2!20}  SCOD$^\dagger$ \\ \midrule
				$M$~$\downarrow$  & 0.057    &\cellcolor{c2!20}  0.054 & 0.049   &\cellcolor{c2!20}  0.046 \\
				$F_\beta$~$\uparrow$ & 0.633    &\cellcolor{c2!20}  0.642  & 0.637   &\cellcolor{c2!20}  0.672 \\
				$E_\phi$~$\uparrow$ & 0.826    &\cellcolor{c2!20}  0.833   & 0.832   &\cellcolor{c2!20}  0.840 \\
				$S_\alpha$~$\uparrow$     & 0.724    &\cellcolor{c2!20}  0.731  & 0.733   &\cellcolor{c2!20}  0.744 \\ \bottomrule
		\end{tabular}}\label{table:GeneralizationMFGCOSW}
	\end{subtable} 
 \begin{subtable}{.318\textwidth}
		\centering
		\setlength{\abovecaptionskip}{0cm}
		\caption{Semi-supervised COS.
		}
		\resizebox{\columnwidth}{!}{
			\setlength{\tabcolsep}{0.6mm}
			\begin{tabular}{c|cc|cc}
				\toprule
				Metrics  & EPS~\cite{lee2023saliency} & \cellcolor{c2!20} EPS$^\dagger$ & CoSOD~\cite{chakraborty2024unsupervised} & \cellcolor{c2!20} CoSOD$^\dagger$ \\ \midrule
				$M$~$\downarrow$  & 0.047    &\cellcolor{c2!20}  0.045 & 0.046   &\cellcolor{c2!20}  0.043 \\
				$F_\beta$~$\uparrow$ & 0.659    &\cellcolor{c2!20}  0.681  & 0.673   &\cellcolor{c2!20}  0.696 \\
				$E_\phi$~$\uparrow$ & 0.806    &\cellcolor{c2!20}  0.829   & 0.813   &\cellcolor{c2!20}  0.835 \\
				$S_\alpha$~$\uparrow$     & 0.761    &\cellcolor{c2!20}  0.773  & 0.767   &\cellcolor{c2!20}  0.776 \\ \bottomrule
		\end{tabular}}\label{table:GeneralizationMFGCOSS}
	\end{subtable} 
 \begin{subtable}{.3663\textwidth}
		\centering
		\setlength{\abovecaptionskip}{0cm}
		\caption{Fully-supervised COS.
		}
		\resizebox{\columnwidth}{!}{
			\setlength{\tabcolsep}{0.6mm}
			\begin{tabular}{c|cc|cc}
				\toprule
				Metrics  & FGANet~\cite{zhaiexploring} & \cellcolor{c2!20} FGANet$^\dagger$ & FEDER~\cite{He2023Camouflaged} &\cellcolor{c2!20}  FEDER$^\dagger$ \\ \midrule
				$M$~$\downarrow$  & 0.032    &\cellcolor{c2!20}  0.030 & 0.032   &\cellcolor{c2!20}  0.030 \\
				$F_\beta$~$\uparrow$ & 0.708    &\cellcolor{c2!20}  0.725  & 0.715   &\cellcolor{c2!20}  0.732 \\
				$E_\phi$~$\uparrow$ & 0.894    &\cellcolor{c2!20}  0.917   & 0.892   &\cellcolor{c2!20}  0.911 \\
				$S_\alpha$~$\uparrow$     & 0.801    &\cellcolor{c2!20}  0.813  & 0.810   &\cellcolor{c2!20}  0.819 \\ \bottomrule
		\end{tabular}}\label{table:GeneralizationMFGCOSF}
	\end{subtable}
    \\ \vspace{1mm}
 \begin{subtable}{.3145\textwidth}
		\centering
		\setlength{\abovecaptionskip}{0cm}
		\caption{Weakly-supervised SOD.
		}
		\resizebox{\columnwidth}{!}{
			\setlength{\tabcolsep}{0.6mm}
			\begin{tabular}{c|cc|cc}
				\toprule
				Metrics  & TEL~\cite{liang2022tree} &\cellcolor{c2!20}  TEL$^\dagger$ & SCOD~\cite{he2022weakly} &\cellcolor{c2!20}  SCOD$^\dagger$ \\ \midrule
				$M$~$\downarrow$  & 0.045    &\cellcolor{c2!20}  0.041 & 0.046   &\cellcolor{c2!20}  0.041 \\
				$F_\beta$~$\uparrow$ & 0.842    &\cellcolor{c2!20}  0.869  & 0.830   &\cellcolor{c2!20}  0.862 \\
				$E_\phi$~$\uparrow$ & 0.901    &\cellcolor{c2!20}  0.920   & 0.892   &\cellcolor{c2!20}  0.918 \\
				$S_\alpha$~$\uparrow$     & 0.863    &\cellcolor{c2!20}  0.874  & 0.855   &\cellcolor{c2!20}  0.872 \\ \bottomrule
		\end{tabular}}\label{table:GeneralizationMFGSODW}
		\vspace{-0.4cm}
	\end{subtable} 
 	\begin{subtable}{.327\textwidth}
		\centering
		\setlength{\abovecaptionskip}{0cm}
		\caption{Semi-supervised SOD.
		}
		\resizebox{\columnwidth}{!}{
			\setlength{\tabcolsep}{0.6mm}
			\begin{tabular}{c|cc|cc}
				\toprule
				Metrics  & EPS~\cite{lee2023saliency} & \cellcolor{c2!20} EPS$^\dagger$ & CoSOD~\cite{chakraborty2024unsupervised} & \cellcolor{c2!20} CoSOD$^\dagger$ \\ \midrule
				$M$~$\downarrow$  & 0.058    &\cellcolor{c2!20}  0.053 & 0.055   &\cellcolor{c2!20}  0.050 \\
				$F_\beta$~$\uparrow$ & 0.806    &\cellcolor{c2!20}  0.819  & 0.817   &\cellcolor{c2!20}  0.831 \\
				$E_\phi$~$\uparrow$ & 0.852    &\cellcolor{c2!20}  0.873   & 0.870   &\cellcolor{c2!20}  0.887 \\
				$S_\alpha$~$\uparrow$     & 0.818    &\cellcolor{c2!20}  0.827  & 0.826   &\cellcolor{c2!20}  0.840 \\ \bottomrule
		\end{tabular}}\label{table:GeneralizationMFGSODS}
		\vspace{-0.4cm}
	\end{subtable} 
	\begin{subtable}{.348\textwidth}
		\centering
		\setlength{\abovecaptionskip}{0cm}
		\caption{Fully-supervised SOD.
		}
		\resizebox{\columnwidth}{!}{
			\setlength{\tabcolsep}{0.6mm}
			\begin{tabular}{c|cc|cc}
				\toprule
				Metrics  & PGNet~\cite{xie2022pyramid} & \cellcolor{c2!20} PGNet$^\dagger$ & MENet~\cite{wang2023pixels} &\cellcolor{c2!20}  MENet$^\dagger$ \\ \midrule
				$M$~$\downarrow$  & 0.027    &\cellcolor{c2!20}  0.025 & 0.028   &\cellcolor{c2!20}  0.025 \\
				$F_\beta$~$\uparrow$ & 0.903    &\cellcolor{c2!20}  0.923  & 0.895   &\cellcolor{c2!20}  0.919 \\
				$E_\phi$~$\uparrow$ & 0.922    &\cellcolor{c2!20}  0.941   & 0.937   &\cellcolor{c2!20}  0.949 \\
				$S_\alpha$~$\uparrow$     & 0.911    &\cellcolor{c2!20}  0.920  & 0.905   &\cellcolor{c2!20}  0.917 \\ \bottomrule
		\end{tabular}}\label{table:GeneralizationMFGSODF}
		\vspace{-0.4cm}
	\end{subtable}
\end{table*}	

\noindent \textbf{Explorations of multi-density prompt extraction}. Here we explore how to best extract prompts to facilitate segmentation in MDPE. For point prompts, ``w/o NBS'' denotes the random selection of all points, while ``NBS \#1'' denotes randomly selecting one foreground and one background point within each block. \cref{table:AblationPromptExtraction} demonstrate that using NBS, selecting center points as foreground prompts, and selecting relatively farthest points as background prompts lead to performance improvements. Additionally, we experimented on the COD training set to investigate the effect of the number $C$ of key points sampled on the quality of the generated pseudo-labels. To ensure comprehensiveness and validity, we performed ten trials and calculated the average $F_\beta$ score. Fig.~\ref{fig:PointNumber} shows that NBS results in higher quality and more stable pseudo-labels, with a smaller standard deviation. Moreover, as the quality of the generated pseudo-labels stabilizes when $C=9$, we set the point number $C$ to 9 in this paper. For the box prompt, we focus on the expanding manner. ``w/o Expand.'' refers to using the original minimal bounding box, while ``Expand. \#1'' and ``Expand. \#2'' denote expanding the bounding box equally in four directions with the maximum or minimal expanding coefficient. The results confirm that expanding the box with the coefficient corresponding to each direction achieves the best performance. Finally, experiments on the mask prompt demonstrate the effectiveness of our filtering strategy.

\noindent \textbf{Ablation studies in pseudo-label storage and supervision}.
For pseudo-label storage, ``Stor. \#1'' and ``Stor. \#2'' refer to updating the optimal pool if any metric or all metrics of the new label outperform the existing labels. As verified in~\cref{table:AblationPLS}, our update rule—requiring at least two metrics to achieve better performance—yields the best result. Additionally, ``Stor. \#3'' and ``Stor. \#4'' represent ablation studies of the comparison rule, corresponding to "using only $U_a$" and "using $U_a$ + $U_r$." The experiments confirm that incorporating all metrics contributes to improved segmentation performance. This is further validated by~\cref{fig:PLSt}, which illustrates the quality of the generated pseudo-labels over epochs (with an interval of 10 epochs), highlighting that our comparison rule yields pseudo-labels with the highest $F_\beta$ score. This suggests that our comparison rule, designed to select the most informative and useful pseudo-labels, effectively aligns to select the highest-quality labels. For pseudo-label supervision, the breakdown studies, where ``Sup. \#1 to \#3'' correspond to ``using all generated pseudo-labels without selection'', `` using only $U_a$ for label selection'', and ``omitting the pixel-level weighting strategy''. \cref{table:AblationPLS} highlights the effectiveness of our proposed strategies in pseudo-label supervision, enabling the student model to receive more precise supervision.

}

}

{\noindent\textbf{Number of augmented views $K$ and stored best labels $B$}.
Table~\ref{table:ParameterAnalysisKB} shows a similar trend for both the number of augmented views $K$ and stored best labels $B$: while increasing these values initially improves performance, the effect diminishes with further increases. For the augmented views, this may occur because the chosen number already fully exploits the benefits of augmentation. For the stored best labels, once $B$ reaches 3, the improved diversity is balanced by the reduced stability, as additional pseudo-labels increase the risk of introducing erroneous supervision. Therefore, we set $K=12$ and $B=3$ to balance performance and efficiency.

\noindent\textbf{Selection rules in image-level selection}. Table~\ref{table:ParameterAnalysisKB} indicates that the best settings for the absolute uncertainty threshold and relative uncertainty threshold are $\tau_{a}=0.1$ and $\tau_{r}=0.5$, respectively. Notably, our method demonstrates robustness and is not highly sensitive to variations in these parameters.   

\noindent \textbf{HGFG in different stages}. To balance performance and efficiency, we apply HGFG only to the bottom layer, yielding $\dot{\mathbf{F}}_i^4$, as the semantic information 
from the bottom layer is most conducive for HGFG to exploit coherence. We verify this by applying the module at different stages. 
As shown in Table~\ref{table:GMFA_Combination}, HGFG gets promising results when applied solely in the $4^{th}$ stage, while extra HGFGs offer limited gains.

\noindent\textbf{Hyperparameters in HGFG}. 
As shown in Table~\ref{table:ParameterAnalysis}, HGFG achieves the best results when the iteration number $T$ is set to 3 and the group and scale settings $(N_1, N_2)$ are configured as $(2,4)$. $(2,4,8)$ needs a third-order RK structure, which introduces extra computational overhead with limited benefit. Therefore, we select the RK2 structure with $(N_1, N_2)=(2,4)$ to strike a balance between performance and efficiency.}

\subsection{Further Analysis}

{\noindent \textbf{Visualization of feature similarity in HGFG}. To investigate how the HGFG module enhances complete segmentation and multi-object segmentation, we visualize the similarity between group prototypes and grid features in Fig.~\ref{fig:GroupVisualization}, where $N_1=2$ and $N_2=4$ denote clustering into two groups and four groups, respectively. 
For complete segmentation, we observe that each individual group can identify local correlations within a single object, aiding in comprehensive segmentation through the combined efforts of multiple groups. Additionally, we find that each group can achieve global coherence across multiple objects by aggregating similar features, thereby improving accuracy in multi-object segmentation.
Fig.~\ref{fig:GroupVisualization} demonstrates that clustering prototypes can learn to aggregate similar features, achieving more fine-grained aggregation at higher granularity levels. Consequently, our HGFG effectively extracts critical cues by integrating hybrid-granularity grouping features.}

\begin{table*}[ht]
\begin{minipage}[c]{\textwidth}
	\centering
	\setlength{\abovecaptionskip}{0cm}
	\caption{Restuls on the weakly-supervised SOD task with scribble and point supervision.}
	\resizebox{\columnwidth}{!}{
		\setlength{\tabcolsep}{0.8mm}
		\begin{tabular}{l|cccc|cccc|cccc|cccc|cccc} 
			\toprule[1.5pt]
			\multicolumn{1}{l|}{}& \multicolumn{4}{c|}{\textit{DUT-OMRON} }& \multicolumn{4}{c|}{\textit{DUTS-test} }& \multicolumn{4}{c|}{\textit{ECSSD} }& \multicolumn{4}{c|}{\textit{HKU-IS} }& \multicolumn{4}{c}{\textit{PASCAL-S} }\\ \cline{2-21}
			\multicolumn{1}{l|}{\multirow{-2}{*}{Methods}} & {\cellcolor{gray!40}$M$~$\downarrow$} &{\cellcolor{gray!40}$F_\beta$~$\uparrow$} &{\cellcolor{gray!40}$E_\phi$~$\uparrow$} & \multicolumn{1}{c|}{\cellcolor{gray!40}$S_\alpha$~$\uparrow$}& {\cellcolor{gray!40}$M$~$\downarrow$} &{\cellcolor{gray!40}$F_\beta$~$\uparrow$} &{\cellcolor{gray!40}$E_\phi$~$\uparrow$} & \multicolumn{1}{c|}{\cellcolor{gray!40}$S_\alpha$~$\uparrow$}& {\cellcolor{gray!40}$M$~$\downarrow$} &{\cellcolor{gray!40}$F_\beta$~$\uparrow$} &{\cellcolor{gray!40}$E_\phi$~$\uparrow$} & \multicolumn{1}{c|}{\cellcolor{gray!40}$S_\alpha$~$\uparrow$}& {\cellcolor{gray!40}$M$~$\downarrow$} &{\cellcolor{gray!40}$F_\beta$~$\uparrow$} &{\cellcolor{gray!40}$E_\phi$~$\uparrow$} & \multicolumn{1}{c|}{\cellcolor{gray!40}$S_\alpha$~$\uparrow$}& {\cellcolor{gray!40}$M$~$\downarrow$} &{\cellcolor{gray!40}$F_\beta$~$\uparrow$} &{\cellcolor{gray!40}$E_\phi$~$\uparrow$} & \multicolumn{1}{c}{\cellcolor{gray!40}$S_\alpha$~$\uparrow$}\\ \midrule
			\multicolumn{21}{c}{Scribble Supervision} \\ \midrule
			SAM~\cite{kirillov2023segment}                                         & 0.077                                 & 0.696                                 & 0.729                                 & 0.734                                 & 0.103                                 & 0.713                                 & 0.741                                 & 0.737                                 & 0.103                                 & 0.728                                 & 0.741                                 & 0.744                                 & 0.119                                 & 0.651                                 & 0.657                                 & 0.657                                 & 0.116                                 & 0.695                                 & 0.716                                 & 0.712                                 \\
			$\text{SAM-W}_\text{S}$~\cite{chen2023sam}                                         & 0.057 & 0.786 & 0.875 & 0.826 & 0.045 & 0.820 & 0.849 & 0.785 & 0.060 & 0.837 & 0.896 & 0.875 & 0.048 & 0.858 & 0.930 & 0.815 & 0.070 & 0.795 & 0.808 & 0.799 \\
			SCWS~\cite{yu2021structure}                                       & 0.060                                 & 0.764                                 & 0.862                                 & 0.812                                 & 0.049                                 & 0.823                                 & 0.890                                 & 0.841                                 & 0.049                                 & 0.900                                 & 0.908                                 & 0.882                                 & 0.038                                 & 0.896                                 & 0.938                                 & 0.882                                 & 0.077                                 & 0.823                                 & 0.846                                 & 0.813                                 \\
			TEL~\cite{liang2022tree}                                           & 0.058                                 & 0.753                                 & 0.864                                 & 0.818                                 & 0.045                                 & {{0.842}}                                 & 0.901                                 & {{0.863}}                                 & 0.039                                 & 0.921                                 & {{0.935}}                                 & 0.907                                 & 0.033                                 & 0.934                                 & 0.944                                 & {{0.906}}                                 & 0.065                                 & 0.834                                 & {{0.883}}                                 & {{0.858}}                                 \\
			SCOD~\cite{he2022weakly}                                         & 0.061                                 & 0.762                                 & 0.865                                 & 0.814                                 & 0.046                                 & 0.830                                 & 0.892                                 & 0.855                                 & 0.042                                 & 0.917                                 & 0.922                                 & 0.910                                 & 0.035                                 & 0.933                                 & 0.942                                 & 0.900                                 & 0.069                                 & 0.828                                 & 0.864                                 & 0.847                                 \\
   GenSAM~\cite{hu2024relax}  & 0.068 & 0.697 & 0.741 & 0.706 & 0.077 & 0.758 & 0.780 & 0.745 & 0.078 & 0.857 & 0.830 & 0.806 & 0.068 & 0.815 & 0.772 & 0.747 & 0.092 & 0.750 & 0.751 & 0.743 \\
			SCOD+~\cite{he2023weaklysupervised} & 0.057                                 & 0.774                                 & 0.872                                 & 0.822                                 & 0.042                                 & 0.841                                 & 0.907                                 & 0.862                                 & 0.038                                 & 0.930                                 & 0.935                                 & 0.917                                 & 0.032                                 & 0.935                                 & 0.947                                 & 0.906                                 & 0.062                                 & 0.838                                 & 0.881                                 & 0.855                                 \\
   GenSAM+~\cite{he2023weaklysupervised}  & 0.062 & 0.718 & 0.765 & 0.715 & 0.072 & 0.766 & 0.795 & 0.749 & 0.073 & 0.869 & 0.845 & 0.813 & 0.062 & 0.826 & 0.785 & 0.751 & 0.088 & 0.763 & 0.758 & 0.749 \\
			WS-SAM~\cite{he2023weaklysupervised} & {\color[HTML]{00B0F0} \textbf{0.050}} & {\color[HTML]{00B0F0} \textbf{0.798}} & {\color[HTML]{00B0F0} \textbf{0.885}} & {\color[HTML]{00B0F0} \textbf{0.844}} & {\color[HTML]{00B0F0} \textbf{0.034}} & {\color[HTML]{00B0F0} \textbf{0.870}} & {\color[HTML]{00B0F0} \textbf{0.931}} & {\color[HTML]{00B0F0} \textbf{0.890}} & {\color[HTML]{00B0F0} \textbf{0.032}} & {\color[HTML]{00B0F0} \textbf{0.942}} & {\color[HTML]{00B0F0} \textbf{0.955}} & {\color[HTML]{00B0F0} \textbf{0.926}} & {\color[HTML]{00B0F0} \textbf{0.026}} & 0.935                                 & {\color[HTML]{00B0F0} \textbf{0.958}} & {\color[HTML]{FF0000} \textbf{0.922}} & {\color[HTML]{00B0F0} \textbf{0.055}} & {\color[HTML]{00B0F0} \textbf{0.869}} & {\color[HTML]{00B0F0} \textbf{0.914}} & {\color[HTML]{00B0F0} \textbf{0.871}} \\
   \rowcolor{c2!20}SCOD++& 0.054                                 & 0.785                                 & 0.882                                 & 0.827                                 & 0.039                                 & 0.849                                 & 0.912                                 & 0.866                                 & 0.035                                 & 0.937                                 & 0.942                                 & 0.922                                 & 0.030                                 & {\color[HTML]{FF0000} \textbf{0.938}} & 0.951                                 & 0.912                                 & 0.060                                 & 0.843                                 & 0.887                                 & 0.860                                 \\
   \rowcolor{c2!20} GenSAM++  & 0.055 & 0.737 & 0.782 & 0.730 & 0.064 & 0.779 & 0.816 & 0.761 & 0.065 & 0.887 & 0.868 & 0.821 & 0.054 & 0.841 & 0.802 & 0.760 & 0.082 & 0.778 & 0.773 & 0.758 \\
   \rowcolor{c2!20}SEE (Ours)& {\color[HTML]{FF0000} \textbf{0.047}} & {\color[HTML]{FF0000} \textbf{0.809}} & {\color[HTML]{FF0000} \textbf{0.893}} & {\color[HTML]{FF0000} \textbf{0.849}} & {\color[HTML]{FF0000} \textbf{0.032}} & {\color[HTML]{FF0000} \textbf{0.877}} & {\color[HTML]{FF0000} \textbf{0.938}} & {\color[HTML]{FF0000} \textbf{0.895}} & {\color[HTML]{FF0000} \textbf{0.031}} & {\color[HTML]{FF0000} \textbf{0.944}} & {\color[HTML]{FF0000} \textbf{0.958}} & {\color[HTML]{FF0000} \textbf{0.928}} & {\color[HTML]{FF0000} \textbf{0.025}} & {\color[HTML]{00B0F0} \textbf{0.936}} & {\color[HTML]{FF0000} \textbf{0.960}} & {\color[HTML]{FF0000} \textbf{0.922}} & {\color[HTML]{FF0000} \textbf{0.053}} & {\color[HTML]{FF0000} \textbf{0.872}} & {\color[HTML]{FF0000} \textbf{0.918}} & {\color[HTML]{FF0000} \textbf{0.873}}\\
   \midrule
			\multicolumn{21}{c}{Point Supervision} \\ \midrule
			SAM~\cite{kirillov2023segment} & 0.077 & 0.696                                 & 0.729                                 & 0.734                                 & 0.103                                 & 0.713                                 & 0.741                                 & 0.737                                 & 0.103                                 & 0.728                                 & 0.741                                 & 0.744                                 & 0.119                                 & 0.651                                 & 0.657                                 & 0.657                                 & 0.116                                 & 0.695                                 & 0.716                                 & 0.712                                 \\
   $\text{SAM-W}_\text{P}$~\cite{chen2023sam}                                         & 0.072                                 & 0.725 & 0.746 & 0.763 & 0.083 & 0.775 & 0.796 & 0.768 & 0.085 & 0.794 & 0.793 & 0.781 & 0.071 & 0.741 & 0.718 & 0.702 & 0.096 & 0.743 & 0.747 & 0.735 \\
			SCWS~\cite{yu2021structure}                                       & 0.076                                 & 0.668                                 & 0.720                                 & 0.720                                 & 0.080                                 & 0.747                                 & 0.769                                 & 0.752                                 & 0.089                                 & 0.815                                 & 0.828                                 & 0.797                                 & 0.076                                 & 0.801                                 & 0.773                                 & 0.738                                 & 0.098                                 & 0.752                                 & 0.742                                 & 0.733                                 \\
			TEL~\cite{liang2022tree}                                           & 0.073                                 & 0.699                                 & 0.736                                 & 0.717                                 & 0.074                                 & 0.758                                 & 0.775                                 & 0.761                                 & 0.078                                 & 0.857                                 & 0.840                                 & 0.813                                 & 0.071                                 & 0.817                                 & 0.785                                 & 0.764                                 & 0.093                                 & 0.761                                 & 0.760                                 & 0.748                                 \\
			SCOD~\cite{he2022weakly}                                         & 0.078                                 & 0.687                                 & 0.729                                 & 0.705                                 & 0.079                                 & 0.741                                 & 0.769                                 & 0.739                                 & 0.082                                 & 0.834                                 & 0.809                                 & 0.792                                 & 0.074                                 & 0.808                                 & 0.769                                 & 0.753                                 & 0.097                                 & 0.749                                 & 0.733                                 & 0.729                                 \\
   GenSAM~\cite{hu2024relax}  & 0.068   & 0.697 & 0.741 & 0.706 & 0.077 & 0.758 & 0.780 & 0.745 & 0.078 & 0.857 & 0.830 & 0.806 & 0.068 & 0.815 & 0.772 & 0.747 & 0.092 & 0.750 & 0.751 & 0.743 \\
			SCOD+~\cite{he2023weaklysupervised} &  0.075  & 0.694 &0.734 &0.712 & 0.076 &0.766 &0.778 &0.747 & 0.080 & 0.851 & 0.820 &0.800 &{{0.071}} &0.816 &0.781 &0.758 & 0.094 &0.758 &0.747 &0.741   \\
   GenSAM+~\cite{he2023weaklysupervised}  & 0.067                                 & 0.703 & 0.752 & 0.713 & 0.072 & 0.766 & 0.787 & 0.749 & 0.075 & 0.863 & 0.838 & 0.811 & 0.066 & 0.820 & 0.780 & 0.752 & 0.089 & 0.760 & 0.762 & 0.746 \\
			WS-SAM~\cite{he2023weaklysupervised} & {\color[HTML]{00B0F0} \textbf{0.064}} & {\color[HTML]{00B0F0} \textbf{0.741}} & {\color[HTML]{00B0F0} \textbf{0.828}} & {\color[HTML]{00B0F0} \textbf{0.790}} & {\color[HTML]{00B0F0} \textbf{0.052}} & {\color[HTML]{00B0F0} \textbf{0.821}} & {\color[HTML]{00B0F0} \textbf{0.880}} & {\color[HTML]{00B0F0} \textbf{0.837}} & {\color[HTML]{00B0F0} \textbf{0.048}} & {\color[HTML]{00B0F0} \textbf{0.910}} & {\color[HTML]{00B0F0} \textbf{0.920}} & {\color[HTML]{00B0F0} \textbf{0.891}} & {\color[HTML]{00B0F0} \textbf{0.048}} & {\color[HTML]{00B0F0} \textbf{0.884}} & {\color[HTML]{00B0F0} \textbf{0.903}} & {\color[HTML]{00B0F0} \textbf{0.863}} & {\color[HTML]{00B0F0} \textbf{0.075}} & {\color[HTML]{00B0F0} \textbf{0.814}} & {\color[HTML]{00B0F0} \textbf{0.865}} & {\color[HTML]{00B0F0} \textbf{0.824}} \\
      \rowcolor{c2!20}SCOD++& 0.073                                 & 0.706                                 & 0.738                                 & 0.714                                 & 0.073                                 & 0.775                                 & 0.791                                 & 0.752                                 & 0.072                                 & 0.863                                 & 0.833                                 & 0.812                                 & 0.064                                 & 0.835                                 & 0.792                                 & 0.766                                 & 0.089                                 & 0.767                                 & 0.755                                 & 0.749                                 \\
      \rowcolor{c2!20}GenSAM++ & {\color[HTML]{00B0F0} \textbf{0.064}} & 0.720 & 0.766 & 0.726 & 0.065 & 0.775 & 0.796 & 0.755 & 0.071 & 0.879 & 0.847 & 0.817 & 0.062 & 0.837 & 0.793 & 0.759 & 0.085 & 0.773 & 0.767 & 0.748 \\
   \rowcolor{c2!20}SEE (Ours)& {\color[HTML]{FF0000} \textbf{0.060}} & {\color[HTML]{FF0000} \textbf{0.752}} & {\color[HTML]{FF0000} \textbf{0.836}} & {\color[HTML]{FF0000} \textbf{0.797}} & {\color[HTML]{FF0000} \textbf{0.050}} & {\color[HTML]{FF0000} \textbf{0.826}} & {\color[HTML]{FF0000} \textbf{0.892}} & {\color[HTML]{FF0000} \textbf{0.839}} & {\color[HTML]{FF0000} \textbf{0.045}} & {\color[HTML]{FF0000} \textbf{0.918}} & {\color[HTML]{FF0000} \textbf{0.922}} & {\color[HTML]{FF0000} \textbf{0.895}} & {\color[HTML]{FF0000} \textbf{0.046}} & {\color[HTML]{FF0000} \textbf{0.891}} & {\color[HTML]{FF0000} \textbf{0.908}} & {\color[HTML]{FF0000} \textbf{0.867}} & {\color[HTML]{FF0000} \textbf{0.072}} & {\color[HTML]{FF0000} \textbf{0.819}} & {\color[HTML]{FF0000} \textbf{0.871}} & {\color[HTML]{FF0000} \textbf{0.828}} \\\bottomrule[1.5pt]
	\end{tabular}}
	\label{table:SODQuanti}
	\vspace{-0.3cm}
 \end{minipage}
\end{table*}

\begin{table*}[ht]
 \begin{minipage}[c]{\textwidth}
	\centering
	\setlength{\abovecaptionskip}{0cm}
	\caption{Restuls on the semi-supervised SOD task.}
	\resizebox{\columnwidth}{!}{
		\setlength{\tabcolsep}{0.8mm}
		\begin{tabular}{l|cccc|cccc|cccc|cccc|cccc} 
			\toprule[1.5pt]
			\multicolumn{1}{l|}{}& \multicolumn{4}{c|}{\textit{DUT-OMRON} }& \multicolumn{4}{c|}{\textit{DUTS-test} }& \multicolumn{4}{c|}{\textit{ECSSD} }& \multicolumn{4}{c|}{\textit{HKU-IS} }& \multicolumn{4}{c}{\textit{PASCAL-S} }\\ \cline{2-21}
			\multicolumn{1}{l|}{\multirow{-2}{*}{Methods}} & {\cellcolor{gray!40}$M$~$\downarrow$} &{\cellcolor{gray!40}$F_\beta$~$\uparrow$} &{\cellcolor{gray!40}$E_\phi$~$\uparrow$} & \multicolumn{1}{c|}{\cellcolor{gray!40}$S_\alpha$~$\uparrow$}& {\cellcolor{gray!40}$M$~$\downarrow$} &{\cellcolor{gray!40}$F_\beta$~$\uparrow$} &{\cellcolor{gray!40}$E_\phi$~$\uparrow$} & \multicolumn{1}{c|}{\cellcolor{gray!40}$S_\alpha$~$\uparrow$}& {\cellcolor{gray!40}$M$~$\downarrow$} &{\cellcolor{gray!40}$F_\beta$~$\uparrow$} &{\cellcolor{gray!40}$E_\phi$~$\uparrow$} & \multicolumn{1}{c|}{\cellcolor{gray!40}$S_\alpha$~$\uparrow$}& {\cellcolor{gray!40}$M$~$\downarrow$} &{\cellcolor{gray!40}$F_\beta$~$\uparrow$} &{\cellcolor{gray!40}$E_\phi$~$\uparrow$} & \multicolumn{1}{c|}{\cellcolor{gray!40}$S_\alpha$~$\uparrow$}& {\cellcolor{gray!40}$M$~$\downarrow$} &{\cellcolor{gray!40}$F_\beta$~$\uparrow$} &{\cellcolor{gray!40}$E_\phi$~$\uparrow$} & \multicolumn{1}{c}{\cellcolor{gray!40}$S_\alpha$~$\uparrow$}\\ \midrule
			\multicolumn{21}{c}{1/8 Labeled Training Data} \\ \midrule
			SAM~\cite{kirillov2023segment}        & 0.077                                 & 0.696                                 & 0.729                                 & 0.734                                 & 0.103                                 & 0.713                                 & 0.741                                 & 0.737                                 & 0.103                                 & 0.728                                 & 0.741                                 & 0.744                                 & 0.119                                 & 0.651                                 & 0.657                                 & 0.657                                 & 0.116                                 & 0.695                                 & 0.716                                 & 0.712                                 \\
SAM-S~\cite{chen2023sam}      & 0.071                                 & 0.720                                 & 0.753                                 & 0.745                                 & 0.093                                 & 0.756                                 & 0.787                                 & 0.750                                 & 0.095                                 & 0.753                                 & 0.765                                 & 0.761                                 & 0.108                                 & 0.677                                 & 0.685                                 & 0.672                                 & 0.105                                 & 0.713                                 & 0.728                                 & 0.720                                 \\
PGCL~\cite{basak2023pseudo}       & 0.071                                 & 0.695                                 & 0.785                                 & 0.754                                 & 0.057                                 & 0.810                                 & 0.858                                 & 0.821                                 & 0.055                                 & 0.876                                 & 0.871                                 & 0.869                                 & 0.056                                 & 0.853                                 & 0.862                                 & 0.836                                 & 0.076                                 & 0.779                                 & 0.847                                 & 0.805                                 \\
EPS~\cite{lee2023saliency}        & 0.070                                 & 0.702                                 & 0.791                                 & 0.766                                 & 0.058                                 & 0.806                                 & 0.852                                 & 0.818                                 & 0.057                                 & 0.869                                 & 0.878                                 & 0.872                                 & 0.057                                 & 0.849                                 & 0.859                                 & 0.832                                 & 0.076                                 & 0.786                                 & 0.853                                 & 0.812                                 \\
CoSOD~\cite{chakraborty2024unsupervised}      & 0.066                                 & 0.717                                 & 0.807                                 & 0.779                                 & 0.055                                 & 0.817                                 & 0.870                                 & 0.826                                 & 0.055                                 & 0.878                                 & 0.888                                 & 0.878                                 & 0.054                                 & 0.862                                 & 0.878                                 & 0.845                                 & 0.074                                 & 0.806                                 & 0.861                                 & 0.818                                 \\
\rowcolor{c2!20}DTEN++     & 0.066                                 & 0.726                                 & {\color[HTML]{00B0F0} \textbf{0.819}} & 0.780                                 & 0.054                                 & 0.806                                 & 0.863                                 & 0.822                                 & 0.053                                 & 0.882                                 & 0.883                                 & 0.872                                 & 0.057                                 & 0.859                                 & 0.863                                 & 0.841                                 & 0.075                                 & 0.783                                 & 0.863                                 & 0.812                                 \\
\rowcolor{c2!20}PGCL++     & 0.066                                 & 0.708                                 & 0.804                                 & 0.772                                 & 0.052                                 & {\color[HTML]{00B0F0} \textbf{0.827}} & 0.872                                 & 0.833                                 & {\color[HTML]{00B0F0} \textbf{0.050}} & {\color[HTML]{00B0F0} \textbf{0.905}} & 0.903                                 & {\color[HTML]{00B0F0} \textbf{0.885}} & 0.055                                 & 0.868                                 & 0.880                                 & 0.852                                 & 0.072                                 & 0.803                                 & 0.870                                 & 0.818                                 \\
\rowcolor{c2!20}EPS++      & 0.067                                 & 0.719                                 & 0.803                                 & 0.777                                 & 0.053                                 & 0.818                                 & 0.867                                 & 0.831                                 & 0.051                                 & 0.893                                 & 0.898                                 & 0.879                                 & 0.055                                 & 0.864                                 & 0.871                                 & 0.849                                 & 0.073                                 & 0.805                                 & 0.867                                 & 0.820                                 \\
\rowcolor{c2!20}CoSOD++    & {\color[HTML]{FF0000} \textbf{0.063}} & {\color[HTML]{00B0F0} \textbf{0.730}} & 0.812                                 & {\color[HTML]{00B0F0} \textbf{0.786}} & {\color[HTML]{00B0F0} \textbf{0.051}} & 0.826                                 & {\color[HTML]{00B0F0} \textbf{0.885}} & {\color[HTML]{FF0000} \textbf{0.837}} & 0.051                                 & 0.902                                 & {\color[HTML]{00B0F0} \textbf{0.906}} & {\color[HTML]{00B0F0} \textbf{0.885}} & {\color[HTML]{00B0F0} \textbf{0.052}} & {\color[HTML]{00B0F0} \textbf{0.873}} & {\color[HTML]{00B0F0} \textbf{0.886}} & {\color[HTML]{00B0F0} \textbf{0.853}} & {\color[HTML]{FF0000} \textbf{0.071}} & {\color[HTML]{FF0000} \textbf{0.822}} & {\color[HTML]{00B0F0} \textbf{0.873}} & {\color[HTML]{FF0000} \textbf{0.831}} \\
\rowcolor{c2!20}SEE (Ours) & {\color[HTML]{FF0000} \textbf{0.063}} & {\color[HTML]{FF0000} \textbf{0.733}} & {\color[HTML]{FF0000} \textbf{0.822}} & {\color[HTML]{FF0000} \textbf{0.788}} & {\color[HTML]{FF0000} \textbf{0.050}} & {\color[HTML]{FF0000} \textbf{0.831}} & {\color[HTML]{FF0000} \textbf{0.889}} & {\color[HTML]{00B0F0} \textbf{0.835}} & {\color[HTML]{FF0000} \textbf{0.048}} & {\color[HTML]{FF0000} \textbf{0.909}} & {\color[HTML]{FF0000} \textbf{0.914}} & {\color[HTML]{FF0000} \textbf{0.891}} & {\color[HTML]{FF0000} \textbf{0.050}} & {\color[HTML]{FF0000} \textbf{0.877}} & {\color[HTML]{FF0000} \textbf{0.892}} & {\color[HTML]{FF0000} \textbf{0.858}} & {\color[HTML]{FF0000} \textbf{0.071}} & {\color[HTML]{00B0F0} \textbf{0.820}} & {\color[HTML]{FF0000} \textbf{0.876}} & {\color[HTML]{00B0F0} \textbf{0.826}}\\
   \midrule
			\multicolumn{21}{c}{1/16 Labeled Training Data} \\ \midrule
SAM~\cite{kirillov2023segment}        & 0.077                                 & 0.696                                 & 0.729                                 & 0.734                                 & 0.103                                 & 0.713                                 & 0.741                                 & 0.737                                 & 0.103                                 & 0.728                                 & 0.741                                 & 0.744                                 & 0.119                                 & 0.651                                 & 0.657                                 & 0.657                                 & 0.116                                 & 0.695                                 & 0.716                                 & 0.712                                 \\
SAM-S~\cite{chen2023sam}      & 0.075                                 & 0.708                                 & 0.742                                 & 0.739                                 & 0.099                                 & 0.730                                 & 0.764                                 & 0.743                                 & 0.101                                 & 0.736                                 & 0.750                                 & 0.749                                 & 0.113                                 & 0.664                                 & 0.671                                 & 0.665                                 & 0.111                                 & 0.705                                 & 0.723                                 & 0.715                                 \\
PGCL~\cite{basak2023pseudo}       & 0.074                                 & 0.677                                 & 0.757                                 & 0.758                                 & 0.065                                 & 0.783                                 & 0.857                                 & 0.820                                 & 0.064                                 & 0.847                                 & 0.857                                 & 0.858                                 & 0.065                                 & 0.825                                 & 0.845                                 & 0.835                                 & 0.080                                 & 0.797                                 & 0.839                                 & 0.810                                 \\
EPS~\cite{lee2023saliency}        & 0.073                                 & 0.684                                 & 0.761                                 & 0.763                                 & 0.067                                 & 0.772                                 & 0.851                                 & 0.817                                 & 0.063                                 & 0.852                                 & 0.861                                 & 0.865                                 & 0.062                                 & 0.837                                 & 0.858                                 & 0.839                                 & 0.085                                 & 0.782                                 & 0.822                                 & 0.797                                 \\
CoSOD~\cite{chakraborty2024unsupervised}      & 0.071                                 & 0.703                                 & 0.778                                 & 0.768                                 & 0.064                                 & 0.788                                 & 0.865                                 & 0.813                                 & 0.060                                 & 0.858                                 & 0.869                                 & 0.871                                 & 0.061                                 & 0.842                                 & 0.863                                 & 0.842                                 & 0.081                                 & 0.793                                 & 0.836                                 & 0.813                                 \\
\rowcolor{c2!20}DTEN++     & 0.068                                 & 0.710                                 & 0.783                                 & {\color[HTML]{00B0F0} \textbf{0.778}} & 0.065                                 & 0.788                                 & 0.858                                 & 0.815                                 & 0.062                                 & 0.854                                 & 0.853                                 & 0.865                                 & 0.057                                 & 0.848                                 & 0.875                                 & {\color[HTML]{FF0000} \textbf{0.850}} & 0.083                                 & 0.795                                 & 0.837                                 & 0.787                                 \\
\rowcolor{c2!20}PGCL++     & 0.071                                 & 0.697                                 & 0.764                                 & 0.768                                 & {\color[HTML]{00B0F0} \textbf{0.060}} & 0.803                                 & 0.869                                 & {\color[HTML]{FF0000} \textbf{0.828}} & 0.059                                 & 0.861                                 & 0.867                                 & 0.869                                 & 0.060                                 & 0.838                                 & 0.863                                 & 0.843                                 & {\color[HTML]{FF0000} \textbf{0.075}} & {\color[HTML]{FF0000} \textbf{0.812}} & 0.853                                 & 0.816                                 \\
\rowcolor{c2!20}EPS++      & 0.068                                 & 0.702                                 & 0.768                                 & 0.772                                 & 0.063                                 & 0.790                                 & 0.863                                 & 0.824                                 & 0.057                                 & 0.866                                 & 0.873                                 & 0.877                                 & 0.059                                 & 0.849                                 & 0.870                                 & 0.846                                 & 0.080                                 & 0.797                                 & 0.846                                 & 0.805                                 \\
\rowcolor{c2!20}CoSOD++    & {\color[HTML]{00B0F0} \textbf{0.067}} & {\color[HTML]{00B0F0} \textbf{0.715}} & {\color[HTML]{00B0F0} \textbf{0.785}} & 0.777                                 & {\color[HTML]{00B0F0} \textbf{0.060}} & {\color[HTML]{00B0F0} \textbf{0.806}} & {\color[HTML]{00B0F0} \textbf{0.875}} & 0.819                                 & {\color[HTML]{00B0F0} \textbf{0.055}} & {\color[HTML]{00B0F0} \textbf{0.871}} & {\color[HTML]{00B0F0} \textbf{0.878}} & {\color[HTML]{FF0000} \textbf{0.880}} & {\color[HTML]{FF0000} \textbf{0.056}} & {\color[HTML]{00B0F0} \textbf{0.853}} & {\color[HTML]{00B0F0} \textbf{0.875}} & 0.848                                 & 0.079                                 & 0.800                                 & {\color[HTML]{00B0F0} \textbf{0.852}} & {\color[HTML]{FF0000} \textbf{0.819}} \\
\rowcolor{c2!20}SEE (Ours) & {\color[HTML]{FF0000} \textbf{0.066}} & {\color[HTML]{FF0000} \textbf{0.717}} & {\color[HTML]{FF0000} \textbf{0.802}} & {\color[HTML]{FF0000} \textbf{0.779}} & {\color[HTML]{FF0000} \textbf{0.057}} & {\color[HTML]{FF0000} \textbf{0.810}} & {\color[HTML]{FF0000} \textbf{0.872}} & {\color[HTML]{00B0F0} \textbf{0.826}} & {\color[HTML]{FF0000} \textbf{0.053}} & {\color[HTML]{FF0000} \textbf{0.875}} & {\color[HTML]{FF0000} \textbf{0.886}} & {\color[HTML]{FF0000} \textbf{0.880}} & {\color[HTML]{FF0000} \textbf{0.056}} & {\color[HTML]{FF0000} \textbf{0.856}} & {\color[HTML]{FF0000} \textbf{0.877}} & {\color[HTML]{00B0F0} \textbf{0.849}} & {\color[HTML]{00B0F0} \textbf{0.078}} & {\color[HTML]{00B0F0} \textbf{0.806}} & {\color[HTML]{FF0000} \textbf{0.855}} & {\color[HTML]{00B0F0} \textbf{0.817}}\\\bottomrule[1.5pt]
	\end{tabular}}
	\label{table:SODQuantiSemi}
	\vspace{-0.3cm}
  \end{minipage}
\end{table*}

{\noindent\textbf{Generalization of HGFG}. As illustrated in Table~\ref{table:GeneralizationMFG}, we prove that our HGFG functions as a plug-and-play addition. Specifically, following the procedures of our method, we integrate HGFG into existing state-of-the-art segmenters, denoted with the suffix $^\dagger$. The new segmenters, equipped with HGFG, are trained using their original strategies.
Table \ref{table:GeneralizationMFGCOSW} shows that in the scribble-based weak supervision setting, HGFG enhances segmenter performance by $2.12\%$ for TEL \cite{liang2022tree} and $3.52\%$ for SCOD \cite{he2022weakly}. Besides, in the semi-supervised setting with 1/8 labeled training data, our HGFG improves the segmenter's performance by $3.00\%$ for EPS \cite{lee2023saliency} and $3.45\%$ for CoSOD \cite{chakraborty2024unsupervised}, as shown in Table~\ref{table:GeneralizationMFGCOSS}. Additionally, we evaluate the impact of HGFG under full supervision in Table~\ref{table:GeneralizationMFGCOSF} and discover segmenters with HGFG generally outperform their original versions by $3.2\%$ for FGANet \cite{zhaiexploring} and $3.0\%$ for FEDER \cite{He2023Camouflaged}. 
{\textit{(2) Salient object detection}. We also examine the application of HGFG in SOD. The results, presented in Tables~\ref{table:GeneralizationMFGSODW}, \ref{table:GeneralizationMFGSODS} and \ref{table:GeneralizationMFGSODF}, are based on the \textit{DUTS-test} dataset \cite{wang2017learning}. In the weakly-supervised setting, HGFG boosts the performance of leading segmenters by $3.9\%$ for TEL \cite{liang2022tree} and $4.9\%$ for SCOD \cite{he2022weakly}. In addition, in the semi-supervised setting, the proposed HGFG module enhances existing methods by $3.45\%$ for EPS \cite{lee2023saliency} and $3.61\%$ for CoSOD \cite{chakraborty2024unsupervised}. For the fully-supervised segmenters, HGFG enhances performance by $3.2\%$ for PGNet \cite{xie2022pyramid} and $4.0\%$ for MENet \cite{wang2023pixels}. 
These positive results comprehensively demonstrate the superiority and high generalizability of HGFG.}
}

\subsection{ Incompletely Supervised Salient Object Detection}
{ 

\noindent \textbf{Weakly-supervised Salient Object Detection}. In the weakly-supervised setting, we use S-DUTS \cite{zhang2020weakly} and P-DUTS \cite{gao2022weakly} to train the models with scribble and point supervision and evaluate on five common benchmarks. As shown in Table~\ref{table:SODQuanti}, our SEE framework achieves the top performance across all benchmarks, greatly surpassing the second-best method, WS-SAM. Besides, we validate our generalizability and find that the SEE framework enhances the performance of SCOD and GenSAM, which are significantly higher than those achieved by WS-SAM.
It is worth noting that the effectiveness of the learnable SAM, specifically SAM-$\text{W}_{\text{S}}$ and SAM-$\text{W}_{\text{P}}$, is significantly improved in the SOD task. This improvement is likely since SOD tasks typically have more distinct discriminative features compared to COS tasks, making it easier to achieve performance gains through fine-tuning.

\noindent \textbf{Semi-supervised Salient Object Detection}. We further validate our effectiveness in the semi-supervised SOD task, with the results reported in Table~\ref{table:SODQuantiSemi}. As indicated in Table~\ref{table:SODQuantiSemi}, our SEE framework consistently achieves a leading position across all benchmarks. In addition, when integrating SEE into other cutting-edge methods, we observe substantial performance gains compared to their original versions. }

\section{Limitations and Future Work}
{The limitations of this work arise from the manually designed strategy for pseudo-label generation. While our approach achieves promising results, it presents challenges in generalizing to complex scenarios, such as concealed objects that are small or involve multiple instances.
For future work, we aim to enhance pseudo-label selection by incorporating not only pixel-level selection but also feature-level selection and refinement to improve the final pseudo-label quality.
Furthermore, our method relies heavily on prompts provided by the teacher model, which presents challenges in unsupervised settings. To address this, we plan to develop a novel SAM-based framework with self-promoting capabilities. This would enable the generation of high-quality pseudo-labels for segmentation without relying on additional prompts, thereby improving the robustness and applicability of our approach.
}

\section{Conclusions}
This paper presents the first unified method, SEE, for ISCOS, which comprises two key components. The first component is a mean-teacher framework that leverages SAM to generate pseudo-labels based on coarse masks produced by the teacher model. A series of pseudo-label generation, storage, and supervision strategies are employed to ensure training the segmenter with reliable pseudo-labels.
The second component is the HGFG module, which extracts nuanced discriminative cues, enhancing feature coherence through a grouping-based approach. HGFG addresses the incomplete segmentation problem and can accurately segment multiple objects.
Experiments demonstrate the superiority of our method.

\balance
\bibliographystyle{IEEEtran}
\bibliography{COS}

{\fontsize{8pt}{\baselineskip}\selectfont \noindent\textbf{Chunming He} received the B.S. degree from Nanjing University of Posts and Telecommunications, China, and the M.E. degree from Tsinghua University, China. He is currently a Ph.D. student at Duke University, Durham, USA. His research interests include computer vision and biomedical image analysis.}

{\fontsize{8pt}{\baselineskip}\selectfont \noindent\textbf{Kai Li} is a research scientist in Meta. He received his Ph.D. degree from Northeastern University, USA, and his master’s and bachelor’s degrees from Wuhan University, China. His research interests include computer vision.
}

{\fontsize{8pt}{\baselineskip}\selectfont \noindent\textbf{Yachao Zhang} works at School of Informatics, Xiamen University as an Assistant Professor. He received Ph.D. degree from Xiamen University, in 2022. His current research interests are machine learning and computer vision, including weakly supervised learning and 3-D scene understanding.
}

{\fontsize{8pt}{\baselineskip}\selectfont \noindent\textbf{Ziyun Yang} received his Ph.D. degree from Department of Biomedical Engineering at Duke University. He received his B.S. degree in Automation Engineering from Beijing Institute of Technology. His research interests include semantic segmentation and salient/camouflaged image detection.
}

{\fontsize{8pt}{\baselineskip}\selectfont \noindent\textbf{Youwei Pang} received the B.E. degree from Dalian University of Technology (DUT) in 2019.
He is pursuing the Ph.D. and is supervised by Prof. Lihe Zhang and Prof. Huchuan Lu (Fellow, IEEE) from the School of Information and Communication Engineering, DUT.
His current research interests include computer vision, neural network design, and object segmentation.}

{\fontsize{8pt}{\baselineskip}\selectfont \noindent\textbf{Longxiang Tang} is a master student at Shenzhen International Graduate School, Tsinghua University. He received his B.S. degree in software engineering from University of Electronic Science and Technology. His research interests cover multi-modal large language model and representation learning.}

{\fontsize{8pt}{\baselineskip}\selectfont \noindent\textbf{Chengyu Fang} received the B.S. degree in software engineering from Southwest University, Chongqing, China in 2024. Now, he is pursuing his M.S. degree at Shenzhen International Graduate School, Tsinghua University. His research interests include computer vision and image processing.}

{\fontsize{8pt}{\baselineskip}\selectfont \noindent\textbf{Yulun Zhang} received a B.E. degree from Xidian University, an M.E. degree from Tsinghua University, and a Ph.D. degree from Northeastern University, USA. He is an associate professor at Shanghai Jiao Tong University. His research interests include image/video restoration and large foundation model. He is an Area Chair for CVPR, ICCV, ECCV, NeurIPS, ICML, ICLR.}

{\fontsize{8pt}{\baselineskip}\selectfont \noindent\textbf{Linghe Kong} is a full professor with the Department of CSE at Shanghai Jiao Tong University (SJTU). He received his Ph.D. degree at SJTU, 2013.
His research interests include satellite network, big data, and artificial intelligence.}

{\fontsize{8pt}{\baselineskip}\selectfont \noindent\textbf{Xiu Li} received the Ph.D. degree from Nanjing University of Aeronautics and Astronautics in 2000. She is a Professor at Shenzhen International Graduate School, Tsinghua University. Her research interests include computer vision.
}

{\fontsize{8pt}{\baselineskip}\selectfont \noindent\textbf{Sina Farsiu} (Fellow, IEEE) received his PhD degree in electrical engineering from the University of California, Santa Cruz in 2005. He is now the Anderson-Rupp Professor of BME, with secondary appointments at the Departments of ECE and CS at Duke University. He has served as the Senior Area Editor of IEEE Transactions on Image Processing, Deputy Editor of the Biomedical Optics Express, and the Associate Editor of SIAM Journal on Imaging Sciences. 
He received 
the “Outstanding Member of the Editorial Board Award” from the IEEE Signal Processing Society in 2018. He is a Fellow of IEEE, Optica, SPIE, ARVO, and AIMBE.}
\end{document}